\documentclass[a4paper,oneside,12pt]{report}

\usepackage{makeidx}

\usepackage[left=4cm,right=2cm,top=2.5cm,bottom=2.5cm]{geometry}

\usepackage{setspace}
\makeindex

\usepackage{thesis}

\ifx\thesishead\undefined
\newcommand*{\thesishead}{}

 \usepackage[round]{natbib}
 \usepackage{amsthm}
 \usepackage{amssymb}
 \usepackage[leqno]{amsmath}
 \usepackage{eufrak}
 \usepackage{graphs}
 \usepackage{enumerate}
 \usepackage{algorithmic}
 \usepackage[plain]{algorithm}
 \usepackage{fancyvrb}
\usepackage{pstricks}
 \usepackage{subfigure}
\usepackage{enumerate}
\usepackage{bm}
\usepackage{graphicx}

\newcommand{\swl}[1]{\rotatebox{90}{\parbox{4cm}{\begin{flushleft}\vspace{-0.1cm}{#1}\vspace{-0.1cm}\end{flushleft}}}}

\DeclareMathOperator{\Ent}{Ent}

\newcommand{\down}[1]{\left\downarrow(#1)\right.}
\newcommand{\downnb}[1]{\left\downarrow#1\right.}
\newcommand{\downe}[1]{\left\downarrow_\vdash(#1)\right.}
\newcommand{\up}[1]{\left\uparrow(#1)\right.}

\DeclareMathOperator{\Sub}{Sub}
\DeclareMathOperator{\FIS}{FIS}
\newcommand{\Par}{\mathrm{Par}}
\newcommand{\IC}{\mathit{IC}}

 \newcommand{\R}{\mathbb{R}}
 \newcommand{\bra}[1]{\langle #1|}
\newcommand{\ket}[1]{|#1 \rangle}
\newcommand{\bracket}[2]{\langle #1| #2\rangle}
\newcommand{\inprod}[2]{\langle #1, #2 \rangle}

\newcommand{\sbra}[1]{\langle #1\|}
\newcommand{\sket}[1]{\|#1 \rangle}
\newcommand{\sbracket}[2]{\langle #1\| #2\rangle}

\newtheorem{prop}{Proposition}[chapter]

 \theoremstyle{definition}
 \newtheorem{defn}[prop]{Definition}
  \newtheorem{conj}[prop]{Conjecture}
 \newtheorem{example}[prop]{Example}
\theoremstyle{remark}

\fi

\usepackage{fancyhdr}
\title{Context-theoretic Semantics for Natural Language: an Algebraic Framework}
\author{Daoud Clarke}
\date{September 2007}

\begin{document}

\begin{titlepage}
\vspace*{2.5cm}\noindent
{\cmtitle Context-theoretic Semantics for Natural Language}\\
{\cmsection an Algebraic Framework}
\vspace{0.5cm}\\
\large{Daoud Clarke}
\vspace{0.5cm}\\
\large{Submitted for the degree of D.Phil.}\\
\large{University of Sussex}\\
\large{September 2007}
\end{titlepage}

\begin{titlepage}
\vspace*{2.5cm}\noindent
{\cmsection Declaration}
\vspace{0.5cm}\\
\noindent
I hereby declare that this thesis has not been and will not be, submitted in whole or in part to another University for the award of any other degree.
\vspace{2cm}\\
\noindent Signature:
\end{titlepage}

\begin{titlepage}
\vspace*{2.5cm}\noindent
{\cmtitle Context-theoretic Semantics for Natural Language}\\
{\cmsection an Algebraic Framework}
\vspace{0.5cm}\\
\large{Daoud Clarke}
\vspace{0.5cm}\\
\large{Submitted for the degree of D.Phil.}\\
\large{University of Sussex}\\
\large{September 2007}
\vspace{1cm}\\
{\cmsection Summary}

\singlespacing
\noindent\normalsize{Techniques in which words are represented as vectors have proved useful in many applications in computational linguistics, however there is currently no general semantic formalism for representing meaning in terms of vectors. We present a framework for natural language semantics in which words, phrases and sentences are all represented as vectors, based on a theoretical analysis which assumes that meaning is determined by context.

In the theoretical analysis, we define a \emph{corpus model} as a mathematical abstraction of a text corpus. The meaning of a string of words is assumed to be a vector representing the contexts it occurs in in the corpus model. Based on this assumption, we can show that the vector representations of words can be considered as elements of an algebra over a field. We note that in applications of vector spaces to representing meanings of words there is an underlying lattice structure; we interpret the partial ordering of the lattice as describing entailment between meanings. We also define the \emph{context-theoretic probability} of a string, and, based on this and the lattice structure, a \emph{degree of entailment} between strings. 

Together these properties form guidelines as to how to construct semantic representations within the framework. A \emph{context theory} is an implementation of the framework; in an implementation strings are represented as vectors with the properties deduced from the theoretical analysis.

We show how to incorporate logical semantics into context theories; this enables us to represent statistical information about uncertainty by taking weighted sums of individual representations. We also use the framework to analyse approaches to the task of recognising textual entailment, to ontological representations of meaning and to representing syntactic structure. For the latter, we give new algebraic descriptions of link grammar.}
\end{titlepage}

\onehalfspacing

\chapter*{Acknowledgements}

\textit{In the name of God, the Merciful the Compassionate. All praise is due to God, Lord of the worlds. He knows what is in the heavens and earth, and none can encompass any of His knowledge except as He wills. Oh God, teach us that which benefits us and benefit us by that which you teach us. Increase us in knowledge, and make us of benefit to mankind. Send Your prayers and deep peace upon our master Muhammad, the unlettered prophet and final messenger, until the end of time.}
\vspace{0.5cm}\\
I am indebted to many people for their help during my time at Sussex: firstly and most importantly to my supervisor David Weir for his support and encouragement, for many in-depth discussions and for his detailed criticism of the thesis.

I am grateful to many people for helpful discussions and suggestions: to Bill Keller, John Carroll, Peter Williams and all my friends and colleagues at the University of Sussex. I am also grateful to Mark Hopkins for discussions via e-mail and for alerting me to the possibility of using Fock space to represent syntax. 

Finally, I wish to thank my parents for supporting me in so many ways throughout my studies, including helping to proof-read the final text, and my wife for moral support in the final stages of writing up.
 \chapter*{Preface}

\textsf{\textsl{My Lord, increase me in knowledge!}}
\newline

\dropcap{I}\textsc{ndeed all praise} belongs to God. We ask for God's prayers and deep peace upon the final 
messenger Muhammad, the messenger of peace, love, mercy, justice and freedom, for it is only by his example that we may reach success, and without his guidance we would be lost. We ask that God may make this work a light and a mercy for mankind; that its benefits may shine out to all humanity and its defects may not cause harm; that He may encourage by it the seekers of His knowledge; and give success by it to those who fight for truth and justice.
 
 We begin in His name, and we pray for His help and guidance, for ``None shall encompass any of His knowledge except by His will''.  Finally, we ask the reader's forgiveness for any errors in what follows, for whatever is good is from Him, and whatever is bad is from ourselves.  
\tableofcontents




\part{The Context-theoretic Framework}
\chapter{Introduction}

This thesis deals with the philosophical and theoretical foundations of computational linguistics. We are interested in the nature of meaning in natural language and the ways in which meaning can be represented computationally, in particular the relationship between vector-based representations of meaning and logical representations.

In recent years, the abundance of text corpora and computing power has allowed the development of techniques to analyse statistical properties of words. These techniques have proved useful in many areas of computational linguistics, arguably providing evidence that they capture something about the nature of words that should be included in representations of their meaning. However, it is very difficult to reconcile these techniques with existing theories of meaning in language, which revolve around logical and ontological representations. The new techniques, almost without exception, can be viewed as dealing with vector-based representations of meaning, placing meaning (at least at the word level) within the realm of mathematics and algebra; conversely the older theories of meaning dwell in the realm of logic and ontology. It seems there is no unifying theory of meaning to provide guidance to those making use of the new techniques.

The problem appears to be a fundamental one in computational linguistics since the whole foundation of meaning seems to be in question. The older, logical theories often subscribe to a model-theoretic \index{logical semantics} philosophy of meaning \citep{Kamp:93, Blackburn:05}. According to this approach, sentences should be translated to a logical form that can be interpreted as a description of the state of the world. The new vector-based techniques, on the other hand, are often closer in spirit to the philosophy of ``meaning as context'',\index{meaning!as context} that the meaning of an expression is determined by how it is used. This is an old idea with origins in the philosophy of \cite{Wittgenstein:53},\index{Wittgenstein, Ludwig} who said that ``meaning just \emph{is} use'' \index{meaning!as use} and \cite{Firth:57},\index{Firth, J.~R.} ``You shall know a word by the company it keeps'', and the distributional hypothesis\index{distributional!hypothesis} of \cite{Harris:68}\index{Harris, Zellig}, that words will occur in similar contexts if and only if they have similar meanings. Whilst the two philosophies are not obviously incompatible --- especially since the former applies mainly at the sentence level and the latter mainly at the word level --- it is not clear how they relate to each other.

While the model-theoretic philosophy of meaning provides us with theories which allow a complete description of natural language from the word level to the sentence level and beyond, the same cannot be said for the philosophy of meaning as context. It is this philosophy that has inspired vector based techniques, yet there is currently no theory explaining how these vectors can be used to represent phrases and sentences. This lack of a firm theoretical foundation has far-reaching implications for computational linguists or engineers implementing systems that represent expressions using vectors.

Such a theoretical foundation would be applicable to a wide range of tasks involving natural language. The task of recognising textual entailment was developed as part of a PASCAL Challenge \index{entailment!textual!PASCAL Challenge} \citep{Dagan:05,Bar-Haim:06} in an attempt to identify a generic task that is inherent in a number of areas within natural language processing, including information retrieval, question answering, machine translation and paraphrase acquisition. The task is to determine, given two sentences or natural language expressions (called the \emph{text} and \emph{hypothesis} sentences), whether the first entails or implies the second, for example in the case of the two sentences
\begin{itemize}
\item \emph{Text:} Once called the ``Queen of the Danube,'' Budapest has long been the focal point of the nation and a lively cultural centre.
\item \emph{Hypothesis:} Budapest was once popularly known as the ``Queen of the Danube.''
\end{itemize}
the text sentence does entail the hypothesis. Finding a solution to this task necessarily means solving the majority of problems within computational linguistics and natural language processing because the task is so general.

The PASCAL Challenge provided a method of evaluating textual entailment systems using a large number of text-hypothesis pairs. A large proportion, 22 of the 41 entered runs, made use of corpus or web-based statistics, yet there is no linguistic theory of meaning that explains how to determine entailment between sentences using such statistics. We might be able to find vector representations for words or multi-word expressions by statistical analysis, but we are left without any guidelines about how sentences should be represented. Entailment systems making use of such statistics thus have to resort to somewhat ad-hoc methods tuned and evaluated empirically by their performance at the task. While this is fine from a practical perspective, it leaves a lot to be desired from a linguistic perspective, since we are left without a deeper understanding of the nature of language.

\index{context-theoretic!framework|(}
In this thesis we attempt to solve these problems by identifying a framework to provide guidelines as to how to deal with vector-based representations of meaning in a principled way. We were looking for specific properties from the framework, namely, we wanted the framework to:
\begin{itemize}
\item provide some guidelines describing in what way the representation of a phrase or sentence should relate to the representations of the individual words as vectors;
\item require information about the probability of a string of words to be incorporated into the representation;
\item provide a way to measure the degree of entailment\index{entailment!degree of} between strings based on the particular meaning representation;
\item be general enough to encompass logical representations of meaning\index{logical semantics};
\item be able to incorporate the representation of ambiguity and uncertainty,\index{uncertainty, representing} including statistical information such as the probability of a parse or the probability that a word takes a particular sense.
\end{itemize}
The framework itself does not provide a recipe for how to represent meaning in natural language, instead it provides restrictions on the set of possibilities. The advantage of the framework is in ensuring that techniques are used in a way that is well-founded in a theory of meaning. For example, given vector representations of words, there is not one single way of combining these to give vector representations of phrases and sentences, but in order to fit within the framework there are certain properties of the representation that need to hold. Any method of combining these vectors in which these properties hold can be considered within the framework and is thus justified according to the underlying theory; in addition the framework instructs us as to how to measure the degree of entailment between strings according to that particular method. In the second part of the thesis, we show how the framework can be applied to problems in natural language processing.

Implementations of the framework are called \emph{context theories} since we think of them as theories about the contexts that strings of the language occur in. By analogy with the term ``model-theoretic'' we use the term ``context-theoretic'' for concepts relating to context theories; in particular we will often call our framework ``the context-theoretic framework''.

\begin{figure}
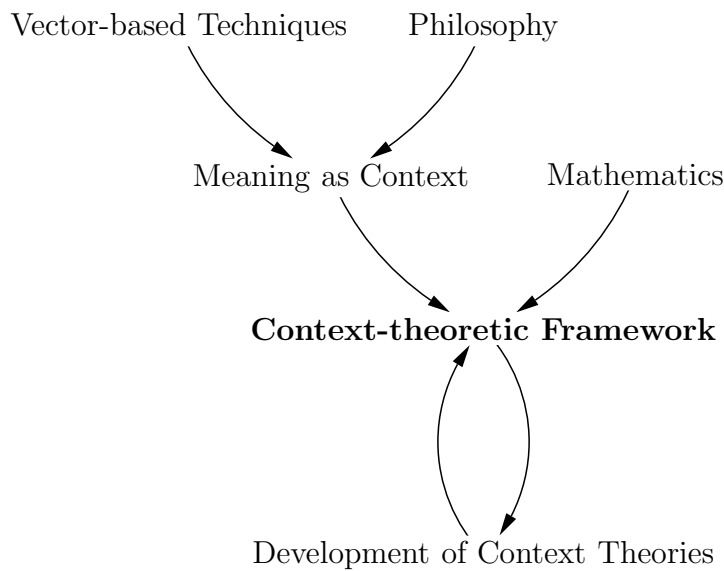

\begin{center}
\begin{graph}(8,8)(0,.5)
\newcommand{\ntext}[4]{\textnode{#1}(#2,#3){#4}[\graphlinecolour{1}]}
\ntext{Mean}{2}{6}{Meaning as Context}
\ntext{Exist}{0}{8}{Vector-based Techniques}
\ntext{Phil}{4}{8}{Philosophy}
\ntext{Math}{6}{6.05}{Mathematics}
\ntext{Cont}{4}{4}{\textbf{Context-theoretic Framework}}
\ntext{Dev}{4}{1}{Development of Context Theories}
\dirbow{Phil}{Mean}{0.1}
\dirbow{Exist}{Mean}{-0.1}
\dirbow{Math}{Cont}{0.1}
\dirbow{Mean}{Cont}{-0.1}
\dirbow{Cont}{Dev}{0.2}
\dirbow{Dev}{Cont}{0.2}
\end{graph}
\end{center}
\caption{Method of Approach in developing the Context-theoretic Framework.}
\label{approach}
\end{figure}

Our approach to identifying the framework can be divided into several components, as depicted in Figure \ref{approach}:
\begin{itemize}
\item We examine the philosophy of meaning as context, looking at the ideas of Wittgenstein, Firth and Harris as well as later developments to these ideas --- see Chapter \ref{background}. In this chapter, we also review statistical techniques that analyse occurrences of words in corpora to determine something about their meaning; such techniques can usually be viewed as representing meaning in terms of vectors. Specifically we look at latent semantic analysis and its variations, and measures of distributional similarity.
\item There are many areas of mathematics that could be of benefit to the problems we are addressing; in the Appendix we summarise those areas that are particularly relevant to our approach.
\item Based on the philosophy of meaning as context, and inspired by the statistical techniques, we develop a mathematical theory of meaning as context by making use of the abstract mathematical idea of a \emph{corpus model}, and we examine the mathematical properties of such models. This theory is vital in formulating the framework; in fact the framework can be viewed as a mathematical abstraction of the properties of the theory (see Section \ref{model-meaning-context}).
\item In Section \ref{context-theoretic-framework} we abstract the theory of meaning as context to define the context-theoretic framework, based on an analysis of the features that were important to include in the framework and the mathematics presented in the Appendix.
\item The second half of the thesis is devoted to describing applications of the context-theoretic framework, in order to demonstrate its usefulness in describing natural language (these are summarised in Table \ref{context-theories-table}). The applications were developed simultaneously with the framework itself; this also helped us to identify which features were important to include in the framework. The areas we look at are as follows:
\begin{itemize}
\item In Chapter \ref{entailment-chapter} we look at the application of the framework to the task of recognising textual entailment, comparing our framework to the approach of others and showing how several existing approaches can be described in terms of context theories.
\item In Chapter \ref{model-theoretic-chapter} we show how the framework can be used to extend standard logical semantics for natural language to include statistical information about uncertainty of meaning.
\item In Chapter \ref{ontologies} we discuss the relationship between ontological representations of meaning and vector-based representations, and show how to construct vector-based representations of meaning from a taxonomy.
\item in Chapter \ref{syntax-chapter} we show how syntactic structure can be represented within the framework, leading to new representations for syntax, and potentially new techniques for statistical parsing of natural language.
\end{itemize}

\end{itemize}

\begin{table}
\begin{center}
\begin{tabular}{|l|p{6cm}|l|}
\hline
\textbf{Context Theory} & \textbf{Purpose} & \textbf{Section}\\
\hline\hline
Document projections & Relate \citeauthor{Glickman:05}'s (\citeyear{Glickman:05}) approach to the task of recognising textual  entailment to the context-theoretic framework. &\ref{document-projections}\\
\hline
Subsequence matching & Estimate the degree of entailment based on the number of shared subsequences. & \ref{subsequence}\\
\hline
Lexical overlap & Relate the degree of lexical overlap, commonly used as a baseline in the task of recognising textual entailment, to the framework. & \ref{subsequence}\\
\hline
Projections for Logic & Represent logical sentences within the framework, allowing statistical information about ambiguity and uncertainty to be incorporated. & \ref{bayesian-uncertainty-section}\\
\hline
Ideal Projection Completion & Represent concepts from an ontology in such a way that words can be represented as weighted sums over the vector representation of its component senses. & \ref{ideal-projection}\\
\hline
Lambek Calculus & Represent syntactic categories in terms of the Lambek Calculus. & \ref{categorial-context}\\
\hline
Link Grammar & Describe syntax in terms of link grammar as operators on a Hilbert space. & \ref{link-context-1}\\
\hline
Semigroups & Construct a context theory from any semigroup. & \ref{semigroup-context}\\
\hline
\end{tabular}
\caption{The context theories described in the thesis, together with a summary of their purpose and the location of their full descriptions.}
\label{context-theories-table}
\end{center}
\end{table}

In summary, the major contributions of the thesis are as follows:
\begin{itemize}
\item the development of a mathematical theory of meaning as context that solidifies ideas implicit in existing philosophies and techniques;
\item the identification of a framework for natural language semantics that abstracts the salient features from the theory of meaning, providing guidelines for implementations that make use of vector-based representations of meaning;\index{context-theoretic!framework|)}
\item a demonstration of the application of the framework to important problems in natural language processing, most importantly the representation of statistical information about uncertainty and ambiguity in logical semantics.
\item in the development of applications of the framework some theoretical discoveries were made:
\begin{itemize}
\item in Chapter \ref{ontologies} we describe \emph{vector lattice embeddings} of partial orderings --- that is, ways to associate vectors with elements of a partially ordered set such as a taxonomy describing a hierarchy of concepts in such a way that the partial ordering is preserved;
\item in Chapter \ref{syntax-chapter} we provide new ways of describing link grammar both in terms of operators on a Hilbert space and in terms of inverse semigroups.
\end{itemize}
\end{itemize}

\chapter{Background}
\label{background}

\section{Philosophy}
\label{philosophy}
\index{philosophy of meaning|(}
The development of a theory of meaning inevitably requires subscription to a philosophy of \emph{what meaning is}. We are interested in describing representations resulting from techniques that make use of context in order to determine meaning, therefore it is natural that we look for a philsophy in which meaning is closely connected to context. The closest we have found is in the ideas of \cite{Firth:57}, and before him, \cite{Wittgenstein:53}.

\subsection{Wittgenstein}
\index{Wittgenstein, Ludwig|(}

Wittgenstein was concerned with understanding language for the purpose of applying it to philosophy. He believed that many errors in philosophical reasoning arose out of an incorrect understanding of what meaning is. In \emph{Philosophical Investigations} Wittgenstein especially combats the idea that the meaning of a word is an object:
\begin{quote}
1. ``When they (my elders) named some object, and accordingly moved towards something, I saw this and I grasped that the thing was called by the sound they uttered when they meant to point it out.  Their intention was shown by their bodily movements, as it were the natural language of all peoples; the expression of the face, the play of the eyes, the movement of other parts of the body, and the tone of the voice which expresses our state of mind in seeking, having, rejecting, or avoiding something.  Thus, as I heard words repeatedly used in their proper places in various sentences, I gradually learnt to understand what objects they signified; and after I had trained my mouth to form these signs, I used them to express my own desires.''\footnote{A quotation from Augustine (Confessions, I.8.)}
 
These words, it seems to me, give us a particular picture of the essence of human language.  It is this: the individual words in language name objects --- sentences are combinations of such names. In this picture of language we find the roots of the following idea: Every word has a meaning.  The meaning is correlated with the word.  It is the object for which the word stands.
\end{quote}
He later continues, ``That philosophical concept of meaning has its place in a primitive idea of the way language functions''.

Wittgenstein's own idea of meaning is later expressed as follows:
\begin{quote}
43. For a large class of cases --- though not for all --- in which we employ the word ``meaning'' it can be defined thus: the meaning of a word is its use in the language.
\end{quote}
In other words, if we know exactly how a word should be used, then in general, we know its meaning.\index{meaning!as use} Note that Wittgenstein requires that we know the ``use'' of a word rather than merely the contexts it is used in. This implies a much stronger knowledge since it seems to require knowing the reason behind using a word in terms of the impact it will produce; knowing the contexts a word occurs in merely means we can list the particular situations in which the use of the word is appropriate.
\index{Wittgenstein, Ludwig|)}

\subsection{Firth}
\index{Firth, J.~R.|(}

\cite{Honeybone:05} describes Firth's perception of language:
\begin{quote}
\ldots Firth saw language as a set of  events which speakers uttered, a mode of action, a way of  ``doing  things'',  and  therefore  linguists should focus on speech events themselves. This rejected  the  common view that speech acts are only interesting for linguists to gain access to the  ``true'' object of study --- their underlying grammatical systems.

As utterances occur in real-life contexts, Firth argued that their meaning derived  just as much from the particular situation in which they occurred as from the string of  sounds  uttered.  This  integrationist  idea, which  mixes  language with the objects  physically present during a conversation to ascertain the meaning involved, is known  as Firth's ``contextual theory of meaning''\ldots
\end{quote}
This is summed up in the famous quote, ``You shall know a word by the company it keeps''  \citep{Firth:57}.

Firth comes closer to the idea of ``meaning as context''\index{meaning!as context} as it used in modern techniques in  computational linguistics in his article \emph{Modes of Meaning} \citep{Firth:57a} in discussing ``collocation'':
\begin{quote}
The following sentences show that a part of the meaning of the word \emph{ass} in modern colloquial English can be by collocation:
\begin{enumerate}
\item An ass like Bagson might easily do that.
\item He is an ass.
\item You silly ass!
\item Don't be an ass!
\end{enumerate}
One of the meanings of \emph{ass} is its habitual collocation with an immediately preceding \emph{you silly}, and with other phrases of address or of personal reference.
\end{quote}
He then clarifies the relationship between what he calls ``meaning by collocation'' and ``contextual meaning'':
\begin{quote}
It must be pointed out that meaning by collocation is not at all the same thing as contextual meaning, which is the functional relation of the sentence to the processes of a context of situation in the context of culture.
\end{quote}
For Firth, part of the meaning of a word may be determined by ``collocation'', but to know its meaning is to know its ``use'' in the general sense of Wittgenstein.

\index{Firth, J.~R.|)}


\subsection{Harris}
\index{Harris, Zellig|(}
\index{distributional!hypothesis|(}

Neither Wittgenstein nor Firth make strong statements connecting meaning to its observed textual context. The first to do this was 
 \cite{Harris:68}, whose work is often cited as first presenting the \emph{distributional hypothesis}: that words will occur in similar contexts if and only if they have similar meanings. Harris is the first to suggest that meanings of words can be determined by statistical analysis of their occurrences in large amounts of text.

He describes this idea as follows \citep[section 2.3 (b)]{Harris:85}:
\begin{quote}
The fact that, for example, not every adjective occurs with every noun can be used as a measure of meaning difference. For it is not merely that different members of the one class have different selections of members of the other class with which they are actually found. More than that: if we consider words or morphemes $A$ and $B$ to be more different in meaning than $A$ and $C$, then we will often find that the distributions of $A$ and $B$ are more different than the distributions of $A$ and $C$. In other words, difference of meaning correlates with difference of distribution.

If we consider \emph{oculist} and \emph{eye-doctor} we find that, as our corpus of actually occurring utterances grows, these two occur in almost the same environments\ldots In contrast, there are many sentence environments in which \emph{oculist} occurs but \emph{lawyer} does not: e.g.~\emph{I've had my eyes examined by the same oculist for twenty years}, or \emph{Oculists often have their prescription blanks printed for them by opticians}. It is not a question of whether the above sentence with \emph{lawyer} substituted is true or not; it might be true in some situation. It is rather a question of the relative frequency of such  environments with \emph{oculist} and with \emph{lawyer}, or of whether we will obtain \emph{lawyer} here if we ask an informant to substitute any word he wishes for \emph{oculist} (not asking which words have the same meaning).
\end{quote}
Harris also proposes the idea that similarity in meaning can be quantified in terms of the difference in their environments (contexts):
\begin{quote}
If $A$ and $B$ have almost identical environments except chiefly for sentences which contain both, we say they are synonyms: \emph{oculist} and \emph{eye-doctor}. If $A$ and $B$ have some environments in common and some not (e.g.~\emph{oculist} and \emph{lawyer}) we say that they have different meanings, the amount of meaning difference corresponding roughly to the amount of difference in their environments. (This latter amount would depend on the numerical relation of different to same environments, with more weighting being given to differences of selectional subclasses.) If $A$ and $B$ never have the same environment, we say that they are members of two different grammatical classes (this aside from homonymity and from any stated position where both these classes occur).
\end{quote}

There is a subtle distinction between the two statements
\begin{enumerate}
\item Words that have similar meanings will occur in similar contexts.
\item Words that occur in similar contexts will have similar meanings.
\end{enumerate}
Harris does not seem to make this distinction explicitly, however it is clear from the above passage that he intends both since he proposes that ``difference of meaning correlates with difference of distribution'' in addition to proposing that words with similar meanings occur in similar contexts. For this reason we have stated the distributional hypothesis as ``words will occur in similar contexts if and only if they have similar meanings''.

While Harris notes that distributional features extend beyond the sentence level, he does not attempt to extend the connection between meaning and context significantly beyond the word level. He also talks only about similarity in meaning, and does not discuss the asymmetric relationship of entailment, and how this relates to context.
\index{Harris, Zellig|)}
\index{distributional!hypothesis|)}

\subsection{Later Developments}

Harris's distributional hypothesis has been the inspiration for much of the statistical work on determining meaning from corpora. Very recently, attempts have been made to refine the distributional hypothesis.\index{distributional!hypothesis}

\cite{Weeds:04} take this one step further with the introduction of the idea of ``distributional generality''.\index{distributional!generality} A term $w_1$ is distributionally more general than another term $w_2$ if $w_2$ occurs in a subset of the contexts that $w_1$ occurs in. They relate this to their measures of precision and recall which they use to define a variety of measures of distributional similarity\index{distributional!similarity}.

The idea is that distributional generality may be connected to \emph{semantic generality}. An example of this is the \emph{hypernymy relation}\index{hypernymy} or ``is a'' relation between nouns: a word $w_1$ is a hypernym of $w_2$ if $w_1$ refers to a concept that generalises the concept referred to by $w_2$, for example the term \emph{animal} is a hypernym of \emph{dog} since a dog is an animal. They explain the connection to distributional generality as follows:
\begin{quote}
Although one can obviously think of counter-examples, we would generally expect that the more specific term \emph{dog} can only be used in contexts where \emph{animal} can be used and that the more general term \emph{animal} might be used in all of the contexts where \emph{dog} is used and possibly others. Thus, we might expect that distributional generality is correlated with semantic generality\ldots
\end{quote}

This has been refined by \cite{Geffet:05} with the introduction of two ``distributional inclusion hypotheses''.\index{distributional!inclusion hypotheses} They define these in terms of ``lexical entailment''\index{entailment!lexical} between senses of words, rather than the hypernymy relation which is more specific in meaning and is defined between words. They also only consider what they call ``syntactic-based features'' which would include, for example, dependency relations, and discount co-occurrences within a window as providing useful knowledge about entailment. Finally, they assume that it is possible to distinguish the ``characteristic'' features --- that is, those features that have an impact on the meaning of a word. Let $s_1$ and $s_2$ be two senses of words. Their hypotheses, then are:
\begin{enumerate}
\item If $s_1$ lexically entails $s_2$ then all the characteristic (syntactic-based) features of $s_1$ are expected to appear with $s_2$.
\item If all the characteristic (syntactic-based) features of $s_1$ appear with $s_2$ then we expect that $s_1$ lexically entails $s_2$. 
\end{enumerate}

The two hypotheses effectively tie the meaning (in terms of lexical entailment) to specific features of the contexts that terms occur in, however, the authors do not go so far as to attempt to equate the two.

\subsection{Discussion}

We view the ideas we have presented here as a progression in our understanding of meaning; this is not to say that each author was aware of the previous author's work, but that the ideas themselves relate to the previous ones. Wittgenstein\index{Wittgenstein, Ludwig} first attempted to free people from existing perceptions of meaning by proposing that knowledge of the meaning of a word meant nothing more than knowing how to use it. Firth\index{Firth, J.~R.} then proposed that part of the meaning of a word may be by collocation in his example of the word ``ass''. Harris\index{Harris, Zellig} went further in his proposal that words occur in similar contexts if and only if they have similar meanings. Recent work arising from computational techniques refines this idea by focussing on distributional and semantic generality, suggesting that a term with a more general meaning will occur in a wider range of contexts.

None of the authors go so far as to equate meaning with context: for example, Harris talks only about how meanings of words relate to one another, and that similarity and difference of meaning can be determined by examining the contexts words occur in. Thus Harris does not contradict earlier philosophers, since it is possible to know how the meanings of words relate to one another without knowing their meaning as Wittgenstein intended it.

For practical purposes of applications in computing however, we argue that knowing how meanings relate to one another is enough. This is something that has become clearer through the development of the notion of textual entailment, which can be applied to so many areas in natural language processing yet only requires a relative understanding of meaning. For this reason, in this thesis we will equate meaning with context, that is, we assume that a relative knowledge of meaning is sufficient. This is not a statement of our philosophical position, rather it is a simplification that is convenient for the problem we are addressing. We hope however that through this simplification and subsequent mathematical analysis we will be able to give a new perspective on meaning that can add to and enrich, rather than detract from, existing ideas of meaning.

\index{philosophy of meaning|)}
\section{Vector Based Representations of Meaning}
\label{vector-based}

By ``vector-based representations of meaning'' we really mean two main areas of research: that of latent semantic analysis\index{latent semantic analysis} and its variants, and that of measures of distributional similarity\index{distributional!similarity} between natural language expressions. In general, both these areas involve representing expressions in terms of vectors which are built according to the contexts that the expression of interest occurs in in some large corpus. Figure \ref{fruit} gives a sample of occurrences of the term ``fruit'' in the British National Corpus; typically context vectors are built from many more occurrences of a term.

In latent semantic analysis, a transformation is applied to the vectors, resulting in a new vector representation of an expression which is supposed to describe ``latent'' features of meaning of the expression. By contrast, measures of distributional similarity leave the initial vector representation intact, but use mathematical analysis to measure the similarity between these vectors in various ways.

Both techniques are dependent on how the initial vectors are built:
\begin{itemize}
\item The vector representation of an expression may depend purely on what document the expression occurs in: the representation is simply the multiset or bag of document identifiers corresponding to occurrences of the expression. The order of occurrences of words in a document is thus deemed unimportant in this model. Each dimension of the vector representation corresponds to a document in the corpus, and the size of a component of the representation of a word will be its frequency of occurrence in the corresponding document.
\item In a \emph{windowing model} the representation of an expression is built from words that occur within a certain ``window'' of $n$ words from the expression of interest; again order of occurrence is unimportant. Each dimension of the vector representation now corresponds to a different word that expressions may co-occur with.
\item The text may be parsed with a dependency parser and some or all of the resulting dependency relations are then used to build vectors. In this case, each dimension would correspond to a different relationship: a noun occurring as object of a verb would be in a different dimension to the same noun occurring as the subject of the verb.
\end{itemize}
The first of these relates closely to information retrieval applications, and it was this application that led to the development of latent semantic analysis; the second representation is also commonly used in latent semantic analysis. Variations on the third representation are more commonly used in measures of distributional similarity.

\begin{figure}
\begin{Verbatim}[fontsize=\scriptsize]
end some medicine for her, but she will need fruit and  milk, and some other special things that
our own. Here we give you ideas for foliage, fruit and  various festive trimmings that you can i
part II). However, other strategies can bear fruit  and are described under three sections which
       supper Ñ tomatoes, potato chips, dried fruit and cake. And  they drank water out of tea-cu
erent days, as  the East Berliners queue for fruit and cheap stereos, a Turkish  beggar sleeps i
dening; and  Pests -- how to control them on fruit and vegetables. Both are  produced by the Hen
me,"Silver Queen" is male so will never bear fruit    At the opposite end of the prickliness sca
 lifted away    Like an orange lifted from a fruit-bowl    And darkness, blacker    Than an oil-
ed in your  wreath. Christmas ribbon and wax fruit can be added for colour.  Essentials are scis
e you need to start developing your very own fruit  collection    KEEPING OUT THE COLD    Need e
ly with Jeyes fluid    THE KITCHEN GARDEN    FRUIT    Cut out cankers on fruit trees, except tho
wn and watered    AUTUMN HUES    Foliage and fruit enrich the autumn garden, whether glowing  th
- have forgotten  the maxim: " tel arbre tel fruit ". If I were  willing  to  unstitch the past 
 of three children of Alfred Roger Ackerley, fruit importer  of London, and his mistress, Janett
rful didactic spirit, much that was to  bear fruit in his years as a mature artist. Although thi
e all made with natural vegetable, plant and fruit ingredients  such as chamomile, kukai nut and
ack in the soup.    He re-visits the Copella fruit juice farm in Suffolk, the  business he told 
rategic relationship" with Lotus, the first  fruit of which is a mail gateway between Office and
, choose your plants  carefully to enjoy the fruit of your labour all year round.    PLACES TO V
 and I love chips.  Otherwise I'll nibble on fruit or something to convince myself  that I'm eat
 tone and felt the  softness and warmth of a fruit ripening against a wall? If she  had she migh
ol place to set. Calories per  slice: 395    Fruit Scones with cinnamon Butter    (makes 12)    
ought me water. Another monster gave me some fruit  to eat. A few monsters lay against my body a
ney fungus.    Cut out diseased wood on most fruit trees    VEGETABLES    Continue winter diggin
age and chafing.    Remove old, unproductive fruit trees by cutting them down to  shoulder heigh
ITCHEN GARDEN    FRUIT    Cut out cankers on fruit trees, except those on peaches, plums  and ch
ps  remain, then stir in the sugar and dried fruit. Using a round-  ended knife, stir in the mil
 of a homeland, well others dream too,    De fruit was forbidden an now yu can't chew,    How ca
onnoisseurs. We take a bite from an unusual  fruit. We come away neither nourished nor ravished,
\end{Verbatim}

\caption{Occurrences and some context of occurrences of the word \emph{fruit} in the British National Corpus.}
\label{fruit}
\end{figure}

\subsection{Latent Semantic Analysis}
\index{latent semantic analysis|(}

The technique of latent semantic analysis and the similar probabilistic techniques that followed it, arose from the work of \cite{Deerwester:90}, in the context of the task of information retrieval. We will give only a brief overview here, since the details are not directly relevant to our work.

It is common in information retrieval to represent a term by the vector of documents it occurs in. Table \ref{fruittable} gives a set of hypothetical occurrences of six terms in eight documents. Given a user query term, the information retrieval software may return the documents that have the most occurrences of that term. From the vector perspective, the documents are identified with the \emph{components} of the vector representation of a term; given a query term, the most suitable document corresponds to the \emph{greatest component} of the term's vector representation.

It is often the case, however, that there are documents that are suitable matches for a given query which do not contain that query term often, or even at all. These will not be returned by the straightforward matching technique, and latent semantic analysis aims to get around this problem. It aims to deduce ``latent'' information about where terms may be expected to appear, by reducing the number of dimensions in which vectors are represented. This is performed in such a way that the most important components of meaning are retained, while those thought to represent noise are discarded.  This \emph{dimensionality reduction} has the effect of moving vectors that were unrelated closer together, as they are ``squashed'' into a space of lower dimensionality. For example, in table \ref{fruittable}, \emph{banana} and \emph{orange} never occur together, however they both occur with \emph{apple} and \emph{fruit} which provides evidence that they are related. Latent semantic analysis aims to deduce this relation.

\begin{figure}
\begin{center}
\input{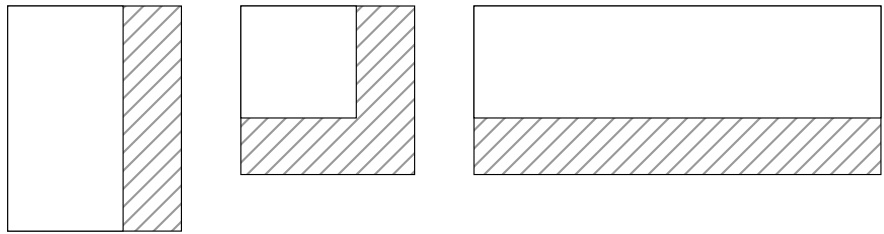}
\end{center}
\caption{Matrix decomposition and dimensionality reduction in latent semantic analysis.}
\label{reduce}
\end{figure}

Figure \ref{reduce} is intended to give an idea of how this works. The outer rectangles represent  the matrices arrived at by singular value decomposition, their product gives the original matrix representing the table of term-document co-occurrences. These matrices are arranged so that the most important information is stored in the top and left areas, with less important information being stored towards the bottom and right. In latent semantic analysis, a rectangle of the most important information is chosen (the inner rectangles); this information is kept and the remaining areas of the matrices (those shaded in the diagram) are discarded --- these are assumed to contain only noise information.
 
Table \ref{approx} shows the latent semantic analysis approximation to table \ref{fruittable}. In this case we chose to keep only two dimensions for the inner rectangles. We can see that in the new table, \emph{banana} and \emph{orange} now have components in common --- latent semantic analysis has forced them into a shared space. Because there were only two dimensions available, the term \emph{computer}, which before only shared components with \emph{apple} has been forced nearer to all the other terms, but remains closest to the term \emph{apple} as we would expect.

Latent semantic analysis works as follows. The matrix $M$ representing the original table can be decomposed into three matrices, $$M = UDV,$$ where $U$ and $V$ are unitary matrices and $D$ is a diagonal matrix containing the \emph{singular values} of $M$. Figure \ref{reduce} shows how the dimensionality reduction is performed. The decomposition can be rearranged so that the most important components --- those with the greatest singular values --- are in the top left of the matrix $D$, the dimensionality reduction is then performed by discarding the less important components, resulting in smaller matrices $U'$, $V'$ and $D'$. The matrix $M$ is then \emph{approximated} by the product of the new matrices, $M \simeq U'D'V'$.

For example, if we take table \ref{fruittable} as matrix $M$, then the decomposed and reduced matrices are those in figure \ref{decompose}. In this case we chose to keep only two dimensions corresponding to the greatest singular values ($12.8$ and $9.46$); keeping more dimensions would mean that more features of the original matrix would be preserved.
\begin{table}
\newcommand\T{\rule[-1.2ex]{0pt}{3.7ex}}

\begin{center}
\begin{tabular}{|l|cccccccc|}
\hline
	 	& $d_1$\T	& $d_2$	& $d_3$	& $d_4$	& $d_5$ & $d_6$ & $d_7$ & $d_8$\\
\hline
banana\T	& 2		& --		& --		& --		& 5		& --		& 5		& --\\
apple\T	& 4		& 3		& 4		& 6		& 3		& --		& --		& --\\
orange\T	& --		& 2		& 1		& --		& --		& 7		& --		& 3\\
fruit\T	& --		& 1		& 3		& --		& 4		& 3		& 5		& 3\\
tree\T	& --		& --		& 5		& --		& --		& 5		& --		& --\\
computer\T& --		& --		& --		& 6		& --		& --		& --		& --\\
\hline
\end{tabular}
\caption{A table of hypothetical occurrences of words in a set of documents, $d_1$ to $d_8$.}
\label{fruittable}
\end{center}
\end{table}

\begin{figure}
$$\left( \begin{array}{cc}
.335  &  -.175 \\
.504   & -.619 \\
.392   & .514\\
.564   & .177\\
.374   & .341\\
.141   & -.415 
\end{array}\right)
\left( \begin{array}{cc}
  12.8   &   0  \\
   0   &   9.46
\end{array}\right)
\left( \begin{array}{cc}
.209   &  -.298  \\
.223    & -.0686 \\
.466   & .0295 \\
.302   &  -.655  \\
.425   &  -.213  \\
.492   &  .617 \\
.351   & .00101\\
.224    & .219 \\
\end{array}\right)^\mathrm{T}$$
\caption{The matrices $U'$, $D'$ and $V'$ formed from singular value decomposition and dimensionality reduction. The product approximates the original matrix in table \ref{fruittable}. Here $A^\mathrm{T}$ is used to mean the transpose of matrix $A$.}
\label{decompose}
\end{figure}

\begin{table}
\newcommand\T{\rule[-1.2ex]{0pt}{3.7ex}}

\begin{center}
\begin{tabular}{|l|cccccccc|}
\hline
	 	& $d_1$\T	& $d_2$	& $d_3$	& $d_4$	& $d_5$ & $d_6$ & $d_7$ & $d_8$\\
\hline
banana\T & 1.40 & 1.08 & 1.95 & 2.40 &  2.19 & 1.09 &  1.51 & .597 \\
apple\T	& 3.11 & 1.85 & 2.84 & 5.80 &  4.00 & -.44 &  2.26 &  .17 \\
orange\T	& -.40 & .795 & 2.48 & -1.68 & 1.10 & 5.49 &  1.77 & 2.20 \\
fruit\T	& 1.02 & 1.50 & 3.41 & 1.08 &  2.71 & 4.60 &  2.53 & 1.99 \\
tree\T	& .041 & .847 & 2.33 & -.68 &  1.35 & 4.36 &  1.68 & 1.78 \\
computer\T& 1.56 & .679 & .731 & 3.13 &  1.62 & -1.53 & .635 & -.455\\
\hline
\end{tabular}
\caption{An approximation to the table obtained from a singular value decomposition followed by a dimensionality reduction to two dimensions.}
\label{approx}
\end{center}
\end{table}

Latent semantic analysis in its original form has some problems many of which have now been resolved to a large degree by new techniques. For example, the new, approximate matrix may contain negative values, as our example shows (table \ref{approx}). This is undesirable, as the matrix is intended to represent expected co-occurrence frequencies, and these cannot be negative; this is a result of the technique's lack of grounding in a sound probabilistic analysis of the situation.
\index{latent semantic analysis|)}
\subsection{Probabilistic Latent Semantic Analysis}
\index{latent semantic analysis!probabilistic|(}

\begin{figure}
\begin{center}
\input{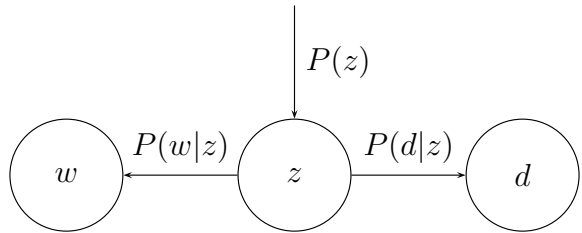}
\end{center}
\caption{The probabilistic latent semantic analysis model of words $w$ and documents $d$ modelled as dependent on a latent variable $z$.}
\label{plsa}
\end{figure}

Probabilistic latent semantic analysis \citep{Hofmann:99} is a technique which has the same aim as latent semantic analysis, but solves the problems of the technique in a probabilistic fashion, resolving the issue of negative values, and putting the technique on a firmer theoretical foundation. It treats the occurrence of a word $w$ and a document $d$ as random variables, and postulates the existence of a hidden variable $z$ (see figure \ref{plsa}), and makes the assumption that $d$ and $w$ are independent conditioned on $z$. The parameters of the model are the probability distributions $P(z)$, $P(w|z)$ and $P(d|z)$. As \cite{Hofmann:99} shows, these can be estimated by the Expectation Maximisation algorithm, using the following equations for the Expectation step
$$P(z|d,w) = \frac{P(z)P(d|z)P(w|z)}{\sum_{z' \in \mathcal{Z}}P(z')P(d|z')P(w|z')}$$
and the Maximisation step
\begin{eqnarray*}
P(w|z) &\propto& \sum_{d\in\mathcal{D}} n(d,w)P(z|d,w)\\
P(d|z) &\propto& \sum_{w\in\mathcal{W}} n(d,w)P(z|d,w)\\
P(z) &\propto& \sum_{d\in\mathcal{D}} \sum_{w\in\mathcal{W}} n(d,w)P(z|d,w)
\end{eqnarray*}
where $\mathcal{D}$ denotes the set of documents, $\mathcal{W}$ the set of words and $\mathcal{Z}$ the set of values that the hidden variable $z$ may take, and $n(d,w)$ represents the observed count of the number of occurrences of word $w$ in document $d$.

\cite{Hofmann:99} demonstrates the results of his analysis by selecting specific values for $z$ and showing the ten most probable words according to $P(w|z)$. For example, they identified two values of $z$ relating to the term ``power'' one which related to radiating objects in astronomy and one relating to electrical engineering (see Table \ref{power-table}).

\begin{table}
\begin{center}
\begin{tabular}{|c|c|}
\hline
$z_1$ & $z_2$\\
\hline\hline
POWER & load\\
spectrum & memori\\
omega & vlsi\\
mpc & POWER\\
hsup & systolic\\
larg & input\\
redshift & complex\\
galaxi & arrai\\
standard & present\\
model & implement\\
\hline
\end{tabular}
\caption{Most probable words given two topic variables relating to the term ``power'' (taken from \cite{Hofmann:99}).}
\label{power-table}
\end{center}
\end{table}


\index{latent semantic analysis!probabilistic|)}
\subsection{Latent Dirichlet Allocation}
\label{lda-section}
\index{latent Dirichlet allocation|(}

Latent Dirichlet allocation \citep{Blei:03} provides an even more in-depth Bayesian analysis of the situation. \citeauthor{Blei:03} claim that the problem with probabilistic latent semantic analysis is that there is an assumed finite number of documents. This is not the true situation, they claim: the documents available should be viewed as a sample from an infinite set of documents. In order to achieve this, they model documents as samples from a \emph{multinomial distribution} --- a generalisation of the binomial distribution.

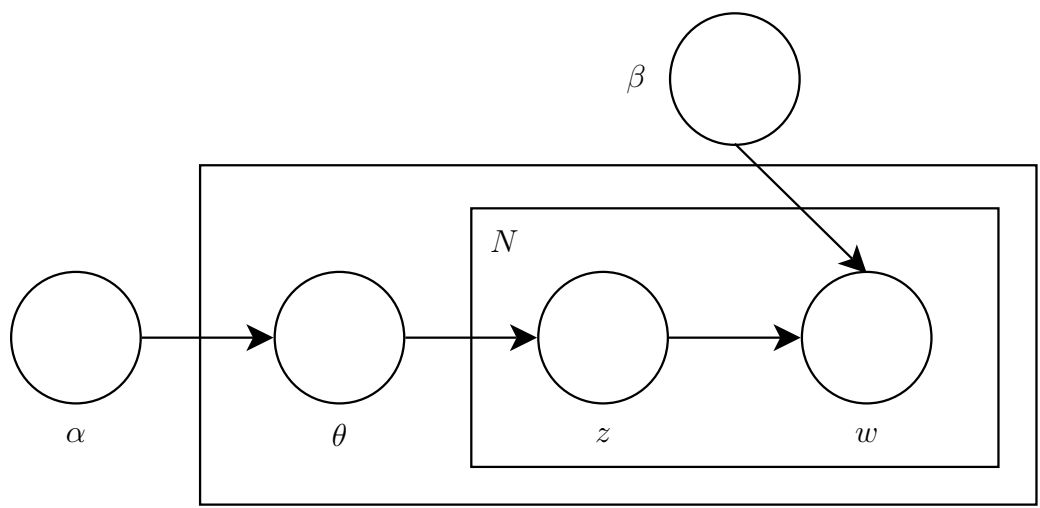
\begin{figure*}
\begin{center}
\psset{xunit=1mm,yunit=1mm,runit=1mm}
\psset{linewidth=0.3,dotsep=1,hatchwidth=0.3,hatchsep=1.5,shadowsize=1}
\psset{dotsize=0.7 2.5,dotscale=1 1,fillcolor=black}
\psset{arrowsize=1 2,arrowlength=1,arrowinset=0.25,tbarsize=0.7 5,bracketlength=0.15,rbracketlength=0.15}
\begin{pspicture}(0,0)(135,65)
\rput{0}(8.67,22.14){\psellipse[](0,0)(8.67,-8.57)}
\rput{0}(43.33,22.14){\psellipse[](0,0)(8.67,-8.57)}
\rput{0}(78,22.14){\psellipse[](0,0)(8.67,-8.57)}
\rput{0}(112.67,22.14){\psellipse[](0,0)(8.67,-8.57)}
\rput{0}(95.33,56.43){\psellipse[](0,0)(8.67,-8.57)}
\psline[arrowsize=3 2]{->}(52,22.14)(69.33,22.14)
\psline[arrowsize=3 2]{->}(17.33,22.14)(34.67,22.14)
\psline[arrowsize=3 2]{->}(86.67,22.14)(104,22.14)
\psline[arrowsize=3 2]{->}(95.33,47.86)(112.67,30.71)
\rput(8.67,9.29){$\alpha$}
\rput(43.33,9.29){$\theta$}
\rput(78,9.29){$z$}
\rput(112.67,9.29){$w$}
\rput(82.33,56.43){$\beta$}
\pspolygon[arrowscale=2 2](60.67,39.29)(130,39.29)(130,5)(60.67,5)
\rput(65,35){$N$}
\pspolygon[arrowscale=2 2](25,45)(135,45)(135,0)(25,0)
\end{pspicture}
\caption{Graphical representation of the Dirichlet model, adapted from \cite{Blei:03}. The inner box shows the choices that are repeated for each word in the document; the outer box the choice that is made for each document; the parameters outside the boxes are constant for the model.}
\label{graphical}
\end{center}
\end{figure*}




\begin{figure}
\begin{center}
\framebox[7.5cm]{\parbox{7cm}{
\begin{enumerate}
\item Choose $\theta \sim$ Dirichlet$(\alpha)$
\item For each of the $N$ words:
\begin{enumerate}
\item Choose $z \sim$ Multinomial$(\theta)$
\item Choose $w$ according to $p(w|z)$
\end{enumerate}
\end{enumerate}}}
\end{center}
\caption{Generative process assumed in the Dirichlet model}
\label{generative}
\end{figure}

Figure \ref{graphical} shows a graphical representation of the latent Dirichlet allocation generative model, and figure \ref{generative} shows how the model generates a document of length $N$. In this model, the probability of occurrence of a word $w$ in a document is considered to be a multinomial variable conditioned on a $k$-dimensional ``topic'' variable $z$. The number of topics $k$ is generally chosen to be much fewer than the number of possible words, so that topics provide a ``bottleneck'' through which the latent similarity in meaning between words becomes exposed.

The topic variable is assumed to follow a multinomial distribution parameterised by a $k$-dimensional variable $\theta$, satisfying $$\sum_{i=1}^k \theta_i = 1,$$
and which is in turn assumed to follow a Dirichlet distribution. The Dirichlet distribution is itself parameterised by a $k$-dimensional vector $\alpha$. The components of this vector can be viewed as determining the marginal probabilities of topics, since:
\begin{eqnarray*}
p(z_i) & = & \int p(z_i|\theta)p(\theta)d\theta\\
	   & = & \int \theta_i p(\theta)d\theta.
\end{eqnarray*}
This is just the expected value of $\theta_i$, which is given by
$$p(z_i) = \frac{\alpha_i}{\sum_j \alpha_j}.$$

The model is thus entirely specified by $\alpha$ and the conditional probabilies $p(w|z)$ which we can assume are specified in a $k\times V$ matrix $\beta$ where $V$ is the number of words in the vocabulary. The parameters $\alpha$ and $\beta$ can be estimated from a corpus of documents by a variational expectation maximisation algorithm, as described by \cite{Blei:03}.

Latent Dirichlet allocation was applied by \cite{Blei:03} to the tasks of document modelling, document classification and collaborative filtering. They compare latent Dirichlet allocation to several techniques including probabilistic latent semantic analysis; latent Dirichlet allocation outperforms these on all of the applications. Recently, latent Dirichlet allocation has been applied to the task of word sense disambiguation \citep{Cai:07,Boyd-Graber:07} with significant success.

\index{latent Dirichlet allocation|)}

\subsection{Measures of Distributional Similarity}

\index{distributional!similarity|(}

The use of distributional similarity measures (or often, more accurately, distance measures) has been an area of intense interest in computational linguistics in recent years \citep{Lin:98a,Lee:99,Curran:02,Kilgarriff:03,Weeds:04}. The technique arose from attempts to apply statistical techniques to Harris' distributional hypothesis \citep{Hindle:90,Pereira:93,Dagan:94}, and has been applied in many areas of computational linguistics, including automatic thesaurus generation \citep{Grefenstette:94, Lin:98a, Curran:02} and word sense disambiguation \citep{Dagan:97,McCarthy:04}. Distributional similarity has also been applied to the problems of determining relationships between phrasal patterns \citep{Lin:01} and detecting compositionality \citep{McCarthy:03}.

 A wide variety of measures have been suggested; we describe here some of the most commonly used. The variety of measures derives from the variety of ways of viewing the occurrences of words in their contexts; of these, some of the most important are as follows (see table \ref{simmeasures}):
\begin{itemize}
\item We can associate a vector $u$ with a word $w$ in the manner previously described; the components of the vector are the frequencies of occurrence of $w$ in each component $c$. Viewing occurrences in contexts from this perspective leads to measures based on the geometric properties of the vectors.
\item We can renormalise the vector $u$ to give a probability distribution $p$ over contexts. This leads to information theoretic measures of dissimilarity based on standard measures of the difference in probability distributions.
\item A consideration of which of the contexts are most important leads to measures which emphasise certain contexts over others, for example, mutual information may be used as an indication of which features are important.
\end{itemize}

\begin{table}
\newcommand\T{\rule[-2.1ex]{0pt}{5.2ex}}
\begin{center}
\begin{tabular}{|l|rcl|}

\hline
\emph{Measure} & \multicolumn{3}{|c|}{\emph{Formula}\T} \\
\hline\hline
Cosine\T & $\cos \theta$&=& $\frac{u\cdot v}{\|u\|\|v\|}$\\
Euclidean distance\T & $\|u - v\|$&=& $\sqrt{\sum_i (u_i - v_i)^2}$\\
City block distance\T & $\|u - v\|_1$ &=& $\sum_i |u_i - v_i|$\\
\hline
Kullback-Leibler\T & $D(p\|q)$ &=& $\sum_c p \log\frac{p}{q}$\\
Jenson-Shannon\T & $\mathrm{dist}_\mathit{JS}(q,p)$ &=& $\frac{1}{2}(D(p\|\frac{p+q}{2}) + D(q\|\frac{p+q}{2}))$\\
$\alpha$-skew\T & $\mathrm{dist}_\alpha(q,p)$ & = & $D(p\|(\alpha q + (1-\alpha)p))$\\
\hline
Jaccard's\T & $\mathrm{sim}_\mathit{ja}(w_2,w_1)$ &=& $\frac{|F(w_1) \cap F(w_2)|}{|F(w_1) \cup F(w_2)|}$\\
Jaccard's (MI)\T & $\mathrm{sim}_\mathit{ja+mi}(w_2,w_1)$ &=& $\frac{|S(w_1) \cap S(w_2)|}{|S(w_1) \cup S(w_2)|}$\\
Lin's\T & $\mathrm{sim}_\mathit{lin}(w_2,w_1)$ &=& $\frac{\sum_{S(w_1)\cap S(w_2)}I(c,w_1) + I(c,w_2)}{\sum_{S(w_1)} I(c,w_1) + \sum_{S(w_2)} I(c,w_2)}$\\
\hline
\end{tabular}
\caption{Eight measures of similarity and distance: geometric measures between vectors $u$ and $v$, where $u_i$ indicates the components of vector $u$, $u\cdot v$ indicates the dot product and $\|u\|$ denotes the Euclidean norm of $u$; measures based on the Kullback-Leibler divergence, where $p$ and $q$ are estimates of probability distributions describing the occurrences of words in contexts $c$; and measures based on the features of a word, either defined with respect to probability of occurrence, $F(w) = \{c : P(c|w) > 0\}$ or with respect to mutual information (this is also called the support of $w$), $S(w) = \{c : I(c,w) > 0\}$, where the mutual information $I$ is given by $I(c,w) = \log(P(c|w)/P(c))$.}
\label{simmeasures}
\end{center}
\end{table}

\subsubsection*{Geometric measures}

The most obvious are those with a clear geometric interpretation, namely measuring angles and distances between vectors (see table \ref{simmeasures}). The cosine of the angle between vectors is often used as a measure of similarity since it takes values between $0$ and $1$ and is equal to $1$ only when the vectors are exactly the same. The Euclidean distance is the measure familiar to us in physical space and the $L^1$ norm\index{L1 norm@$L^1$ norm} or ``city block'' distance corresponds to the distance measured using only vertical and horizontal lines (in two dimensions).

\subsubsection*{Information theoretic measures}

The more complex measures are more probabilistic in nature; vectors are normalised so that they can be considered as an estimate of a probability distribution over contexts. The basis of many of these measures is the Kullback-Leibler (KL) divergence $D(p\|q) = \sum_c p \log\frac{p}{q}$ \index{Kullback-Leibler divergence} of two distributions $p$ and $q$. This measures the inefficiency of describing the true distribution $p$ while assuming the distribution is $q$, and is thus an (asymmetric) measure of the difference between the two distributions. Using the KL divergence directly is not generally practical however, as it will be infinite if there is a context $c$ for which $q(c) = 0$ and $p(c) \neq 0$. The Jenson-Shannon and $\alpha$-skew measures get around this problem.

\subsubsection*{Feature-based measures}

The ``features'' of a word are those contexts which are considered to provide interesting information about the word. The features can simply be the contexts that occur with non-zero probability with a word, as used in Jaccard's coefficient, which measures the proportion of contexts occurring with either word that are shared by both words.  An alternative is to include only those contexts with positive mutual information, $I$, where $I(c,w) = \log(P(c|w)/P(c))$, this can be applied directly to the formula for Jaccard's coefficient,\index{Jaccard's coefficient} and also leads to Lin's measure,\index{Lin, Dekang} which is based on an information theoretic analysis of similarity \citep{Lin:98a}.

\index{distributional!similarity|)}

\section{Discussion}

The most important aspects of the work we have discussed for our 
purposes are those which they have in common --- they are all techniques 
which attempt to describe something about the meaning of a term based on the 
contexts the term appears in. The techniques are all flexible as to the 
exact interpretation of what context is --- for example, a window of text 
may be used or a bag of grammatical relations. The input to the techniques 
is a bag or multiset of pairs of terms and contexts. We can also view this 
input as representing a term as a function from the set of contexts to the 
natural numbers, or more generally, as positive, real-valued functions on 
the set of contexts. Equivalently, we can think of such functions as 
positive elements of a real vector space whose dimensionality is given by 
the number of contexts. It is this latter perspective that is the starting 
point for our theory of meaning as context that is developed in the next 
chapter. 
 
From here the techniques differ: latent semantic analysis and its variants 
attempt to transform these vectors to extract ``latent'' information about 
their meaning, while measures of distributional similarity leave the vectors 
as they are but make use of various methods to measure the similarity or 
difference between the vectors. Whilst our theory builds on what is common 
between the techniques, there are elements of each that will be of 
importance to us later. The concept of \emph{corpus model} that we describe 
in the next chapter is a generalisation of the generative model of a corpus 
that is used in latent Dirichlet allocation. Distributional similarity has 
made use of various norms; this is an important topic for us in the 
development of \emph{context theoretic probability} in the next chapter. We 
also discuss the relationship between certain measures of distributional 
similarity and context theories in Section \ref{dist-sim-projections-section}.

\chapter{Meaning as Context}
\label{meaning-context}
\index{meaning!as context}

We discussed in the previous chapter how vectors intended to represent the meaning of terms can be formed by looking at the contexts that terms appear in. However, these techniques do not provide any guidance as to how such vectors may be composed to form representations of larger constituents. Our approach to solving this problem is to build an abstract model of language based on the notion of meaning as context. In this chapter we first describe this model, in which both words and sequences of words are represented by vectors; we are then able to examine the mathematical properties of this model to providing guidelines as to how to combine vector representations of  words to form representations of phrases and sentences. These properties form the basis of the context-theoretic framework, described in the second part of this chapter.


In particular there are three key properties that will be incorporated into the framework:
\begin{itemize}
\item The vectors associated with strings can be endowed with a lattice structure,\index{lattice} making the object of study a \emph{vector lattice}.\index{vector lattice} This can be seen by looking in a very general manner at the way in which vectors in computational linguistics are derived. We shall interpret the associated partial ordering relation\index{partial ordering} of the lattice as \emph{entailment};\index{entailment} thus the lattice structure can be thought of as carrying the ``meaning''.\index{meaning!and lattice structure}
\item We can define multiplication on the vector space in such a manner that the vector associated with the concatenation of two strings is the product of the vectors associated with each individual string. Remarkably, the multiplication makes the vector space an \emph{algebra over a field} \index{algebra!over a field}--- a structure which has been the object of much study in mathematics.
\item We shall show that according to this model, the size of the context vector of a term should correspond to its frequency of occurrence, we call this measure the \emph{context theoretic probability}\index{context-theoretic!probability} of a vector, denoted $\phi$. This value makes two probability spaces from context vectors in two separate ways: the lattice structure of the vector space can be viewed as a (traditional, measure-theoretic) probability space using $\phi$, while the algebra becomes a \emph{non-commutative probability space} with $\phi$ as a linear functional (see Section \ref{operators}).
\end{itemize}
These properties put strong requirements on the nature of an algebra to represent natural language; and it is these properties that will be required of any implementation of our framework. We will also show how a \emph{degree of entailment} can be defined in terms of context vectors according to the ideas of distributional generality described previously. Later, when we discuss implementations of the framework, the same definition of the degree of entailment can be employed by the implementations because they have the same properties as the structure we derive in this chapter. This approach ensures that we can measure entailment for any implementation of the framework in a manner consistent with the context-theoretic philosophy.

In this chapter we will be making use of mathematical concepts that are summarised in the appendix; bold entries in the index indicate the page number of the relevant definition.

\section{A Model of Meaning as Context}
\label{model-meaning-context}

We wish to build an abstract mathematical model based on techniques which build vector representations of words in terms of their contexts.
We choose a very simple definition of what we mean by ``context'' --- the context of a string will be the pair of strings surrounding that string on either side in a document, that is, the whole document except the string itself. While this definition of context does not correspond directly to that used in the techniques, simplifying the definition of context allows us to easily examine the mathematical properties of our model. We will generalise the idea of a text corpus by assuming we have at our disposal an infinite amount of data, thus we do not attempt to overcome the problem of data sparseness that real-world techniques have to deal with. Because of this we are able to choose as ``context'' something that would be impractical in most applications --- using this definition most strings would share virtually no contexts in common given any real-life corpus.

We view a real world text corpus (a finite collection of documents) as a sample of some hypothetical infinite collection of documents. \index{corpus model|textbf}
 Specifically, we assume a \emph{probabilistic generative model} of corpora \citep{Blei:03}; one way to define such models is as follows:
\begin{defn}[Corpus Model]
A corpus model $C$ on a set $A$ of symbols is a probability distribution over $A^*$.
\end{defn} 

We can view a (real world) corpus as having been produced by a machine which repeatedly outputs strings according to the probability distribution $C$. Note that the machine is oblivious to what strings it has output previously; we can think of the individual strings output by the machine as \emph{documents}: the order of the strings is unimportant with respect to the machine, and typically the order of documents in a corpus is unimportant (whereas the order of sentences, for example, often is important). Of course it may be useful in practice to think of the strings as sentences, paragraphs or any other unit of text.

Abstracting in this way allows us to discover the nature of meaning as context according to our assumptions in the hypothetical situation of having an infinite amount of data available to us by analysing the mathematical properties of the resulting mathematical structure. It also allows us to make use of techniques which build corpus models from finite corpora, such as Latent Dirichlet Allocation\index{latent Dirichlet allocation} and associate meanings with strings according to the corpus model generated from the finite corpus.

\subsection{Meaning as Context}
\index{meaning!as context}

 How should we think about the meaning of an expression? For many applications in computational linguistics it suffices to know the relationships between the meanings of expressions: for example we should know if one entails another, or if two expressions are contradictory. For the purposes of what follows, we shall assume a purely \emph{relative} interpretation of the word ``meaning''; that is knowing the meaning of an expression means knowing how the expression relates to other expressions.

Techniques such as those discussed in the previous chapter typically build vector representations of meaning based on the context in which words or phrases appear; such representations only describe meaning in the relative sense described above.

Because of the problem of data sparseness,\index{data sparseness} these techniques typically only make use of a part of the context of a string, for example using a limited window and ignoring the order of words in this window. Because we are assuming we have at our disposal a corpus model in which data sparseness is not a problem, we instead make full use of the context: the context of an expression in a document is  everything surrounding the expression in the document.

\index{context vector|textbf}
Mathematically, the context vector of a string $x$ will be a function over pairs of strings $(u,v)$ with $u,v \in A^*$ such that $uxv$ is a document. More formally:
\begin{defn}
The context vector of a string $x \in A^*$ in a corpus model $C$ is a real-valued function $\hat{x}\in L^\infty(A^*\times A^*)$  on the set of contexts $A^* \times A^*$, defined by
$$\hat{x}(u,v) = C(uxv).$$
\end{defn}

\index{L999@$L^\infty$}
We stated here that the context vector of a string lives in the vector space $L^\infty(A^*\times A^*)$, that is, the set of bounded functions\index{bounded function} from $A^*\times A^*$ to the real numbers; we know that the functions are bounded because they are formed from the probability distribution $C$ (see Section \ref{lp-space-section}).

Thus, from this definition, we are able to associate with each string in $A^*$ a vector representing the contexts in which it occurs in the corpus model $C$, accordingly, we have all the properties of vector spaces at our disposal to study strings with respect to $C$: for example, we can add, subtract and scale their associated context vectors.


\subsection{Entailment}
\index{entailment|!and lattice structure(}

Consider the methods of forming vectors for terms described in the last chapter. In each method, vectors are formed from components which correspond to different contexts in which the term may occur: the components may correspond to words the term can occur with, or its possible dependency relations with other terms. What is important to note is that in computational linguistic applications we are always able to give an interpretation to each dimension; the exact method of determining the vectors is not of importance to us here.

In fact, this situation is somewhat special in comparison to other applications of vectors. For example, we live in a universe with three (observable) spatial dimensions. If we want, we can find a \emph{basis} (see \ref{basis})\index{basis} for this space, consisting of three vectors, $x$, $y$ and $z$, say, allowing us to locate any point in space by a linear combination of these vectors; equivalently we can decompose any vector into components with respect to this basis. However, in general, we don't have a \emph{preferred} choice of basis. There may be a basis which is convenient for us to use (for example we may choose $x$ and $y$ to be north and east and $z$ to be up, with a length of 1 meter, for some particular location on earth), but there is no fundamental reason for us to prefer that basis.

In contrast, in computational linguistics, we are automatically provided with a basis purely by the way in which vectors are formed from components. For example, if we build the vector representation of a term by looking at the words occurring in a certain window of text, the dimensionality of the resulting vector space will be the same as the number of different words, and by default we will make use of a basis which has a basis vector corresponding to each different word. This fact is so obvious that its importance has been overlooked, but in reality it has profound implications for the properties we should expect from vector spaces in computational linguistics.

Careful consideration of this fact, can, we believe, lead us to answer the following question: how can vector representations of meaning such as those obtained by latent semantic analysis and in measures of distributional similarity be reconciled with ontological and logical representations of meaning?\index{meaning!and lattice structure} The former make use of vector spaces while the latter make use of structures resembling those of lattice theory --- is there a way the two can be combined? This issue is touched on by \cite{Widdows:04}, where the implication is that the solution lies in generalising vector and lattice structures by weakening the mathematical requirements. In contrast, we will argue that all the necessary structure is already present and implicit in existing representations, which can be simultaneously be considered as vector spaces and lattices --- they are \emph{vector lattices} (see Section \ref{vector-lattices}).\index{vector lattice}

\begin{figure}
\begin{center}
\input{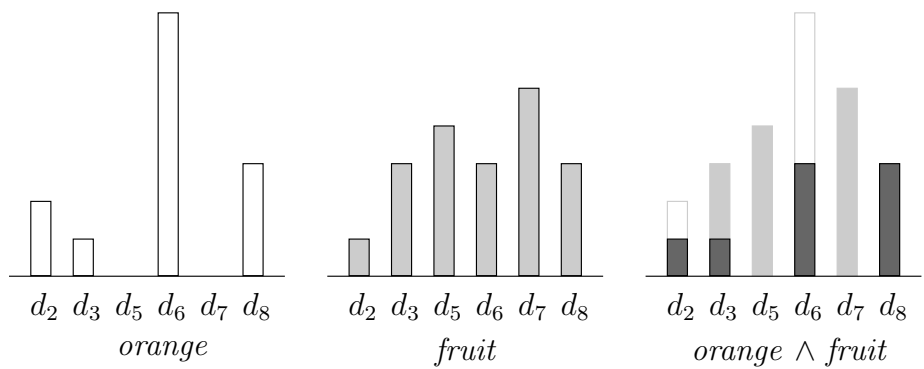}
\caption{Vector representations of the terms \emph{orange} and \emph{fruit} based on hypothetical occurrences in six documents (see the previous chapter) and their vector lattice meet (the darker shaded area).}
\label{orangefruit}
\end{center}
\end{figure}

Any vector space together with a basis can be considered as a vector lattice:\index{vector lattice} the meet and join operations can be defined as the component-wise minimum and maximum respectively. Figure \ref{orangefruit} shows two vectors representing the contexts of the terms \emph{orange} and \emph{fruit} based on their hypothetical occurrences in six documents, described in the previous chapter, and shows how their meet (component-wise minimum) is derived. Note that it is only because we are able to describe these vectors in terms of their components that we can define the lattice operations: the lattice operations are defined with respect to that particular basis, and if we had chosen a different basis the lattice operations would be different.

The vectors we discussed in the previous chapter were finite-dimensional; the context vector $\hat{x}$ of a string just defined is potentially infinite-dimensional. The same argument applies however: we can decompose the vector into components relating to individual contexts; for example, the basis vector corresponding to the context $(u,v)$, for $u,v \in A^*$ is the function which takes the value 1 on $(u,v)$ and 0 everywhere else on $A^*\times A^*$. Because we can decompose vectors in this way, we can again define lattice operations as component-wise minimum and maximum.

As with any lattice, there is an associated partial ordering; in this case, we write $\hat{x} \le \hat{y}$ if each component of $\hat{x}$ is less than or equal to the corresponding component of $\hat{y}$. In terms of contexts, this means that the string $x$ occurs in each context that $y$ occurs in at least as frequently. Relating this back to the concept of distributional generality discussed in the previous chapter, we may state the following hypothesis: \emph{a string fully entails another string if and only if the first occurs with equal or lower probability in all the contexts that the second occurs in}; or $x$ entails $y$ if and only if $\hat{x} \le \hat{y}$. It is this that we believe provides the link between vector space and ontological representations of meaning: the lattice structure already implicit in vector space representations of meaning can be viewed as describing the entailment relationship between concepts in a similar manner to an ontology.

In fact, we expect the situation of a string fully entailing another string to occur rarely; it is more likely that a string will share a proportion of its contexts with other strings. In order to be able to describe such ``partial entailment'' we need to have a way of measuring the size of such vectors, and this is the topic of the next section.

\index{entailment!and lattice structure|)}

\subsection{Context-Theoretic Probability}


Probability theory is central to modern techniques in computational linguistics, and it is thus important that our framework can inform us about the probabilistic aspect of language. In fact, we will show that in our model, the probability of a string is intimately connected to the ``size'' of its vector representation, as long as we choose a particular measure of size, the $l^1$ norm; this norm is thus particularly well suited to the purposes of our framework.
The $l^1$ norm of a vector is simply the sum of the absolute value of its components:  if $u$ is a vector with components $u_i$ for $1 \le i \le n$, then the $l^1$ norm of $u$ is given by
$$\|u\|_1 = |u_1| + |u_2| +\ldots + |u_n|.$$

\begin{figure}
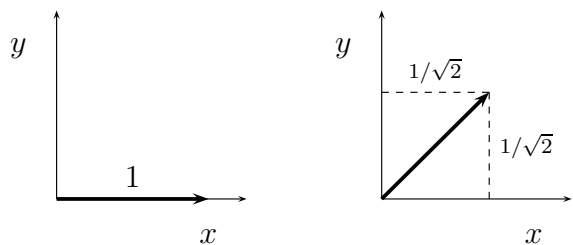

\begin{center}
\subfigure{\input{rotation.pst}}
\subfigure{\input{rotation2.pst}}
\caption{The length of a vector under the $l^1$ norm is not invariant under rotation.}
\end{center}
\end{figure}

 There are many norms we could choose --- why should the $l^1$ norm be special? The answer is that it has properties that are particularly well suited to our needs, linking the vector space to probability theory, while other norms have properties making them the most suitable in other applications of vector spaces.
For example, physical law in our three spatial dimensions has the special property that it is invariant with respect to rotation; this means that the $l^2$ norm occurs frequently in physical laws. The $l^2$ norm of a vector corresponds to the familiar Euclidean notion of its length: if $u$ is a vector with components $u_i$ for $1 \le i \le n$, then the $l^2$ norm of $u$ is given by
$$\|u\|_2 = (u_1^2 + u_2^2 +\ldots + u_n^2)^{1/2}.$$
It has the special property that lengths remain the same under rotation, which is something we expect to observe in our universe. 
To see that the $l^1$ norm, for example, doesn't preserve lengths under rotation, consider a vector in the $x$-$y$ plane which has a zero $y$ component and an $x$ component of 1. This vector has length 1 under both the $l^1$ and $l^2$ norms. Rotating this by $45^\circ$, however we find the length under the $l^1$ norm is $\sqrt{2}$, whereas under the $l^2$ norm, the length remains 1.

There is however no reason for us to expect the same properties for vectors in computational linguistics. We can consider other, more exotic norms, such as the generalisations of the $l^1$ and $l^2$ norms, the $l^p$ norms:
$$\|u\|_p = (|u_1|^p + |u_2|^p +\ldots + |u_n|^p)^{1/p},$$
where $1 \le p < \infty$, and the $l^\infty$ norm, where $\|u\|_\infty$ is the supremum over all components of $u$.

The $l^1$\index{L1 norm@$L^1$ norm} norm, though, has a special property with regards to vectors in computational linguistics. In the previous chapter, we saw  how, in practice, the vector representation of a term is built according to the frequencies of occurrence of that term in different contexts. In the simplest construction, the components of a vector are simply the frequencies of occurrence in each context.\footnote{Measures of distributional similarity often use more complex techniques to weight components of vectors.} Summing these frequencies is equivalent to summing the components of the vector representing the term; thus in the simplest methods of building vector representations we would expect this sum to be proportional to the frequency of occurrence of the term itself, or equivalently, proportional to its probability of occurrence.

In fact, there is a deeper connection to probability theory. Under the $l^1$ norm, the vector space becomes an \emph{Abstract Lebesgue}\index{abstract Lebesgue space} or \emph{AL} space (see section \ref{ALspace}) --- under the other $l^p$ norms this would not be the case. As the name suggests, the space can be considered as an abstraction of Lebesque spaces which form the foundation of measure theory, and hence the theory of probability. The key property is additivity of disjoint elements: in an AL space, if $x$ and $y$ are positive elements with $x \land y = 0$ then $\|x + y\| = \|x\| + \|y\|$. This is precisely the property we expect from probability: if we have two areas $A$ and $B$ in a Venn diagram which don't overlap, then we know that $P(A\cup B) = P(A) + P(B)$. In a vector lattice, we have
$$x \lor y = x + y - x \land y,$$
so if $x\land y = 0$ the above condition is the same as requiring $\|x \lor y\| = \|x\| + \|y\|$, an exact match for the Venn diagram requirement. Thus using the $l^1$ norm allows us to think of the vector space as simultaneously being a probability space. Although the structure is not what we normally think of as being a probability space (i.e.~a set of elements which we can interpret as events) the mathematical properties are the same, and this is what is so attractive about using the $l^1$ norm in computational linguistics applications: we can treat the lattice with the $l^1$ norm \emph{as if it is a probability space}.


Not all corpora are guaranteed to have finite $l^1$ norms. For example, consider the corpus model $C$ on $A = \{a\}$ defined by
$$C(a^{2^n}) = 1/2^{n+1}$$
for integer $n \ge 0$, and zero otherwise, where by $a^n$ we mean $n$ repetitions of $a$, so for example, $C(a) = \frac{1}{2}$, $C(aa) = \frac{1}{4}$, $C(aaa) = 0$ and $C(aaaa) = \frac{1}{8}$. Then $\|\hat{a}\|_1$ is infinite, since each non-zero document contributes $1/2$ to the value of the norm, and there are an infinite number of non-zero documents. To prevent difficulties we shall restrict ourselves to considering corpus models for which $\|\hat{\epsilon}\|_1$ is finite. This does not limit us much: note that $\|\hat{\epsilon}\|_1 = \sum_{x\in A^*}(|x| + 1)C(x)$ is just the mean of 1 + document length, over the documents in the corpus; the average document length of the above example is infinite.

We also make use of $\|\hat{\epsilon}\|_1$ to define the context theoretic probability:
\index{context-theoretic!probability|textbf}
\begin{defn}[Context-theoretic Probability]
The (context theoretic) probability $\phi$ is a real-valued linear function on $L^\infty(A^*\times A^*)$ defined by
$$\phi(u) = \frac{\|u\|_1}{\|\hat{\epsilon}\|_1}$$
\end{defn}
\noindent
Normalising in this way means that we can interpret $\phi(\hat{x})$ as the probability that $x$ occurs at a particular point in a document chosen at random.\footnote{This includes the end of the document, where no string except $\epsilon$ may ``occur''.} As we will see later, it also means that in addition to thinking of the vector space as a probability space with respect to the lattice operations, we can also think of it as a non-commutative probability space, with respect to a distributive multiplication which we shall define shortly.

The function $\phi$ is not guaranteed to be finite for all elements $u \in L^\infty(A^*\times A^*)$; however we are really interested in the value of $\phi$ on the vector space generated by context vectors. 
The following proposition shows that $\phi$ is defined on this space and clarifies the relationship between the context-theoretic probability of the context vector of a string and what we normally think of as the ``probability of a string'':
\begin{prop}
Let $C$ be a corpus model such that $\|\hat{\epsilon}\|_1$ is finite. Then:
\begin{enumerate}
\item for $x \in A^*$, $\phi(\hat{x})$ satisfies $\phi(\hat{x}) \le 1$ with $\phi(\hat{x}) = 1$ if and only if $x = \epsilon$.
\item $\sum_{a \in A} \phi(\hat{a}) < 1.$
\item If $v$ is a vector in $L^\infty(A^*\times A^*)$ constructed from context vectors using the vector and lattice operations, then $\phi(v)$ is finite.
\end{enumerate}
\end{prop}

\begin{proof}
\mbox{}
\begin{enumerate}
\item A document $d \in A^*$ contributes $(|d|+1)C(d)$ to $\|\hat{\epsilon}\|_1$. The string $x \in A^* - \{\epsilon\}$ can occur at most $|d| - |x| + 1$ times in $d$, so $d$ can contribute at most $(|d|- |x| + 1)C(d)$ to $\|\hat{x}\|_1$. Thus
$$\frac{\|\hat{x}\|_1}{\|\hat{\epsilon}\|_1} \le \frac{\sum_{d\in A^*} (|d| - |x| + 1)C(d)}{\sum_{d\in A^*} (|d|+1)C(d)} < 1,$$
i.e.~$\phi(\hat{x}) < 1$; we also have $\phi(\hat{\epsilon}) = 1$.

\item Note that 
$$\sum_{a\in A} \phi(\hat{a}) = \frac{\sum_{a\in A} \|\hat{a}\|_1}{\|\hat{\epsilon}\|_1}.$$
Call the numerator of the right hand side $S$. A document $d\in A^*$ must contribute $|d| C(d)$ to $S$ since there are $|d|$ symbols in the document. Thus 
$$S = \sum_{d\in A^*} |d| C(d) < \sum_{d\in A^*} (|d| + 1) C(d) = \|\hat{\epsilon}\|_1$$
and hence $\sum_{a \in A} \phi(\hat{a}) < 1$.

\item It follows from the first part of the proof that the  $L^1$ norm is finite for context vectors, thus they live in $L^1(A^*\times A^*)$. This space is closed under the vector and lattice operations, and so $\phi$ must be finite for all vectors generated by context vectors under the vector and lattice operations.
\end{enumerate}
\end{proof}

In contrast to our definition, when talking about the ``probability of a string'' in the context of language modelling, we would expect to find the property $\sum_{a \in A} \phi(\hat{a}) = 1$.
We can explain this by interpreting the value $\phi(\hat{x})$ in the following way. Consider a machine that outputs strings according to the probability distribution $C$, and at the end of each string outputs an additional symbol to denote the end of the document. Then $\phi(\hat{x})$ is the probability that if you stop the machine at a random point, the next $|x|$ symbols output by the machine will form the string $x$. In a language model, there would be no symbol denoting the end of a document and thus the sum of the probabilities of the symbols is 1. From another perspective, if we wished to encode the corpus model we would need an additional symbol to denote the end of a document; we can think of this additional symbol as absorbing the lost probability.

The benefit of defining $\phi$ in this way is that it allows us to relate our definition to the theory of non-commutative probability, for which it is necessary for the context-theoretic probability of the empty string to be 1; we discuss this more later in the chapter.



\subsection{Degrees of Entailment}
\index{entailment!degree of|(}

As we discussed previously, the performance of many tasks in computational linguistics rests on our ability to determine entailment between strings. Thus ultimately we are interested in being able to determine whether one string entails another based on their context vectors.
We propose that rather than having a black and white measure of entailment there should be \emph{degrees} of entailment. Within computational linguistics, this concept does not seem to have been developed. For example, in the Recognising Textual Entailment Challenges, participants were required to determine the existence or non-existence of entailment, together with a degree of confidence in the result. However within the theory of probability and logic, in particular in Bayesian interpretations of probability, the concept of degrees of entailment has been around for a long time \citep{Kyburg:01}.

 We wish to define a degree of entailment based on the context-theoretic probability. Like \cite{Glickman:05} we believe that entailment is closely connected to the nature of conditional probability. This is what we would expect from a Bayesian perspective; according to the Bayesian philosophy the correct formalism for reasoning about uncertainty is the mathematics of probability; from this perspective conditional probability can be viewed as a Bayesian implication. Because the $l^1$ norm together with the lattice operations define a an AL-space, the following definition of entailment has all the properties of a conditional probability:
\begin{defn}[Degree of Entailment]
The degree of entailment $\Ent(x,y)$ between two strings $x$ and $y$ is defined as
$$\Ent(x,y) = \frac{\phi(\hat{x}\land\hat{y})}{\phi(\hat{x})}$$
\end{defn}\noindent
when $\phi(\hat{x}) \neq 0$, and is undefined otherwise. This value is a measure of the degree to which the contexts string $x$ occurs in are shared by the contexts string $y$ occurs in.  According to this definition, complete entailment exists between $x$ and $y$ when $\Ent(x,y) = 1$, which will be true when $x \le y$. There will be no degree of entailment when $\Ent(x,y) = 0$, which is true when $x \land y = 0$.

As we will see in the following chapters, this definition provides us with a unified measure of the degree of entailment for any implementation of the framework we will define, based on the context-theoretic philosophy.

\index{entailment!degree of|)}


%



\subsection{Multiplication on Contexts}
\label{mult-contexts-section}
\index{algebra!over a field|(}


In this section we compare the representation of strings of words in the model to the representation of the individual words, and we are able to show that our model places strong restrictions between the two. 



A crucial feature of our definition is that it applies to strings as well as to individual symbols: strings of any size are attributed with a context vector. In particular, given two strings $x$ and $y$, not only do the strings have their own context vectors, but their concatenation $xy$ has a context vector $\widehat{xy}$ associated with it.  What we will show in this section is that context vectors can be considered as elements of an \emph{algebra over a field}, or simply \emph{algebra} (note that this is a much more specific sense of the word than is normally intended) --- a vector space together with multiplication such that the addition of the vector space distributes with respect to the multiplication (see Section \ref{algebras} for a formal definition). As far as we are aware this is a new, if fairly straightforward result, however it opens up the potential for the use of the extensive mathematics of algebras to studying corpus models in terms of their context vectors. More importantly for our purposes, it provides us with a concrete foundation for the context-theoretic framework --- because of the elegance and widespread nature of the mathematical structure of an algebra, we choose to require all context theories to incorporate an algebra to represent meaning.

This leads us to answer an important question: given vector representations for two strings, how are we to combine these representations to find a representation suitable for the concatenation of the strings, or more accurately, what ways of doing this should be considered suitable in the context-theoretic framework? We can think of this process as defining a product on the vector space: then the representation of the concatenation of two strings is the product of the individual representations. For example \cite{Clark:07} suggest that a suitable representation of the concatenation of two strings could be the tensor product of the representations of the individual strings. The following analysis will indicate that certain products are particularly suitable according to our model of meaning as context: namely those with respect to which the addition of the vector space distributes. Thus the tensor product of \citeauthor{Clark:07} would be acceptable according to the model since it satisfies this requirement of distributivity.


The question we are addressing is: does there exist some algebra $\mathcal{A}$ containing the context vectors of strings in $A^*$ such that $\hat{x}\cdot \hat{y} = \widehat{xy}$ where $x,y\in A^*$ and $\cdot$ indicates multiplication in the algebra? As a first try, consider the vector space $L^\infty(A^*\times A^*)$ in which the context vectors live. Is it possible to define multiplication on the whole vector space such that the condition just specified holds?

Consider the corpus $C$ on the alphabet $A = \{a,b,c,d,e,f\}$ defined by $C(abcd) = C(aecd) = C(abfd) = \frac{1}{3}$ and $C(x) = 0$ for all other $x \in A^*$. Now if we take the shorthand notation of writing the basis vector in $L^\infty(A^*\times A^*)$ corresponding to a pair of strings as the pair of strings itself then
\begin{eqnarray*}
\hat{b} &=& \tfrac{1}{3}(a,cd) +  \tfrac{1}{3}(a,fd)\\
\hat{c} &=& \tfrac{1}{3}(ab,d) +  \tfrac{1}{3}(ae,d)\\
\widehat{bc} &=& \tfrac{1}{3}(a,d)
\end{eqnarray*}
%
%
It would thus seem sensible to define multiplication of contexts so that $ \tfrac{1}{3}(a,cd)\cdot  \tfrac{1}{3}(ab,d) =  \tfrac{1}{3}(a,d)$. However we then find
$$\hat{e}\cdot \hat{f} =  \tfrac{1}{3}(a,cd)\cdot  \tfrac{1}{3}(ab,d) \neq \widehat{ef} = 0$$
showing that this definition of multiplication doesn't provide us with what we are looking for. In fact, if there did exist a way to define multiplication on contexts in a satisfactory manner it would necessarily be far from intuitive, as, in this example, we would have to define $(a,cd)\cdot (ab,d) = 0$ meaning the product $\hat{b}\cdot\hat{c}$ would have to have a non-zero component derived from the products of context vectors $(a,fd)$ and $(ae,d)$ which don't relate at all to the contexts of $bc$.


As an alternative to the approach of defining multiplication directly on contexts, we can consider instead defining multiplication on a \emph{subspace} of $L^\infty(A^*\times A^*)$, specifically the subspace generated by all context vectors, that is the space of all vectors that can be formed from the context vectors of strings by a countable number of additions and multiplications by scalars. This is in fact the subspace we are interested in, since in general we are interested in the relationships between meanings of words, described in terms of their context vectors.

Because we are interested in the context theoretic probability $\phi$ of strings, we will extend $\phi$ to all vectors in this subspace by requiring it to be linear: $\phi(\alpha_1 \hat{x}_1 + \alpha_2\hat{x}_2) = \alpha_1\phi(\hat{x}_1) + \alpha_2\phi(\hat{x}_2)$ for all $\alpha \in \R$ and $x_1, x_2 \in A^*$. Note that this doesn't contradict the earlier definition of $\phi$ because of the properties of the $l^1$ norm with respect to which $\phi$ is defined. We might want to consider infinite sums of context vectors, but we will not be interested in those which have infinite context theoretic probability, so we define the subspace $\mathcal{A}$ that we are interested in as follows:
\begin{defn}[Generated Subspace $\mathcal{A}$]
The subspace $\mathcal{A}$ of $L^\infty(A^*\times A^*)$ is the set defined by
$$\mathcal{A} = \{a : a = \sum_{x\in A^*}\alpha_x \hat{x}\text{ for some }\alpha_x \in \R\text{ and }|\phi(a)| < \infty\}$$
\end{defn}


Because of the way we define the subspace, there will always exist some basis\index{basis} $\mathcal{B} = \{\hat{u} : u \in B\}$ where $B \subseteq A^*$, and we can define multiplication on this basis by $\hat{u}\cdot\hat{v} = \widehat{uv}$ where $u,v \in B$. Defining multiplication on the basis defines it for the whole vector subspace, since we define multiplication to be linear, making $\mathcal{A}$ an algebra.

However there are potentially many different bases we could choose, each corresponding to a different subset of $A^*$, and each giving rise to a different definition of multiplication. Remarkably, this isn't a problem:


\index{context algebra|textbf}
\begin{prop}[Context Algebra]
Multiplication on $\mathcal{A}$ is the same irrespective of the choice of basis $B$.
\end{prop}
\begin{proof}

We say $B \subseteq A^*$ defines a basis $\mathcal{B}$ for $\mathcal{A}$ when $\mathcal{B} = \{\hat{x}: x\in B\}$. Assume there are two sets $B_1, B_2 \subseteq A^*$ that define corresponding bases $\mathcal{B}_1$ and $\mathcal{B}_2$ for $\mathcal{A}$. We will show that multiplication in basis $\mathcal{B}_1$ is the same as in the basis $\mathcal{B}_2$.

We represent two basis elements $\hat{u}_1$ and $\hat{u}_2$ of $\mathcal{B}_1$ in terms of basis elements of $\mathcal{B}_2$:
$$\hat{u}_1 = \sum_i \alpha_i \hat{v}_i \quad\text{and}\quad
\hat{u}_2 = \sum_j \beta_j \hat{v}_j,$$
for some $u_i \in B_1$, $v_j \in B_2$ and $\alpha_i, \beta_j  \in \R$.
 First consider multiplication in the basis $\mathcal{B}_1$. Note that $\hat{u}_1 = \sum_i \alpha_i \hat{v}_i$ means that $C(xu_1y) = \sum_i \alpha_i C(xv_iy)$ for all $x,y \in A^*$. This includes the special case where $y = u_2y'$ so $$C(xu_1u_2y') = \sum_i \alpha_i C(xv_iu_2y')$$ for all $x, y' \in A^*$.
Similarly, we have $C(xu_2y) = \sum_j \beta_j C(xv_jy)$ for all $x,y \in A^*$ which includes the special case $x = x'v_i$, so $C(x'v_iu_2y) = \sum_j \beta_j C(x'v_iv_jy)$ for all $x',y \in A^*$. Inserting this into the above expression yields
$$C(xu_1u_2y) = \sum_{i,j} \alpha_i\beta_j C(xv_iv_jy)$$
for all $x,y \in A^*$ which we can rewrite as
$$\hat{u}_1\cdot\hat{u}_2 = \widehat{u_1u_2} = \sum_{i,j}\alpha_i\beta_j (\hat{v}_i\cdot\hat{v}_j)
= \sum_{i,j}\alpha_i\beta_j \widehat{v_iv_j}.$$
Conversely, the product of $u_1$ and $u_2$ using the basis $\mathcal{B}_2$ is
$$\hat{u}_1\cdot \hat{u}_2 = \sum_i \alpha_i \hat{v}_i \cdot \sum_j \beta_j \hat{v}_j =  \sum_{i,j}\alpha_i\beta_j (\hat{v}_i\cdot\hat{v}_j)$$
thus showing that multiplication is defined independently of what we choose as the basis.
\end{proof}


Returning to the previous example, we can see that in this case multiplication is in fact defined on $L^\infty(A^*\times A^*)$ since we can describe each basis vector in terms of context vectors:
\begin{eqnarray*}
(a,fd)\cdot(ae,d) &=& 3(\hat{b} - \hat{e})\cdot 3(\hat{c} - \hat{f}) = -3(a,d)\\
(a,cd)\cdot(ae,d) &=& 3\hat{e}\cdot 3(\hat{c} - \hat{f}) = 3(a,d)\\
(a,fd)\cdot(ab,d) &=& 3(\hat{b} - \hat{e})\cdot 3\hat{f} = 3(a,d)\\
(a,cd)\cdot(ab,d) &=& 3\hat{e}\cdot 3\hat{f} = 0,
\end{eqnarray*}
thus confirming what we predicted about the product of $\hat{b}$ and $\hat{c}$.

\index{algebra!over a field|)}

\subsection{Discussion}

The fact that context vectors live in an algebra has profound implications for the nature of meaning according to the context-theoretic philosophy. The essential property is distributivity:\index{distributivity} the vector representations of two strings can be decomposed into components such that the vector representation of the concatenation of strings is the sum of the distributed product of the components.

This in fact places strong constraints on the theory of meaning. It means that if two words share a component of meaning, that component will remain in common between them when they are concatenated with another string (unless the component becomes zero on concatenation).

For example, we may assume the word \emph{square} has some component of meaning in common with the word \emph{shape}. Then we would expect this component to be preserved in the sentences \emph{He drew a square} and \emph{He drew a shape}. However, in the case of the two sentences \emph{The box is square} and \emph{*The box is shape} we would expect the second to be represented by the zero vector since it is not grammatical; \emph{square} can be a noun and an adjective, whereas \emph{shape} cannot. Distributivity of meaning means that the component of meaning that \emph{square} has in common with \emph{shape} must be disjoint with the adjectival component of the meaning of \emph{square}.

As we will see however, this constraint does not prevent us from representing many important properties of meaning in natural language; rather it provides us with guidelines as to how best to represent meaning according to the context-theoretic philosophy.

\subsection{Non-commutative Probability}

We have already stated that it is important for us that our framework is well grounded in probability theory. The context theoretic probability $\phi$ defines a probability space with respect to the vector lattice --- this is the most important aspect of the definition of $\phi$ since it allows us to define the degree of entailment. The results of the previous section allow us to think about the algebra $\mathcal{A}$ as an entirely different probabilistic structure, a \emph{non-commutative probability space}.
\index{non-commutative probability|textbf}
\begin{defn}[Non-commutative Probability]
A non-commutative probability space is a unital algebra (an algebra with unity 1) together with a linear functional $\psi$ such that $\psi(1) = 1$.
\end{defn}
A \emph{linear functional} is a linear function from a vector space to the real numbers. In our definition, $\hat{\epsilon}$ is a unity of the algebra, and the linear functional $\phi$ which we called the context theoretic probability satisfies $\phi(\hat{\epsilon}) = 1$, and thus $\mathcal{A}$ together with $\phi$ defines a non-commutative probability space. This means that we can think of context vectors as forming a probability space in two ways: they have a measure theoretic probability structure in terms of their vector lattice properties, and a non-commutative probability structure in terms of their algebraic properties, both defined with respect to the context theoretic probability $\phi$. The measure theoretic probability structure arises from the lattice operations $\land$ and $\lor$ together with the linear functional $\phi$, while the non-commutative probability space arises from the multiplication operation together with $\phi$.

The theory of non-commutative probability allows the description of a concept called freeness which is similar to independence in classical probability, but deals with non-commuting variables. It is our hope that freeness will eventually play an important role in applications of our theory by providing a method for the combination of context theories; for example, syntactic and semantic aspects of the representations of words may be considered to combine freely. Given a context theory for syntax and a context theory for lexical semantics, it may be possible to combine them using a free product of algebras.

Whilst there may be no immediate practical benefit from having this structure available, it allows our theory to be related to an established formalism for probability. The potential benefits of this relationship have led us to include it in the definition of the framework, although it may turn out to be an unnecessary inclusion.


\subsection{Further Work}

There are some unanswered questions relating to the theoretical properties of the model we have just described. We have shown that lattice operations can be defined on the vector space of possible contexts, and we have also shown that multiplication can be defined on the subspace of this vector space generated by context vectors to form an algebra; we have not, however, defined multiplication on the vector space generated by the lattice operations. This would be useful for us to show since this would make the space a \emph{lattice-ordered} algebra\index{algebra!lattice-ordered}; all the implementations of the context-theoretic framework\index{context-theoretic!framework} have this structure, thus we have included it in the framework. Proving that this structure is inherent in the model of meaning as context would give further justification for its inclusion. Instead we make the following conjecture:
\begin{conj}\label{conjecture}
Let $\mathcal{A}^{\land\lor}$ denote the vector lattice generated by a context algebra $\mathcal{A}$ under the lattice operations. There exists some multiplication on $\mathcal{A}^{\land\lor}$ that is an extension of the multiplication of $\mathcal{A}$ that makes it a lattice-ordered algebra.\index{algebra!lattice-ordered}
\end{conj}\noindent
Our attempts to prove this conjecture have not yet succeeded, nor have we been able to find a counter-example to disprove it. The difficulty in defining multiplication on this space lies in how to define it between two elements of  $\mathcal{A}^{\land\lor}$ which are not also elements of $\mathcal{A}$ --- if, for example, the left hand multiplier is assumed to always be in $\mathcal{A}$, we can use a definition akin to that used for operators on a vector lattice (see Section \ref{positive-operators}). A proof of this conjecture may be of benefit in understanding the theoretical underpinnings of the theory, for example throwing light on why it is difficult to define the product of the algebra on the space of contexts (as we attempted to do in Section \ref{mult-contexts-section}) rather than the subspace generated by context vectors of strings.

Another interesting theoretical question relating to the model is the question of completeness\index{completeness} with respect to the norm defined by the linear functional $\phi$; if the vector space is complete with respect to this norm then we would have a Banach space\index{Banach space} (see Section \ref{completeness-section}). However, what the implications of this would be for the nature of meaning as context are unclear; as far as we can see answering this question either way would have little impact on the practical use of the framework, though of course the long term benefits of answering such theoretical questions are hard to predict.

\section{The Context-theoretic Framework}
\label{context-theoretic-framework}


In this section we define the context theoretic framework based on the theory of meaning as context we have just discussed; the framework is formed from the central mathematical properties of the theory. These properties are derived from the assumption that the meaning of a string is purely determined by context; because of this, we can think of implementations of the framework as describing a theory about the contexts in which a string can occur --- for this reason we call such implementations ``context theories''.


There are certain things we require of the framework: it must provide guidelines about how to represent phrases and sentences, about determining the probability of a string and determining the degree of entailment between strings. Based on the preceding analysis of our model of meaning as context, we are now in a position to make a fuller set of requirements --- specifically we can identify those properties of the model which we wish to include in our framework:
\begin{itemize}
\item Words and strings of words should be represented as vectors. We may wish to make use of techniques such as latent semantic analysis to derive vector representations of words; this ensures that such representations can be incorporated, but requires that strings of words are also represented by vectors, based on our analysis in the previous chapter.
\item The vector space should in addition have a lattice structure. As we have seen, it is the lattice structure that informs us about entailment between strings, it is thus essential that this structure be incorporated into the framework.
\item The representation of the concatenation of strings can be viewed as a product of the representations of the individual strings for some distributive product (i.e.~the vector space forms an algebra\index{algebra!over a field}); this is a strong requirement to place on the mathematical structure. Imposing this structure is justified by the analysis of meaning as context and not only simplifies things from a mathematical perspective, but potentially opens up the vast amount of research available on these structures to be applied to computational linguistics.
\item There is a linear functional $\phi$ (the context-theoretic probability)\index{context-theoretic!probability} on the vector space such that the lattice operations together with $\phi$ can be used to define an AL-space.\index{abstract Lebesgue space} This requirement ensures that $\phi$ behaves like a probability with respect to the lattice operations. This is important since the degree of entailment is defined in terms of $\phi$ and the lattice operations. We wish the degree of entailment to have the form of a conditional probability, and placing this requirement ensures that this will be the case for any implementation of the framework.
\end{itemize}
It may be that over the course of time, other important properties will come to light, or some of these properties may not seem so important, and thus the framework will need to be revised. With this in mind, in addition to the above requirements we will require that the algebra together with $\phi$ defines a non-commutative probability space. Although the immediate benefits of this requirement are unclear there is some justification for it from the preceding analysis, moreover it has not proven a limitation in the development of applications of the framework: all the context theories we have developed naturally have this property.

We are able to combine these properties within the following definition:\index{context theory|textbf}\index{context-theoretic!framework|textbf}
\begin{defn}[Context Theory]\index{algebra!lattice-ordered}
A context theory for an alphabet $A$ is a unital lattice-ordered algebra $\mathcal{A}$ together with a semigroup homomorphism from $A^*$ to $\mathcal{A}$, denoted $a \mapsto \hat{a}$ and a positive linear functional $\phi$ such that $\phi(\hat{\epsilon}) = 1$.
\end{defn}
In addition we shall also often require that the set $I = \{u : \phi(u) = 0\}$ is a sub-vector lattice\index{sub-vector lattice} of $\mathcal{A}$ --- that is, a subspace of $\mathcal{A}$ that is a vector lattice under the same partial ordering. We call a context theory that satisfies this condition a \emph{strong context theory}.\index{context theory!strong|textbf}

This definition incorporates all the properties we require:
\begin{itemize}
\item A string $x$ is represented by the vector $\hat{x}$; requiring this to be a semigroup homomorphism ensures that we can view strings as elements of an algebra.
\item We require that the algebra is lattice-ordered. While the lattice structure is essential, requiring a lattice-ordered algebra is a stronger requirement; this would be justified in our theory if the conjecture at the end of the last chapter is proven correct. We have made this requirement since in practice it is not a limitation: all the structures we will describe in the second half of this thesis are naturally lattice-ordered algebras.
\item requiring $\phi(\hat{\epsilon}) = 1$  makes $\mathcal{A}$ together with $\phi$ a non-commutative probability space;
\item for a strong context theory we can define a norm on a space formed from $\mathcal{A}$ and $\phi$ that makes it an AL-space:
\end{itemize}
\begin{prop}[$\phi$-norm]
Given a strong context theory $\mathcal{A}$ with positive linear functional $\phi$ we can define a norm $\|\cdot\|_\phi$ on $\mathcal{A}/I$, where $I = \{u : \phi(u) = 0\}$ that defines an AL-space:
$$\|u\|_\phi = \phi(u^+) + \phi(u^-)$$
\end{prop}
\begin{proof}
The space $\mathcal{A}/I$ is the quotient space $\mathcal{A}/\equiv$ where $u\equiv v$ if $u - v \in I$, and is a vector lattice under the ordering of $\mathcal{A}$ \citep{Aliprantis:85}.\footnote{Effectively, it is the space formed by setting the subspace $I$ to zero.} The linear functional $\phi$ is well defined on this quotient space and satisfies $\phi(u) = 0$ if and only if $u$ is the zero of the quotient space. We need to show that $\|\cdot\|_\phi$ has the properties of a norm. For all $u \in \mathcal{A}/I$ we have:
\begin{itemize}
\item Positivity: $\|u\|_\phi \ge 0$ since $\phi$ is positive, and both $u^+$ and $u^-$ are positive.
\item Positive scalability: for $\alpha \in \R$, if $\alpha > 0$ then $\|\alpha u\|_\phi = \alpha\phi(u^+) + \alpha\phi(u^-)$. If $\alpha < 0$ then $(\alpha u)^+ = -\alpha u^-$ and $(\alpha u)^- = -\alpha u^+$ so $\|\alpha u\|_\phi = -\alpha\phi(u^+) - \alpha\phi(u^-)$. If $\alpha = 0$ then $\|\alpha u\|_\phi = 0$, thus for all $\alpha \in \R$, $\|\alpha u\|_\phi = |\alpha| \cdot \|u\|_\phi$.
\item Triangle inequality: we have $(u+v)^+ \le u^+ + v^+$ and $(u+v)^- \le u^- + v^-$. Then $\|u+v\|_\phi = \phi((u+v)^+) + \phi((u+v)^-) \le \phi(u^+ + v^+) + \phi(u^- + v^-) = \|u\|_\phi + \|v\|_\phi$.
\item Positive definiteness: it follows from $\phi(u) = 0$ if and only if $u = 0$ that $\|u\|_\phi = 0$ if and only if $u = 0$.
\end{itemize}
Finally, for $u,v \in \mathcal{A}$ with $u \ge 0$ and $v \ge 0$ we have $\|u + v\|_\phi = \phi(u+v) = \|u\|_\phi + \|v\|_\phi$ thus $\|\cdot\|_\phi$ defines an AL-space\index{abstract Lebesgue space} on $\mathcal{A}/I$.
\end{proof}


The requirements that we placed on a context theory ensured that the space is a probability space in two ways. In particular, the definition of the degree of entailment that we defined previously in the form of a conditional probability applies equally well in the case of a context theory. We restate it here for the case of a context theory:\index{entailment!degree of|textbf}
\begin{defn}[Degree of Entailment]
The degree of entailment $\Ent(x,y)$ between two strings $x$ and $y$ is defined as
$$\Ent(x,y) = \frac{\phi(\hat{x}\land\hat{y})}{\phi(\hat{x})}$$
when $\phi(\hat{x}) \neq 0$, and is undefined otherwise.
\end{defn}

\part{Context-theoretic Semantics for Natural Language}
\chapter{Textual Entailment}
\label{entailment-chapter}
\index{entailment!textual}


In this chapter we examine the task of recognising textual entailment from the context-theoretic perspective. Textual entailment recognition is the task of determining, given two sentences, whether the first sentence entails or implies the second. The task is particularly well suited to the application of the context-theoretic framework since it concerns detecting entailment between strings of words, which is what a context theory predicts. However, while the task requires determining the existence or non-existence of entailment, from the context-theoretic perspective it is more accurate to talk about a \emph{degree} of entailment\index{entailment!degree of} between strings. Nevertheless we will show later in the chapter how several existing approaches to the task relate to the framework.

We first give an overview of the task and summarise existing approaches. In Section \ref{glickman-probabilistic-section} we examine the approach of \cite{Glickman:05} who define a probabilistic framework for textual entailment. Then in Section \ref{logical-approaches-section} we look at systems for recognising textual entailment that make use of logical representations of meaning. This is relevant for our discussion in Chapter \ref{model-theoretic-chapter} of how to represent statistical information about uncertainty with logical representations of meaning within the framework.

In Section \ref{context-entailment} we relate the context-theoretic framework to existing approaches to textual entailment by defining some context theories. These not only serve to illustrate the application of the framework but also suggest new modifications to the existing approaches.

\section{The Recognising Textual Entailment Challenge}
\index{entailment!textual!PASCAL Challenge|(}

\begin{table}
\begin{center}
\begin{tabular}{|c|p{5cm}|p{5cm}|c|}
\hline
Task & Text & Hypothesis & Ent.\\
\hline \hline
IE &
A bus collision with a truck in Uganda has resulted in at least 30 fatalities and has left a further 21 injured. &
30 die in a bus collision in Uganda. &
Yes\\
\hline
IR &
Chirac needed a new mandate for his government from the electorate, or a new left government was needed that could count on the support of the trade union bureaucracy and among the working class and so would encounter less resistance.&
Parliamentary elections create new government in France.&
No\\
\hline
QA &
Brazilian cardinal Dom Eusbio Oscar Scheid, Archbishop of Rio de Janeiro , harshly criticized Brazilian President Luiz In‡cio Lula da Silva after arriving in Rome on Tuesday.&
The Brazilian President is Luiz In‡cio Lula da Silva. &
Yes\\
\hline
SUM &
The mine would operate nonstop seven days a week and use tons of cyanide each day to leach the gold from crushed ore. &
A weak cyanide solution is poured over it to pull the gold from the rock.&
No\\
\hline
\end{tabular}
\caption{Sample text and hypothesis sentences from the Third Recognising Textual Entailment Challenge and whether entailment is judged to hold between them, with examples from the sub-tasks of information extraction (IE), information retrieval (IR), question answering (QA) and summarisation (SUM).}
\end{center}
\label{entailment-examples}
\end{table}

The task of recognising textual entailment has reached prominence recently with the launch of the Recognising Textual Entailment Challenge \citep{Dagan:05, Bar-Haim:06}, in which participants develop systems to analyse pairs of sentences to automatically determine whether entailment exists. An example from the development set of the third challenge is the pair
\begin{enumerate}
\item UK Foreign Secretary Jack Straw said Iraqis had ``shown again their determination to defy the terrorists and take part in the democratic process''.
\item Jack Straw holds the position of UK Foreign Secretary.
\end{enumerate}
In this case entailment does hold, since we can deduce the content of the second sentence (called the \emph{hypothesis}) from the first (called the \emph{text}); see Table \ref{entailment-examples} for more examples.

\begin{table}
\begin{center}
\begin{tabular}{|l||c|c|c|c|c|c|}
\hline
Challenge & \swl{Corpus / web-based statistics} & \swl{Lexical relation DB} & \swl{Syntactic matching} & \swl{World knowledge / Paraphrase templates} & \swl{Logical inference} & \swl{Total number of submissions}\\
\hline \hline
RTE-1 & 13 & 10 & 13 & 3 & 7 & 28\\
\hline
RTE-2 & 22 & 32 & 28 & 5 & 2 & 41\\
\hline\hline
Total (\%) & 51\% & 61\% & 59\% & 12\% & 13\% & 100\%\\
\hline
\end{tabular}
\caption{Number of submitted runs using various techniques in Recognising Textual Entailment Challenges 1 and 2 (RTE-1 and RTE-2 respectively).}
\label{RTE-techniques}
\end{center}
\end{table}

It is immediately clear from the generality of this task that a wide range of language processing tools and resources are required to tackle this problem comprehensively; this is demonstrated in the approaches that have been attempted in the two Recognising Textual Entailment Challenges (see Table \ref{RTE-techniques}):
\begin{itemize}
\item \textbf{Morphological and syntactic analysis:} various levels of analysis have been performed in tackling this task, ranging from simple stemming of words to full dependency parsing of sentences. 59\% of runs submitted to both challenges used some kind of syntactic analysis.
\item \textbf{Lexical semantic knowledge:} An even greater proportion of runs submitted, 61\% made use of a lexical relations database such as WordNet, while 51\% of runs used corpus or web-based statistics such as measures of distributional similarity. Indeed it seems that the major focus of approaches to the task to date has been on analysis at the lexical level, while deep semantic analysis has received far less attention.
\item \textbf{Inference and world knowledge:} Only 13\% of submitted runs in both challenges, and only 2 runs in the second challenge used some kind of logical inference. This could be because of the complexity of implementing the task, with no choice but to deal with problems such as anaphora resolution and lexical ambiguity. However, we believe that deeper semantic analysis is necessary to achieve high accuracy in this task: accuracy is low in current approaches, with no team achieving greater than 75\%. Deep semantic analysis and use of world knowledge are areas that have not been explored fully; indeed, we will argue that current systems have not achieved their full potential because of failure to deal effectively with the ambiguity and uncertainty inherent in analysis of natural language.
\end{itemize}

It does seem that in order to perform well at this task it is necessary to combine various tools and techniques: the two best performing entries to the second entailment challenge improved the accuracy of their systems in this way. One of these \citep{Hickl:06} achieved 75\% accuracy using a system that essentially treated the task as a classification problem; in doing so however, they made use of a part-of-speech tagger, parser, named entity recogniser, semantic parser for determining dependency relations, a lexical alignment system, and a method of acquiring paraphrases from the world-wide web. It is not clear however whether all these components are essential to achieving this level of accuracy in their system. \citeauthor{Tatu:06}'s (\citeyear{Tatu:06}) system achieves 74\% accuracy by combining a simple lexical alignment method with a deep semantic analysis using world knowledge and inference.

Despite the benefits of combining several techniques, part of the attraction of the task is that very simple methods perform relatively well. For example, in the second Recognising Textual Entailment Challenge, \cite{Zanzotto:06} achieve 60\% accuracy (which is as good as the best entries in the first challenge) purely by measuring lexical overlap\index{lexical overlap} between the text and hypothesis sentences.



\subsection{Discussion}

We believe it is vital when implementing a system that it is based within a framework with a firm theoretical foundation. The framework provides guidance at each stage of construction of the system and ensures that decisions that are taken in implementing it are made in a consistent, logical manner.

It is also important for us that the theoretical foundation of a textual entailment system be \emph{linguistic} in nature; that is, the framework provides guidance as to how to deal with language. Conversely, many approaches to the task of recognising textual entailment make use of a framework which requires abstraction of the task to a level where language is irrelevant --- an example of this is systems such as that of \citet{Hickl:06} which treat the problem as one of classification of pairs of sentences. Whilst their approach is successful, it provides no insight into the linguistic nature of textual entailment because of the framework in which it is based; instead it provides engineering insight about the task itself and approaches to the task.

We also believe that such a framework should be grounded in the mathematics of probability. Statistical approaches to dealing with language have proved successful in dealing with syntax and, to a degree, lexical semantics because of the possibility of gaining wide coverage by using large amounts of data, and providing robustness, for example by allowing the representation of uncertainty of the correctness of a parse. Thus it is only natural that a framework for textual entailment should also allow for such representation of uncertainty, and the mathematics of probability is the most established and well-founded way of doing this.

In the following sections, we will describe some of the approaches to this task, concentrating specifically on those that we believe can benefit most from the context-theoretic approach.

\subsection{\citeauthor{Glickman:05}'s Probabilistic Setting}
\label{glickman-probabilistic-section}

The work of \citet{Glickman:05}\index{Glickman and Dagan!textual entailment framework} is of great interest to us because they describe a probabilistic framework which is also linguistic in nature, and deals specifically with the nature of textual entailment. As we will see however, in our opinion their framework is not ideal and leaves areas upon which our context-theoretic framework can improve.

Their framework is defined as follows: let $T$ denote the set of possible texts and $H$ the set of possible hypotheses. The set $W$ denotes the set of all mappings from $H$ to $\{0,1\}$; it is called the set of \emph{possible worlds}, and each element of $W$ is interpreted as assigning truth values of true (1) or false (0) to elements of $H$.

The authors describe their setting as follows:
\begin{quote}
We assume a probabilistic generative model for texts and possible worlds. In particular, we assume that texts are generated along with a concrete state of affairs represented by a possible world. Thus, whenever the source generates a text $t\in T$, it generates also corresponding hidden truth assignments that constitute a possible world $w\in W$.

The probability distribution of the source, over all possible texts and truth assignments $T\times W$, is assumed to reflect inferences that are based on the generated texts. That is, we assume that the distribution of truth assignments is not bound to reflect the state of affairs in a particular ``real'' world, but only the inferences about the proposition's truth which are related to the text.
\end{quote}

The term ``possible world'' relates to assignments of truth values to elements of $H$. Thus the authors' setting states that for each text $t$, there is some conditional probability distribution $P(\text{Tr}_h = 1| t)$ over truth assignments to hypotheses; $P(\text{Tr}_h = 1| t)$ denotes the probability that hypothesis $h$ is true given that the text $t$ has been generated. The authors consider entailment to exist between $t$ and $h$ if this probability is greater than the prior probability of $h$ being true, $P(\text{Tr}_h = 1)$.

\subsection{Lexical Entailment Model}

\citeauthor{Glickman:05} apply their probabilistic setting using a simple model of entailment based on occurrences of words in web documents.\index{Glickman and Dagan!lexical entailment model} In order to do this, they allow individual words to be assigned truth values; they suggest a possible interpretation for this as the existence of a concept related to the word, so that $\text{Tr}_\text{book} = 1$ when text $t$ is generated if ``it can be inferred in $t$'s state of affairs that a book exists''. A hypothesis is considered to be true if all its component words are true; in addition it is assumed that the probabilities of individual terms being true in a hypothesis are independent of each other:
\begin{gather*}
P(\text{Tr}_h = 1) = \prod_{u\in h}P(\text{Tr}_u=1)\\
P(\text{Tr}_h = 1|t) = \prod_{u\in h}P(\text{Tr}_u=1|t)
\end{gather*}
where the product is over all words $u$ in $h$, considered as a bag or multiset of words. In order to estimate $P(\text{Tr}_u=1|t)$ for a given word $u$ in the hypothesis, they assume that ``the majority of the probability mass comes from a specific entailing word in $t$:
$$P(\text{Tr}_u = 1 | t) = \max_{v\in t} P(\text{Tr}_u = 1|T_v)$$
where $T_v$ denotes the the event that a generated text contains the word $v$.'' Finally, they make the assumption that ``all hypotheses stated verbatim in a document are true and all others are false and hence $P(\text{Tr}_u=1|T_v) = P(T_u|T_v)$''. That is, the probability of a hypothesis (word) $u$ being true given that a document containing the word $v$ is generated is just the probability that a document contains word $u$ given that it contains word $v$. These values can easily be estimated based on frequency counts:
$$P(T_u|T_v) \simeq \frac{n_{u,v}}{n_v}$$
where $n_{u,v}$ is the number of documents containing both $u$ and $v$, and $n_v$ is the number of documents containing $v$.

For the first Recognising Textual Entailment Challenge, the authors used estimates of these probabilities based on frequency counts from web search engines; their system achieved 59\% accuracy, one of the best scores achieved in the first challenge.



\subsection{Analysis of \citeauthor{Glickman:05}'s Approach}

\citeauthor{Glickman:05}'s framework aims to achieve the same as we wish to achieve, namely, the incorporation of the representation of uncertainty into logical reasoning. This includes the representation of all kinds of uncertainty involved in recognising textual entailment, as they state:
\begin{quote}
An implemented model that corresponds to our probabilistic setting is expected to produce an estimate for $P(\text{Tr}_h=1|t)$. This estimate is expected to reflect all probabilistic aspects involved in the modelling, including inherent uncertainty of the entailment inference itself\ldots, possible uncertainty regarding the correct disambiguation of the text\ldots, as well as probabilistic estimates that stem from the particular model structure.
\end{quote}

Their framework requires combining knowledge about generation of text with reasoning about the probability of the \emph{truth} of propositions. An example they give that illustrates this is the sentence ``His father was born in Italy'', which (one would imagine) should entail with high probability the sentence ``He was born in Italy''. However, according to the authors, examining the texts containing the sentence ``His father was born in Italy'', we find that in these texts the son was more often \emph{not} born in Italy (presumably because the father of someone born in Italy is likely to also be born in Italy, meaning that the sentence is unlikely to be used when this is the case). Hence, in their framework, the probability of entailment would be low, since the probability of the second sentence being true, given the generation of the first sentence, is low.

From our perspective, there are several problems with their framework:
\begin{itemize}
\item The framework requires the hypothesis to be interpretable as a logical proposition. Many textual entailment implementations do not make use of logical representations, however, including the authors' own implementation. This forces them to allow truth values to be assigned to words, which is not ideal since there is no satisfactory interpretation of what it means for a word to be ``true'' since words do not in general refer to propositions; in order to allow words to be interpreted as propositions we need further assumptions.
\item Because of the limitation just mentioned, their framework does not make predictions about the entailment of phrases or words, only sentences.
\item The combination of the probability of truth of propositions with generation of text is confusing. For someone implementing a textual entailment recognition system within their framework, it is unclear how these probabilities are to be obtained. In the authors' implementation, they assume that ``all hypotheses stated verbatim in a document are true and all others are false'', resulting in a system that is much closer in nature to our context-theoretic framework. 
\end{itemize}

Glickman and Dagan seem to be the only entrants to the challenge that attempt to define a framework specifically for textual entailment. Also relevant to our framework however are those entries to the challenge that make use of logical representations of meaning, since we will do this within our framework in Chapter \ref{model-theoretic-chapter} with the purpose of handling statistical information about uncertainty and ambiguity; thus such systems are the subject of the next section.


\subsection{Logical Approaches}
\label{logical-approaches-section}
\index{logical semantics!and textual entailment|(}

\begin{table}
\begin{tabular}{|p{2.4cm}|p{2.4cm}|p{4.3cm}|p{2.3cm}|p{2.1cm}|}%
\hline%
Author(s) & Parser & Inference technique & {\raggedright Inference\\ system(s)} & {\raggedright Accuracy\\(Coverage)}\\ %
\hline\hline %
{\raggedright \citet{Akhmatova:05}\\} & Link Parser & {\raggedright Theorem proving to detect entailment between atomic propositions\\} & OTTER & 52\%\\ %
\hline
{\raggedright \citet{Bayer:05}\\} & {\raggedright Link Parser, MITRE dependency analyser\\} & {\raggedright Probabilistic inference\\} &{\raggedright EPILOG} & 52\% (73\%)\\
\hline %
{\raggedright \citet{Delmonte:05}\\} & {\raggedright VENSES} & {\raggedright Score based comparison of semantic representations\\} &{\raggedright VENSES} & 61\% (62\%)\\ \hline
{\raggedright \citet{Fowler:05}\\} & {\raggedright Unknown} & {\raggedright Theorem proving with scores for dropped predicates / relaxed arguments\\} & {\raggedright COGEX (based on OTTER)\\} & 55\%\\ \hline
{\raggedright \citet{Raina:05}\\} & {\raggedright \citet{Klein:03}\\} &{\raggedright Abductive theorem proving with costs, classifier\\} &{\raggedright EPILOG} & 56\%\\ \hline \hline
{\raggedright \citet{Bos:06}\\} & {\raggedright CCG-parser \citep{Bos:05}\\} &{\raggedright Theorem proving and model building, decision tree\\} &{\raggedright Vampire, Paradox, Mace} & 61\%\\ \hline
{\raggedright \citet{Tatu:06}\\} & MINIPAR &{\raggedright Theorem proving with scores for dropped predicates / relaxed arguments, lexical alignment, classifier\\} &{\raggedright COGEX} & 74\%\\ \hline
\end{tabular}
\caption{Summary of logical approaches to textual entailment. Coverage indicates the proportion of pairs for which the system returned answers, if not 100\%.}
\label{logical-approaches}
\end{table}

Textual entailment recognition systems that make use of logic rarely take the straightforward approach of translating a sentence into a logical form and seeing if the representation of the text logically entails the representation of the hypothesis; in fact no entry to either challenge took such an approach, while those that came closest to doing so \citep{Akhmatova:05, Delmonte:05} are not robust enough to perform well at the task. It seems logical representations are inherently brittle and on their own are not suited to the flexibility of reasoning that is required to deal with natural language. To get around this problem, several strategies have been employed (see also Table \ref{logical-approaches}):
\begin{itemize}
\item \textbf{Score / cost based systems:}\index{scoring} In logical representations a single proposition in the hypothesis that is not in the text will prevent entailment from holding; this is an example of the brittleness of logical representations. Score based systems \citep{Delmonte:05, Fowler:05, Raina:05, Tatu:06} address this, by relaxing the conditions on entailment holding, but adding a ``cost'' or score to such relaxations (for example the addition of an extra proposition to the hypothesis) to indicate a lack of certainty about entailment holding.

This effectively allows ``degrees'' of entailment, where the degree is determined by the score (although this interpretation is not given by the authors). It is practically useful, since it allows more flexibility in determining the existence of entailment, however it is theoretically unfounded, and thus questions remain as to exactly how the scores are to be assigned, what values they are to take, and how they can be interpreted.

\item \textbf{Classification based systems:}\index{classification task} Another approach \citep{Raina:05, Tatu:06} often used in combination with scoring is to treat detection of entailment between pairs of sentences as a classification problem: the task is to classify such pairs as either showing entailment or not. The results of logical inference would then be considered as one ``feature'' of the pair, which together with other features, (for example word overlap), would form the input to a classifying system. The parameters of the classifier are then determined by training on the development set of pairs.

This approach is useful since it allows different techniques to be combined by describing them as features; the weaknesses of one technique can be compensated for by strengths of another. For example, measuring word overlap is a robust technique, but not terribly accurate, so in cases where logical analysis fails, word overlap provides a good back-up measure.

The problem with the clasification approach is that it is not tackling the problem at its source, merely compensating for failures in each technique with other, also imperfect techniques. Instead of trying to understand the nature of entailment, this is a useful way of engineering systems to do the best with the techniques at hand.

Ultimately, it seems hard to imagine such a system performing extremely well, if each of the component ``features'' involved are really flawed measures of entailment. It would always be possible to think of example pairs of sentences which fall outside the range of the development set and thus exploit flaws in each component system.

In addition, it seems unlikely that such analyses will bring us closer to understanding the nature of language or textual entailment itself, instead it will merely provide us with insight as to how best to approach the \emph{task} of recognising textual entailment using existing techniques.
\item \textbf{Model building:} Another interesting approach \citep{Bos:06} is to build models of $T$ and $T\land H$, where $T$ and $H$ are the logical representations of the text and hypothesis, if they are satisfiable, and compare the sizes of the models built. If $T\land H$ has a model that is not much larger than $T$ then it is reasonable to assume that a lot of the information in the hypothesis is also contained in the text, and thus that the text entails the hypothesis; the result is again a relaxing of the conditions for entailment allowing degrees of entailment based on the comparison of model sizes.

However this approach lacks a firm theoretical foundation, and thus questions such as how to best measure model size are unanswered --- for example, a measure that one might use is domain size, which measures the number of entities in the model. In addition to this \citeauthor{Bos:06} use the product of the domain size and the number of all instances of relations in the model as a measure of model size.
\item \textbf{Probabilistic reasoning:} The EPILOG system used by \cite{Bayer:05} allows reasoning about probabilistic statements such as ``If $x$ is a person, then with probability $\ge$ 0.95, $x$ lives in a building.'' \citep{Kaplan:00}. It is not clear however, if and how \citeauthor{Bayer:05}~incorporate this feature into their system; moreover their system performs poorly in terms of robustness and accuracy.
\end{itemize}


\index{logical semantics!and textual entailment|)}
\index{entailment!textual!PASCAL Challenge|)}

\section{Context Theories for Textual Entailment}
\label{context-entailment}
\index{context theory}

The only existing framework for textual entailment that we are aware of is that of \citet{Glickman:05}.\index{Glickman and Dagan!textual entailment framework} However this framework does not seem to be general enough to deal satisfactorily with many techniques used to tackle the problem since it requires interpreting the hypothesis as a logical statement.

Conversely, systems that use logical representations of language are often implemented without reference to any framework, and thus deal with the problems of representing the ambiguity and uncertainty that is inherent in handling natural language in an ad-hoc fashion.

Thus it seems what is needed is a framework which is general enough to satisfactorily incorporate purely statistical techniques and logical representations, and in addition provide guidance as to how to deal with ambiguity and uncertainty in natural language. It is this that we hope our context-theoretic framework will provide.

In this section we analyse approaches to the textual entailment problem, showing how they can be related to the context-theoretic framework, and discussing potential new approaches that are suggested by looking at them within the framework. We first discuss some simple approaches to textual entailment based on subsequence matching and measuring lexical overlap. We then look at how Glickman and Dagan's approach can be considered as a context theory in which words are represented as projections on the vector space of documents. This leads us to an implementation of our own in which we used latent Dirichlet allocation as an alternative approach to overcoming the problem of data sparseness.

\subsection{Subsequence Matching and Lexical Overlap}
\label{subsequence}
\index{subsequence matching}

We call a sequence $x \in A^*$ a ``subsequence'' of $y \in A^*$ if each element of $x$ occurs in $y$ in the same order, but with the possibility of other elements occurring in between, so for example $abba$ is a subsequence of $acabcba$ in $\{a,b,c\}^*$. We denote the set of subsequences of $x$ (including the empty string) by $\Sub(x)$. Subsequence matching compares the subsequences of two sequences: the more subsequences they have in common the more similar they are assumed to be. This idea has been used successfully in text classification \citep{Lodhi:02} and also formed the basis of the author's entry to the second Recognising Textual Entailment Challenge \citep{Clarke:06}.

\begin{table*}
\begin{center}
\begin{tabular}{|l||c|c|c|c|c|c||c|}
\hline
Run & Acc. & Av. Prec. & IE & IR & QA & SUM & Dev. Acc\\
\hline
Word Matching  & 0.53 & 0.56 & 0.46 & 0.49 & 0.59 & 0.59 & 0.57 \\
Subsequence Matching  & 0.55 & 0.53 & 0.49 & 0.56 & 0.54 & 0.60 & 0.57 \\
Substr. + Corpus Occ. & 0.53 & 0.53 & 0.50 & 0.50 & 0.53 & 0.59 & 0.55\\
\hline
\end{tabular}
\caption{Results for for the author's entry to the Second Recognising Textual Entailment Challenge using subsequence matching. Estimates of overall accuracy and average precision, and accuracy results for each of the subtasks (Information Extraction, Information Retrieval, Question Answering and Summarisation) are shown for the test set, together with accuracy for the development set. Results are reported for a baseline of simple word matching and the two entered runs: subsequence matching and subsequences combined with corpus occurrences. Error for accuracy values on the test and development sets is approximately 2\%, and on the subtasks, 4\%.}
\label{table:subsequence}
\end{center}
\end{table*}

If $S$ is a semigroup, $L^1(S)$ is a lattice ordered algebra under the multiplication of convolution:
$$(f\cdot g)(x) = \sum_{yz = x} f(y)g(z)$$
where $x,y,z \in S$, $f,g \in L^1(S)$. For a sequence $x\in A^*$, we define $\hat{x} \in L^1(A^*)$ by
$$\hat{x} = (1/2^{|x|})\sum_{y\in Sub(x)} e_y,$$
where $e_y$ is the unit basis element associated with $y$; that is, the function that is $1$ on $y$ and $0$ elsewhere. This is clearly a semigroup homomorphism and thus together with the linear functional $\phi$,
$$\phi(u) = \|u^+\|_1 - \|u^-\|_1$$
defines a context theory\index{context theory} for $A$. Under this context theory, a sequence $x$ completely entails $y$ if and only if it is a subsequence of $y$. In our experiments \citep{Clarke:06}, we have shown that this type of context theory can perform significantly better than straightforward lexical overlap (see Table \ref{table:subsequence}). Many variations on this idea are possible, for example using more complex mappings from $A^*$ to $L^1(A^*)$.


The simplest approach to textual entailment is to measure the degree of lexical overlap:\index{lexical overlap} the proportion of words in the hypothesis sentence that are contained in the text sentence. Though simple, variations on this approach can perform  comparably to much more complex techniques \citep{Dagan:05}.

This approach can be described as a context theory in terms of a free commutative semigroup on a set $A$, defined by $A^*/\equiv$ where $x \equiv y$ in $A^*$ if the symbols making up $x$ can be reordered to make $y$. Following the reasoning of subsequence matching, for a sequence $x$ we can define $\hat{x} \in L^1(A^*/\equiv)$ by
$$\hat{x} = (1/2^{|x|})\sum_{y\in Sub(x)} e_{[y]},$$
where $[y]$ is the equivalence class of $y$ in $A^*/\equiv$. Defining a linear functional similarly gives us a context theory\index{context theory} in which entailment depends on the words in the sequences but not their order. Again, more complex definitions of $\hat{x}$ can be used, for example to weight different words by their probabilities.

\subsection{Document Projections}
\label{document-projections}
\index{Glickman and Dagan!lexical entailment model|(}

\cite{Glickman:05} give a probabilistic definition of entailment in terms of ``possible worlds'' which they use to justify their lexical entailment model based on occurrences of words in web documents. They estimate the lexical entailment probability $\text{\textsc{lep}}(u,v)$ to be
$$\text{\textsc{lep}}(u,v) \simeq \frac{n_{u,v}}{n_v}$$
where $n_v$ and $n_{u,v}$ denote the number of documents that the word $v$ occurs in and the words $u$ and $v$ both occur in respectively. From the context theoretic perspective, we view the set of documents the word occurs in as its context vector. To describe this situation in terms of a context theory, consider the vector space $L^\infty(D)$ where $D$ is the set of documents. With each word $u$ we associate an operator $P_u$ on this vector space by
$$P_u e_d = \left\{\begin{array}{ll} e_d & \text{if $u$ occurs in document $d$} \\ 0 & \text{otherwise.} \end{array}\right.$$
where $e_d$ is the basis element associated with document $d \in D$. $P_u$ is a projection, that is $P_uP_u = P_u$; it projects onto the space of documents that $u$ occurs in. These projections are clearly commutative: $P_uP_v = P_vP_u = P_u \land P_v$ projects onto the space of documents in which both $u$ and $v$ occur.

In their paper, \citeauthor{Glickman:05} assume that probabilities can be attached to individual words, as we do, although they interpret these as the probability that a word is ``true'' in a possible world. In their interpretation, a document corresponds to a possible world, and a word is true in that world if it occurs in the document.

They do not, however, determine these probabilities directly; instead they make assumptions about how the entailment probability of a sentence depends on lexical entailment probability. Although they do not state this, the reason for this is presumably data sparseness: they assume that a sentence is true if all its lexical components are true: this will only happen if all the words occur in the same document. For any sizeable sentence this is extremely unlikely, hence their alternative approach.

It is nevertheless useful to consider this idea from a context theoretic perspective. The probability of a term being true can be estimated as the proportion of documents it occurs in. This is the same as the context theoretic probability defined by the linear functional $\phi$, which we may think of as determined by a vector $p$ in $L^\infty(D)$ given by $p(d) = 1/|D|$ for all $d \in D$. In general, for an operator $U$ on $L^\infty(D)$ the context theoretic probability of $U$ is defined as
$$\phi(U) = \|U^+p\|_1 - \|U^-p\|_1,$$
where by $U^+$ and $U^-$ we mean the positive and negative parts of $U$ in the vector lattice of operators (see Section \ref{positive-operators}). 
The probability of a term is then $\phi(P_u) = n_u /|D|$. More generally, the context theoretic representation of an expression $x = u_1u_2\ldots u_m$ is $P_x = P_{u_1}P_{u_2}\ldots P_{u_m}$. This is clearly a semigroup homomorphism (the representation of $xy$ is the product of the representations of $x$ and $y$), and thus together with the linear functional $\phi$ defines a context theory\index{context theory} for the set of words.

The degree to which $x$ entails $y$ is then given by $\phi(P_x\land P_y) / \phi(P_x)$. This corresponds directly to \citeauthor{Glickman:05}'s entailment ``confidence'' without the additional assumptions they make; it is simply the proportion of documents that contain all the terms of $x$ which also contain all the terms of $y$.

\index{Glickman and Dagan!lexical entailment model|)}

\subsection{Latent Dirichlet Projections}

This formulation suggests an alternative approach to that of \citeauthor{Glickman:05} to cope with the data sparseness problem. We consider the finite data available $D$ as a sample from a corpus model $D'$; the vector $p$ then becomes a probability distribution over the documents in $D'$. In our own experiments, we used Latent Dirichlet Allocation (see Section \ref{lda-section}) to build a corpus model based on a subset of around 380,000 documents from the Gigaword corpus. Having the corpus model allows us to consider an infinite array of possible documents, and thus we can use our context-theoretic definition of entailment since there is no problem of data sparseness.

Consider the vector space $L^\infty(A^*)$ for some alphabet $A$, the space of all bounded functions on possible documents. In this approach, we define the representation of a string $x$ to be a projection $P_x$ on the subspace representing the (infinite) set of documents in which all the words in string $x$ occur. Again we define a vector $q(d)$ for $d\in A^*$ where $q(d)$ is the probability of document $d$ in the corpus model, we then define a linear functional $\phi$ for an operator $U$ on $L^\infty(A^*)$ as before by $\phi(U) = \|U^+q\|_1 - \|U^-q\|_1$. $\phi(P_x)$ is thus the probability that a document chosen at random contains all the words that occur in string $x$. In order to estimate $\phi(P_x)$ we have to integrate over the Dirichlet parameter $\theta$:
$$\phi(P_x) = \int_\theta\left(\prod_{a\in x}p_\theta(a)\right)p(\theta)d\theta,$$
where by $a\in x$ we mean that the word $a$ occurs in string $x$, and $p_\theta(a)$ is the probability of observing word $a$ in a document generated by the parameter $\theta$. We estimate this by
$$p_\theta(a) \simeq 1 - \left(1 - \sum_z p(a|z)p(z|\theta)\right)^N,$$
where $z$ is the topic variable described in Section \ref{lda-section} and we have assumed a fixed document length $N$. The above formula is an estimate of the probability of a word occurring at least once in a document of length $N$, i.e.~$1 -$ the probability that the word does not occur $N$ times. The sum over the topic variable $z$ is the probability that the word $a$ occurs at any one point in a document given the parameter $\theta$. We approximated the integral using Monte-Carlo sampling to generate values of $\theta$ according to the Dirichlet distribution.

\begin{table}
\begin{center}
\begin{tabular}{|l||l|l|}
\hline
\textbf{Model} & \textbf{Accuracy} & \textbf{CWS}\\
\hline\hline
Dirichlet ($10^6$) & 0.584 & 0.630\\
Dirichlet ($10^7$) & 0.576 & 0.642\\
\hline
Bayer (MITRE) & 0.586 & 0.617 \\
Glickman (Bar Ilan) & 0.586 & 0.572\\
Jijkoun (Amsterdam) & 0.552 & 0.559\\
Newman (Dublin) & 0.565 & 0.6\\
\hline
\end{tabular}
\caption{Results obtained with our Latent Dirichlet projection model on the data from the first Recognising Textual Entailment Challenge for two document lengths $N = 10^6$ and $N = 10^7$ using a cut-off for the degree of entailment of $0.5$ at which entailment was regarded as holding. CWS is the confidence weighted score --- see \citep{Dagan:05} for the definition.}
\label{table:lda-results}
\end{center}
\end{table}

The results we obtained using this method on the data from the first Recognising Textual Entailment Challenge were comparable to the best results in the first challenge (see Table \ref{table:lda-results}).

\subsection{Discussion}

We have shown how some existing approaches to the task of recognising textual entailment can be described in terms of context theories. The potential for extending these ideas is great:
\begin{itemize}
\item Context theories based on substring matching can be extended
\begin{itemize}
\item by using different weighting schemes for strings based on the length of the string or the probability of its occurrence in large corpora;
\item by replacing words with vectors representing their context; instead of using concatenation of words to form representations of the substring use the tensor product of the vectors;
\item by allowing partial commutativity --- a hybrid of lexical and substring matching could be made by allowing some words to commute and not others; this could be based on an analysis of the relative frequency of pairs of words in a corpus.
\end{itemize}
\item Glickman and Dagan's approach can be extended by considering other corpus models --- there are many possible alternatives to latent Dirichlet allocation, for example using $n$-grams or other models in which words do not commute.
\item The evidence from entries to the challenge suggest that to perform well at the task a number of different approaches need to be combined. The context theoretic framework makes it easy to do this while remaining true to the framework. For example, given two context theories that map a string $x$ to vectors $\hat{x}_1$ and $\hat{x}_2$ with linear functionals $\phi_1$ and $\phi_2$, a new context theory can be defined in terms of the direct sum of the vectors so that $x$ maps to $\hat{x}_1 \oplus \hat{x}_2$, with linear functional $\phi(\hat{x}_1 \oplus \hat{x}_2) = \alpha\phi_1(\hat{x}_1) + \beta\phi_2(\hat{x}_2)$, with $\alpha$ and $\beta$ positive real numbers such that $\alpha + \beta = 1$. This ``weighted sum'' could be used to combine any number of context theories possibly describing quite different approaches.
\end{itemize}

While the simple techniques of this chapter are useful, to perform well at the task of textual entailment some form of in-depth reasoning seems essential because of the semantic nature of the task. Earlier we discussed approaches to the challenge that make use of logical representations of language. Because such representations on their own lack robustness it is clear that ways need to be found to extend them. One way of doing this would be to use a weighted sum to combine a context theory that describes the logical interpretation of a sentence with a context theory describing a more robust technique such as lexical matching. Such an approach would help increase robustness, however it would not get to the root of the problem, which we believe to lie in the lack of flexibility of the logical representations themselves.

 In the next chapter we will show how logical approaches can be described in terms of context theories. The context-theoretic approach suggests ways in which statistical information about uncertainty can be incorporated into such representations, allowing us to represent a sentence as a weighted sum over many possible logical interpretations of a sentence to take into account statistical information from a parser, word sense disambiguation systems, anaphora resolution and so on. We hope that this will ultimately lead to entailment systems that are able to reason with logic in a robust and principled manner.






\chapter{Uncertainty in Logical Semantics}
\label{model-theoretic-chapter}
\index{logical semantics|(}

The standard approach to representing meaning in natural language is to represent sentences of the language by some logical form. This is useful in situations where it is necessary to perform in-depth reasoning, however it brings with it many problems. Such systems require accurate parses of sentences in order to reason effectively, yet existing parsers\index{parsers} do not provide sufficient accuracy or coverage. Parsers will typically return a probability distribution over possible parses, however systems using logical reasoning do not make use of these probabilities. Similarly, these systems typically need to know which sense of a word was intended by the speaker in a particular context; the task of word sense disambiguation\index{word sense disambiguation} attempts to determine this, however current performance at this task is poor. Whilst there are many problems encountered by such systems, these are two that we will look at in this chapter; it is our hope however that it will be possible to generalise the ideas presented here to deal with other problems in a similar way.

It is possible that these problems are inherent in the nature of language: perhaps in general there is not one correct parse for a sentence, nor is there only ever one sense of a word that can apply in a particular context. In this case it is vital that representations of the meaning of natural language can incorporate such uncertainty. Alternatively, it may be that these problems will be eventually be solved satisfactorily; in the meantime we need a way to deal with the uncertainty that results from using these techniques.

To our knowledge, existing methods of representing uncertainty\index{uncertainty, representing} and ambiguity\index{ambiguity} are founded in formal semantics and do not incorporate statistical information such as the probability of a parse or the probability of the sense of a word. It is vital to make use of this information when dealing with natural language because there are so many possible sources for uncertainty. Many of the tools that are used in reaching the logical representation (part of speech taggers, parsers, word-sense disambiguation systems etc.) can provide us with statistical information about the uncertainty in a representation, thus it makes sense to incorporate this information into the representation itself. In this chapter we show how the context-theoretic framework can be used to do this, first giving a theoretical analysis of the problem, and then outlining how the ideas can be implemented practically.


The approach we take in this chapter is to first formulate logical semantics within the context-theoretic framework; this gives us the flexibility of vector spaces that we need to represent statistical information about uncertainty. For example, this will allow us to represent the meaning of a word as a weighted sum of its individual senses.

Instead of dealing with a specific version of model-theoretic semantics, we give a very general treatment that can deal with any system in which strings are translated into a logical form with an implication relation defined on it; thus the ideas presented here can be applied to just about any conceivable logical system.

The contributions of this chapter are as follows:
\begin{itemize}
\item In section \ref{logical-projections} we show how logical semantics can be interpreted in a context-theoretic manner: given a way of translating natural language expressions into logical forms, we can define an algebra which represents the meaning equivalently. Given also a way of attaching probabilities to logical forms (which can be given a Bayesian interpretation), we can define a context theory allowing us to deduce degrees of entailment between expressions.
\item In section \ref{uncertainty-section} we show how the vector-based representation allows statistical information about uncertainty of meaning and ambiguity to be incorporated; the representation of an ambiguous sentence is a weighted sum over the vector representations of its unambiguous meanings. These may be the result of syntactic ambiguity, such as multiple parses returned by a statistical parser, due to ambiguous words or some other source of uncertainty. 
\item Computing with the representations we describe is far from straightforward. In Section \ref{practical-issues} we outline how the ideas we present may be implemented in a concrete manner, showing how a system may be built to compute a degree of entailment between two sentences.
\item Most of this chapter relates to the representation of natural language sentences that are translated into logical form, however to demonstrate the general applicability of the context-theoretic framework, we show how (in Section \ref{context-theoretic-analysis}), given a logical representation of sentences, entailment can be defined between words and phrases, based on a context-theoretic analysis of the situation.
\end{itemize}

\section{From Logical Forms to Algebra}
\label{logical-projections}

Model-theoretic approaches generally deal with a subset of all possible strings, the language under consideration, translating sequences in the language to a logical form, expressed in another, logical language. Relationships between logical forms are expressed by an entailment relation on this logical language.


This section is about the algebraic representation of logical languages. Representing logical languages in terms of an algebra will allow us to incorporate statistical information about language into the representations. For example, if we have multiple parses for a sentence, each with a certain probability, we will be able to represent the meaning of the sentence as a probabilistic sum of the representations of its individual parses.

By a \emph{logical language} we mean a language $\Lambda \subset {A'}^*$ for some alphabet $A'$, together with a relation $\vdash$ on $\Lambda$ that is reflexive and transitive; this relation is interpreted as entailment on the logical language. We will show how each element $u \in \Lambda$ can be associated with a projection on a vector space; it is these projections that define the algebra. Later we will show how this can be related to strings in the natural language $\lambda$ that we are interested in.

For a subset $T$ of a set $S$, we define the projection $P_T$ on $L^\infty(S)$ by 
$$P_T e_s = \left\{
\begin{array}{ll}
e_s & \text{if $s \in T$}\\
0 & \text{otherwise}
\end{array}
 \right.$$
 Where $e_s$ is the basis element of $L^{\infty}(S)$ corresponding to the element $s\in S$.
Given $u \in \Lambda$, define $\left\downarrow_\vdash(u)\right. = \{v : v \vdash u\}$. As a shorthand we write $P_u$ for the projection $P_{\left\downarrow_\vdash(u)\right.}$ on the space $L^\infty(\Lambda)$.

The projection $P_u$ can be thought of as projecting onto the space of logical statements that entail $u$. This is made formal in the following proposition:
\begin{prop}\label{modelth}
$P_u \le P_v$ if and only if $u \vdash v$.
\end{prop}
\begin{proof}
Clearly
\begin{equation}
\tag{$*$}\label{projectionprod}
P_uP_v e_w = \left\{
\begin{array}{ll}
e_w & \text{if $w \vdash u$ and $w \vdash v$}\\
0 & \text{otherwise}
\end{array},
 \right.
 \end{equation}
so if $u \vdash v$ then since $\vdash$ is transitive, if $w \vdash u$ then $w \vdash v$, so we must have $P_uP_v = P_u$. The projections $P_u$ and $P_v$ are commutative, so $P_uP_v = P_u$ if and only if $P_u \le P_v$ \citep{Aliprantis:85}.

Conversely, if $P_uP_v = P_u$ then it must be the case that $w \vdash u$ implies $w \vdash v$ for all $w \in \Lambda$, including $w = u$. Since $\vdash$ is reflexive, we have $u \vdash u$, so $u \vdash v$ which completes the proof.
\end{proof}

To help us understand this representation better, we will show that it is closely connected to the ideal completion\index{ideal completion} of partial orders (see Proposition \ref{ideal-completion-prop}). Define a relation $\equiv$ on $\Lambda$ by $u \equiv v$ if and only if $u \vdash v$ and $v \vdash u$. Clearly $\equiv$ is an equivalence relation; we denote the equivalence class of $u$ by $[u]$. Equivalence classes are then partially ordered by $[u] \le [v]$ if and only if $u \vdash v$. Then note that $\bigcup \down{[u]} = \left\downarrow_\vdash(u)\right.$, thus $P_u$ projects onto the space generated by the basis vectors corresponding to the elements $\bigcup \down{[u]}$, the ideal completion representation of the partially ordered equivalence classes.

What we have shown here is that logical forms can be viewed as projections on a vector space. Since projections are operators on a vector space, they are themselves vectors; viewing logical representations in this way allows us to treat them as vectors, and we have all the flexibility that comes with vector spaces: we can add them, subtract them and multiply them by scalars; since the vector space is also a vector lattice, we also have the lattice operations of meet and join. As we will see in the next section, in some special cases such as that of the propositional calculus, the lattice\index{lattice!and logic} meet and join coincide with logical conjunction and disjunction.

\subsection{Application: Propositional Calculus}
\label{propositional}
\index{propositional calculus|(}

In this section we apply the ideas of the previous section to an important special case: that of the propositional calculus. We choose as our logical language $\Lambda$ the language of a propositional calculus with the usual connectives $\lor$, $\land$ and $\lnot$, the logical constants $\top$ and $\bot$ representing ``true'' and ``false'' respectively, with $u \vdash v$ meaning ``infer $v$ from $u$'', behaving in the usual way. Then:
\begin{align*}
P_{u\land v} &= P_uP_v
	& P_{\lnot u} &= 1 - P_u + P_\bot\\
P_{u\lor v} &= P_u + P_v - P_uP_v
	& P_{\top} &=1
\end{align*}

To see this, note that the equivalence classes of $\vdash$ form a Boolean algebra\index{Boolean algebra} under the partial ordering induced by $\vdash$, with
\begin{align*}
[u\land v] & = [u] \land [v]
 & [u\lor v] & = [u] \lor [v]
  & [\lnot u] & = \lnot[u].
\end{align*}
Note that while the symbols $\land$, $\lor$ and $\lnot$ refer to logical operations on the left hand side, on the right hand side they are the operations of the Boolean algebra of equivalence classes; they are completely determined by the partial ordering associated with $\vdash$.\footnote{In the context of model theory, the Boolean algebra of equivalence classes of sentences of some theory $T$ is called the \emph{Lindenbaum-Tarski} algebra of $T$ \citep{Hinman:05}.}

Since the partial ordering carries over to the ideal completion we must have
\begin{align*}
\downnb{[u\land v]} &= \downnb{[u]} \cap \downnb{[v]}
& \downnb{[u\lor v]} &= \downnb{[u]} \cup \downnb{[v]}
\end{align*}
Since $u \vdash \top$ for all $u\in \Lambda$, it must be the case that $\downnb{[\top]}$ contains all sets in the ideal completion. However the Boolean algebra of subsets in the ideal completion is larger than the Boolean algebra of equivalence classes; the latter is embedded as a Boolean sub-algebra of the former. Specifically, the least element in the completion is the empty set, whereas the least element in the equivalence class is represented as $\downnb{[\bot]}$. Thus negation carries over with respect to this least element:
$$\downnb{[\lnot u]} = (\downnb{[\top]} - \downnb{[u]})\cup \downnb{[\bot]}.$$

We are now in a position to prove the original statements:
\begin{itemize}
\item Since $\downnb{[\top]}$ contains all sets in the completion, $\bigcup\downnb{[\top]} = \downe{\top} = \Lambda$, and $P_\top$ must project onto the whole space, that is $P_\top = 1$.
\item Using the above expression for $\downnb{[u \land v]}$, taking unions of the disjoint sets in the equivalence classes we have $\downe{u\land v} = \downe{u} \cap \downe{v}$. Making use of \eqref{projectionprod} in the proof to Proposition \ref{modelth}, we have $P_{u\land v} = P_uP_v$.
\item In the above expression for $\downnb{[\lnot u]}$, note that $\downnb{[\top]} \supseteq \downnb{[u]} \supseteq{\downnb{[\bot]}}$. This allows us to write, after taking unions and converting to projections, $P_{\lnot u} = 1 - P_u + P_\bot$, since $P_\top = 1$.
\item Finally, we know that $u\lor v \equiv \lnot(\lnot u \land \lnot v)$, and since equivalent elements in $\Lambda$ have the same projections we have
\begin{align*}
P_{u\lor v} &= 1 - (P_{\lnot u \land \lnot v}) + P_\bot\\
		&= 1 - (P_{\lnot u}P_{\lnot v}) + P_\bot\\
		&= 1 - (1 - P_u + P_\bot)(1 - P_v + P_\bot) + P_\bot\\
		&= P_u + P_v - P_u P_v - 2P_\bot + P_\bot P_u + P_\bot P_v\\
		&= P_u + P_v - P_u P_v
\end{align*}
\end{itemize}

It is also worth noting that in terms of the vector lattice operations $\lor$ and $\land$ on the space of operators on $L^\infty(\Lambda)$, we have $P_{u\lor v} = P_u \lor P_v$ and $P_{u\land v} = P_u \land P_v$.

\index{propositional calculus|)}

\section{Representing Uncertainty}
\label{uncertainty-section}
\index{uncertainty, representing|(}

In the context of logical representations of meaning, there are certain properties that we would expect from representations of ambiguity; we give an initial list of these and discuss them here --- there are potentially other features we may wish to incorporate into a more complete analysis at a later stage.

\subsubsection*{Bayesianism}
\index{Bayesianism|(}

We would expect our representation to be tied closely to Bayesian reasoning, since this is the standard approach to reasoning with uncertainty. Bayesianism asserts that the correct calculus for modelling uncertainty is the mathematics of probability theory. A ``probability'' assigned to a logical sentence is then merely taken as an indication of our certainty of the truth of the sentence; it is not intended to be a scientifically measurable quantity in the sense that probabilities are often assumed to be. We would expect to be able to incorporate such probabilities into our system.

\index{Bayesianism|)}

\subsubsection*{Ambiguity and Logic}

When dealing with ambiguity in the context of logical representations, we expect certain relationships between the representation of ambiguity and the logical representations. Specifically, if we have an ambiguous expression with two meanings, we would expect the ambiguous representation to entail the logical disjunction of two expressions. In general, we would not expect the converse, since the two are not equivalent. To see this, for example, consider the sentence $s$ = ``He saw a plant''. We wish to represent the lexical ambiguity in the word ``plant'' which we will consider can either mean an industrial plant or an organism. The two disambiguated meanings roughly correspond to the sentences $s_1$ = ``He saw an industrial plant'' and $s_2$ = ``He saw a plant organism''. We expect that each disambiguated sentence $s_i$ entails the ambiguous sentence $s$, and for the reverse we would expect some degree of entailment to exist. 

\subsubsection*{Statistical Features}

Similarly, we would expect the ambiguous sentence to entail the disambiguated meanings to the degree that we expect the ambiguous word to carry the relevant sense. For example, if ``plant'' is used in the sense of industrial plant 40\% of the time, then we would expect that $s$ entails $s_1$ to degree 0.4.

\subsection{Representing Bayesian Uncertainty}
\label{bayesian-uncertainty-section}
\index{Bayesianism|(}

The projection representation of translations to logical form allows us to associate an algebra (of projections) with the logical language $\Lambda$, however it does not quite give us a context theory. For that, we need a linear functional on the algebra of projections, and we will show how this can be done if we take a Bayesian approach to reasoning.

We need to associate probabilities with (logical) sentences in a way that is compatible with their logical structure. For example if a sentence $s_1$ entails $s_2$ then $s_2$ should be assigned a probability at least as large as that of $s_1$. This can be done using a probability distribution over the sentences of the logical language:



\index{logical language!probabilistic|textbf}
\begin{defn}[Probabilistic Logical Language]
Let $\Lambda$ be a logical language with entailment relation $\vdash$, a probability distribution $p$ over elements of $\Lambda$ and a distinguished element $\bot \in \Lambda$ such that
\begin{itemize}
\item $\bot \vdash u$, and
\item $p(u) = 0$ if $u \vdash \bot$,
\end{itemize}
for all $u \in \Lambda$. We call $\langle \Lambda, \vdash, \bot, p\rangle$ a probabilistic logical language. For an arbitrary subset $X$ of $\Lambda$, define $p(X) = \sum_{u\in X}p(u)$. For $u \in \Lambda$ define $p_\vdash(u) = p(\downe{u})$.
\end{defn}
\begin{prop}
The function $p_\vdash$ defines a probability measure on the lattice defined by $\downarrow_\vdash$. Specifically:
\begin{enumerate}
\item $p_\vdash(\bot) = 0$
\item if $\downe{u} \cap \downe{v} = \downe{\bot}$ then $p(\downe{u} \cup \downe{v}) = p_\vdash(u) + p_\vdash(v)$
\end{enumerate}
\end{prop}
\begin{proof}
\mbox{}
\begin{enumerate}
\item $p_\vdash(\bot) = \sum_{u\vdash\bot}p(u) = 0$.
\item if $\downe{u} \cap \downe{v} = \downe{\bot}$ then $p(\downe{u} \cup \downe{v}) = p_\vdash(u) + p_\vdash(v) - p_\vdash(\bot) =  p_\vdash(u) + p_\vdash(v)$
\end{enumerate}
\end{proof}\noindent
Thus we can think of the function $p_\vdash$ as describing the probability of a logical sentence since it has all the properties of a probability measure with respect to the lattice of equivalent sentences. This means, for example that if $s_1 \vdash s_2$ then $p_\vdash(s_1) \le p_\vdash(s_2)$ as we would expect.

We can now define a linear functional on the algebra of projections:
\begin{defn}
Given a probability distribution $p$ over a logical language $\Lambda$, we define a vector $\hat{p}$ in $L^\infty(\Lambda)$ by
$$\hat{p} = \sum_{u \in \Lambda} p(u)e_u$$
We define a linear functional $\phi$ on  the space of bounded operators on $L^\infty(\Lambda)$  by
$$\phi(F) = \|F_+(\hat{p})\|_1 - \|F_-(\hat{p})\|_1$$
where $F_+$ and $F_-$ are the positive and negative parts of the bounded operator $F$ respectively.
\end{defn}
\begin{prop} If $u \in \Lambda$ for some logical language $\Lambda$ with probability distribution $p$, then
$\phi(P_u) = p_\vdash(u)$
\end{prop}
\begin{proof}
$$\phi(P_u) = \|P_u\hat{p}\|_1 = \sum_{v \in \downe{u}}p(v) = p_\vdash(u)$$
\end{proof}

Using the linear functional we can define a context theory\index{context theory} for logical sentences. Since context theories are defined in terms of an alphabet, we have to define it in terms of a finite set $A$ of symbols with each symbol representing a sentence in $\Lambda$. We associate a bounded operator $\hat{x}$ on the space $L^\infty(\Lambda)$ with each element $x\in A$: we have $\hat{x} = P_u$, where $u$ is the logical sentence corresponding to $x$; thus we have a context theory, although only for a finite subset of sentences of $\Lambda$. In practice this should not be a problem, since we only need to be able to interpret a finite number of sentences at any one time.

\index{Bayesianism|)}

\subsection{Representing Syntactic Ambiguity}

One of the major problems facing engineers of natural language systems is how to deal with syntactic ambiguity. Most modern wide-coverage parsers will return many parses for a single sentence, together with a probability distribution over these parses. How are we to make use of this probability distribution while reasoning with the logical representations of the sentences?

We make the simplifying assumption that different parses of a sentence apply in different contexts, so for each context the sentence can occur in there is exactly one parse that applies. We also assume for now that there is a single interpretation $s_i$ of the sentence $s$ for each possible parse.
Thus we can view the context vector of a sentence as the sum of the context vectors of the individual interpretations of the sentence that are attached to each parse:
$$\hat{s} = \sum_i \hat{s}_i,$$
where $\hat{s}_i$ is the context vector representing interpretation $s_i$. We can interpret the probability given to each parse by the parser as contributing to the context theoretic probability of the corresponding interpretation of the sentence as follows:
$$\phi(\hat{s}_i) = p(s_i)p_\vdash(u_i),$$
where $u_i$ is the logical representation of interpretation $s_i$; i.e.~the probability of the interpretation is the probability of the meaning of the interpretation multiplied by the probability of the parse. This will be the case if we represent a sentence as a weighted sum of its individual interpretations, where the weights are given by the probability of the corresponding parse:
$$\hat{s} = \sum_i p(s_i)P_{u_i}.$$
where $P_{u_i}$ is the projection representing the interpretation $s_i$ of sentence $s$. The probability of the sentence as a whole is thus given by $\phi(\hat{s}) = \sum_i p(s_i)p_\vdash(u_i)$.

Note that because of the vector lattice based framework, we are able to take probabilistic sums of the representations of sentences, and the lattice operations are still well defined, enabling us to calculate the degree of entailment between two sentences represented in this way.

This recipe can of course be applied more generally to deal with other forms of uncertainty: for example, any uncertainty about lexical ambiguity, anaphora resolution, part of speech tagging etc.~can be incorporated into a probabilistic sum of the resulting semantic representations of the different analyses. The situation would be similar to the one above: we would have a set of interpretations of a single sentence, each with a probability and a logical representation; the sentence itself would then be represented as a weighted sum of the vector representation of the individual interpretations, with weights given by the probabilities.

\subsection{A Context Theoretic Analysis of Logical Representations}
\label{context-theoretic-analysis}

The algebraic description of logic given in the previous section is useful for giving us an intuition of how logic can be interpreted geometrically as projections, however it can only deal with descriptions of logic at the sentence level. It would be useful in addition to have a description of the logical representation of language in terms of vectors that also told us how words and phrases should be represented. Such a description would allow us to  examine representations of phrases and sentences and compute entailment between them. In this section, we show how such a description can be constructed. This will allow us to represent a word in terms of a sum of its senses, where the logical behaviour of each individual sense is well defined, and provide us with a deeper insight into the relationship between model-theoretic and context-theoretic descriptions of meaning.

In order to represent words however, we are going to need a more comprehensive representation. There are potentially many ways to do this, for example, we could attempt to construct an algebra in terms of semigroups that contains the properties we are looking for. The approach we will describe, however, is context-theoretic in nature, bearing many similarities to the context vectors of a string defined previously.

The approach we will take is as follows: we first associate with each string a function that maps contexts to vectors representing the logical interpretation of the string in that context. We define an appropriate linear functional for this vector space and show that the representation incorporates the logical structure. We then show that the vector representation can be viewed as originating from a generalisation of a corpus model, demonstrating the context-theoretic nature of the definition.

\index{probabilistic logical translation|textbf}
\begin{defn}[Probabilistic Logical Translation]
A probabilistic logical translation is a tuple $\langle\Lambda,\vdash, \bot, p, \lambda, \mu\rangle$ such that $\langle\Lambda,\vdash, \bot, p\rangle$ is a probabilistic logical language, $\lambda$ is some language and $\mu$ is a function from $\lambda$ to $\Lambda$.
\end{defn}

The language $\lambda$ is intended to represent a natural language, and the function $\mu$ the process of obtaining a logical sentence in $\Lambda$ for each sentence in $\lambda$.

\begin{defn}
\label{x-tilde}
Let $\langle\Lambda,\vdash, \bot, p, \lambda, \mu\rangle$ be a probabilistic logical translation where $\lambda \subseteq A^*$. For $x\in A^*$ we define the function $\tilde{x}$ from $A^*\times A^*$ to $L^1(\Lambda)$ by
$$\tilde{x}(a,b) = \begin{cases}
\sum_{u \in \down{\mu(axb)}} p(u)e_u & \text{if $axb \in \lambda$}\\
0 & \text{otherwise.}
\end{cases}$$
\end{defn}

The function $\tilde{x}$ maps a context $(a,b)$ to a vector representing the sum of all the logical representations that entail the logical translation of $axb$. Note that since $\tilde{x}$ is a function to a vector lattice, it can be viewed itself as a vector lattice, with the vector and lattice operations defined point-wise: for example $(\tilde{x} + \tilde{y})(a,b) = \tilde{x}(a,b) + \tilde{y}(a,b)$, $(\alpha \tilde{x})(a,b) = \alpha \tilde{x}(a,b)$ and $(\tilde{x} \land \tilde{y})(a,b) = \tilde{x}(a,b) \land \tilde{y}(a,b)$.

We also define a linear functional $\varphi$ on the vector space by
$$\varphi(u) = \|u_+(\epsilon,\epsilon)\|_1 - \|u_-(\epsilon,\epsilon)\|_1$$

This description in terms of functions incorporates information about entailments between sentences, whilst remaining context-theoretic in nature. The next proposition shows how logical and probabilistic properties of sentences of $\lambda$ are preserved in the vector representation.
\begin{prop}
If $x,y \in \lambda$ and $\mu(x) \vdash \mu(y)$ then $\Ent(x,y) = 1$; if $p$ is non-zero everywhere on $\Lambda$ except $\bot$ then the converse also holds. Moreover, if $x\in \lambda$, then $\varphi(\tilde{x}) = p_\vdash(\mu(x))$
\end{prop}
\begin{proof}
If $x,y \in \lambda$ and $\mu(x) \vdash \mu(y)$ then clearly $\tilde{x}(\epsilon,\epsilon) \le \tilde{y}(\epsilon,\epsilon)$, so $\Ent(x,y) = 1$. Conversely, if $\Ent(x,y) = 1$ then by the definition of $\varphi$ it must be the case that $0 < \tilde{x}(\epsilon,\epsilon) \le \tilde{y}(\epsilon,\epsilon)$, hence $x,y \in \lambda$. If $p$ is non-zero everywhere on $\Lambda - \{\bot\}$ then the only way this can be true is if $\mu(x) \vdash \mu(y)$. To see this, assume that $\mu(x) \nvdash \mu(y)$; then there exists an element of $\down{\mu(x)}$ that is not in $\down{\mu(y)}$, hence $\tilde{x}(\epsilon,\epsilon)$ will be non-zero in a component for which the corresponding component of $\tilde{y}(\epsilon,\epsilon)$ will be zero; hence $\tilde{x}(\epsilon,\epsilon)\nleq \tilde{y}(\epsilon,\epsilon)$, thus by contradiction, $\mu(x) \vdash \mu(y)$.
\end{proof}

Thus we have a vector based description of the language which preserves the logical and probabilistic nature of the translation, however we have not yet shown that this description is context-theoretic in nature --- i.e.~that the definition we have given has properties in common with context theories (other than that strings are represented by vectors). In fact, we will show a very close relationship between the description and the definition of meaning in terms of context that we used in the discussion on corpus models in chapter \ref{meaning-context}. We will need a more general definition than that used previously however --- the probabilistic nature of a corpus model is too restrictive to encompass the description we have given. Instead we define a \emph{general corpus model}\index{corpus model!general|textbf} on an alphabet $A$ to be a positive real-valued function over $A^*$. The definition of the context vector $\hat{x}$ of a string $x\in A^*$ still holds with a general corpus model; and again the vector space $\mathcal{A}$ generated by all such vectors is an algebra under the multiplication defined by concatenation of strings. What we will show is that a general corpus model can be associated with every probabilistic logical translation of a language.

\begin{prop}
Given a probabilistic logical translation $T = \langle\Lambda,\vdash, \bot, p, \lambda, \mu\rangle$ for $\lambda\subseteq A^*$  there exists a general corpus model $C_T$ over an alphabet $B$ and a one-to-one function $\psi$ from the space $V$ of functions from $A^*\times A^*$ to $L^1(\Lambda)$ to $L^\infty(B^*\times B^*)$ such that $\psi(\tilde{x}) = \hat{x}$.
\end{prop}
\begin{proof}
Let $B = A\cup A' \cup \{\diamond\}$ where $\diamond$ is an additional symbol, $\diamond \notin A \cup A'$. Define $C_T$ by:
$$C_T(x\diamond m) = p(m)$$
for all $x \in \lambda$ and all $m \in \down{\mu(x)}$, and $C_T$ is zero for all other elements of $B^*$. Let $u(a,b,m)$ be the basis element of $V$ which maps $(a,b) \in A^*\times A^*$ to $e_m$ in $L^1(\Lambda)$ and maps all other elements of $A^*\times A^*$ to $0$. Then we define $\psi$ by its operation on these basis elements:
$$\psi(u(a,b,m)) = e_{(a,b\,\diamond\,m)}.$$
Because $\diamond$ is not in $A$ or $A'$ this function must be one to one. Then using Definition \ref{x-tilde},
$$\tilde{x}(a,b\diamond m) = C_T(axb\diamond m) = p(m)$$
if $axb \in \lambda$ and $m\in\down{\mu(x)}$. Thus
$$\psi(\tilde{x}) = \sum_{a,b\,:\,axb \in \lambda}\left[ \sum_{m \in \down{\mu(axb)}} p(m)e_{(a,b\diamond m)}\right] = \hat{x}.$$
\end{proof}

This is an important result since it means that given a logical description of a language, we can construct a general corpus model incorporating this logical description, allowing us to make a strong link between logical and context-theoretic approaches: it allows us to think of the logical representation of a string as arising from the contexts in which the string occurs in a general corpus model. It also means that since the vector space $\mathcal{A}$ generated by the context vectors of $C$ is an algebra\index{algebra}, the vector space $\mathcal{A}'$ generated by the vectors $\{\tilde{x} : x \in A^*\}$ is also an algebra, again with multiplication defined by concatenation: $\tilde{x}\cdot \tilde{y} = \widetilde{xy}$. This is guaranteed to be well defined since it is well defined in the vector space $\mathcal{A}$ defined by $C_T$.

\subsection{Semantic Corpus Models}
\index{corpus model!semantic|(}

According to the context theoretic framework we have developed, the linear functional $\varphi$ when applied to the vector representation of a string is supposed to give the ``probability'' of that string. Clearly there is no such concept in model-theoretic semantics --- we can attach probabilities to logical forms giving them a Bayesian interpretation, but the concept of a probability of a string itself is foreign to model theoretic semantics. In fact the linear functional $\varphi$ we have defined behaves exactly like this: when $x$ is a sentence of the language $\lambda$, $\varphi(x)$ is the probability of the logical representation of $x$; if $x$ is not in the language, $\varphi(x)$ is zero; thus $\varphi$ does not conform to the context-theoretic ideal.

Yet if we are to truly find a way to combine context-theoretic techniques with model-theoretic approaches, we must find a way to link the concept of a probability of a string with these logical approaches; we should look for a linear functional that behaves more like a probability while still not ignoring the model theoretic nature of the representation.

In the linear functional $\varphi$ we are only using one context $(\epsilon,\epsilon)$; one way to give a non-zero probability to phrases that aren't sentences would be to consider other contexts. However here we face a practical problem; if we use all contexts, the value is not guaranteed to be finite.

One solution is to think of the probability\index{probability!of a string} of a string as being composed of two parts: the probability of the meaning of the string, and the probability that the meaning is expressed in that particular way. We can describe this by a probability distribution $q(x)$ over elements of $\lambda$, where $q$ satisfies the requirement
$$\sum_{x\in A^*}\left(\sum_{u \in \down{\mu(x)}}p(u)\right)q(x) = 1.$$
We can interpret this value as the conditional probability of observing string $x$ given that a string with a meaning at least as specific as the meaning of $x$ (its logical translation entails the logical translation of $x$) has been observed. Thus $q$ satisfies the requirement



We then give a new definition of the representation of a string in the vector space: it is still a function from $A^*\times A^*$ to $L^1(\Lambda)$; we define
$$\tilde{x}_q(a,b) = \begin{cases}
\sum_{u \in \down{\mu(axb)}} q(axb)p(u)e_u & \text{if $axb \in \lambda$}\\
0 & \text{otherwise.}
\end{cases}$$
When we use the general corpus model translation we defined in the previous section, we must define
$$C(x\diamond m) = q(x)p(m)$$
for all $x \in \lambda$ and all $m \in \down{\mu(x)}$, with $C$ zero for all other elements of $B^*$. We can view $C$ as being generated by a two stage process:
\begin{enumerate}
\item Choose a sentence $m\in\Lambda$ according to the probability distribution $p(m)$.
\item Choose a sentence $x\in\lambda$ such that $\mu(x) \in \down{m}$ according to $q(x)$.
\end{enumerate}
Because of the requirement we placed on $q$, we must have $\|C\|_1 = \sum_{u \in B^*} C(u) = 1$, so $C$ is a corpus model. Having a corpus model allows us to use the original linear functional $\phi$ defined for corpus models to measure the probability of a string. How are we to interpret this probability? We can think of $C$ as a ``semantic'' corpus model: it generates strings according to the probability $p$ of their meaning as well as the probability $q$ that this meaning is expressed in that particular way.


\index{corpus model!semantic|)}



\subsection{Representing Lexical Ambiguity}
\index{ambiguity!lexical|(}

The work of the previous section gives us the tools with which to describe lexical ambiguity within the context-theoretic framework. We are interested in descriptions of word sense ambiguity that allow us to incorporate statistical information about the probabilities of different senses and reason  about these in a way that is consistent with the context-theoretic philosophy.

Let us take a simple model of word sense ambiguity in which each word $w$ takes a finite number $n$ of senses $S(w) = \{w_1, w_2, \ldots w_n\}$. We assume that given a particular context $(a,b)$, we know which sense of the word is intended: each context is associated with exactly one sense in $S(w)$ so that the context completely disambiguates $w$. We can associate with each sense $w_i$ a set $[w_i]$ of contexts in which the word $w$ takes sense $w_i$. Similarly, given a corpus model $C$ we can define a context vector $\hat{w}_i$ with each sense $w_i$ which represents the contexts that that particular sense of $w$ occurs in:
$$\hat{w}_i(a,b) = \begin{cases}
\hat{w}(a,b) = C(awb) & \text{if } (a,b) \in [w_i]\\
0 & \text{otherwise.}
\end{cases}$$
Given this definition, we see that the context vectors of the senses of a word are disjoint in the vector lattice, and the context vector of a word is equal to the sum of the context vector of its senses: $\hat{w} = \sum_{i} \hat{w}_i$. Note that, just as we would expect, the context-theoretic probability of a word is the sum of the probability of its individual senses. Note also that the representations of the senses are disjoint because we assumed that each context completely disambiguated the word; if we relax this condition they will not necessarily be disjoint. Disjointness is thus not an essential feature, indeed it may not be useful in cases where a word has senses that are closely related.

We can define multiplication of senses with context vectors in a very straightforward way:
$$(\hat{w}_i \cdot u)(a,b) = \begin{cases}
(\hat{w}\cdot u)(a,b) & \text{if } (a,b) \in [w_i]\\
0 & \text{otherwise,}
\end{cases}$$
for $u \in \mathcal{A}$, and similarly for left-hand multiplication. This allows us to see how ambiguous words are partially disambiguated as they are concatenated with other words: we have
$$\hat{w}\cdot\hat{x} = \sum_i\hat{w}_i\cdot\hat{x}.$$
Thus as $w$ is concatenated with a string $x$, its representation remains the sum of its senses multiplied by $\hat{x}$, however since each sense only occurs in a subset of the possible contexts, $x$ has the effect of partially disambiguating $w$, and the left hand side of the equation becomes more similar to one of the summands.

This analysis provides us with a simple formula for representing a word in terms of its senses, given the methods of the previous sections: we treat each sense exactly as if it were an unambiguous word; build a context-theoretic representation using the senses, then represent the ambiguous word as the sum of the representation of its individual senses. For example, using the ideas of the previous section, we can define a semantic corpus model based on a probabilistic logical translation in which we assume we only ever deal with senses, for which the logical translation can be well defined. We can then represent the word as the sum of the representation of its individual senses. The probabilistic logical translation and the function $q$ can be interpreted as disambiguating\index{word sense disambiguation} the word. There are several kinds of disambiguation that can occur:
\begin{itemize}
\item We do not distinguish between different parts of speech when we talk about senses; for example the representation of a word like ``book'' will include both noun and verb parts. As words are concatenated with this word, only the senses that can make the phrase grammatical (i.e.~that occur as a substring of $\lambda$) will remain, disambiguating parts of speech\index{parts of speech, disambiguating}.
\item The entailment relation and $p$ provide \emph{semantic disambiguation}\index{semantic disambiguation}: in a particular context those senses which lead to sentences which are meaningless and thus whose meaning is assigned a value of $0$ by $p$ will be eliminated so that only senses which are meaningful in the context remain. Similarly, senses which produce a meaning in the given context which is very unlikely will be assigned a low probability by $p$.
\item The function $q$ provides \emph{statistical disambiguation}\index{statistical disambiguation} --- it reduces emphasis on senses of words which are statistically unlikely based on the context, although the resulting meaning may not be unlikely; thus this function has a r\^ole similar to current word sense disambiguation techniques.
\end{itemize}

We have shown in this section that the framework provides ample room for the representation of word sense ambiguity and its disambiguation; an ambiguous word can be represented by summing the representations of individual senses. Although this is a simple analysis of the situation it gives us a method for representing lexical ambiguity and a picture of how words are disambiguated within the framework: as more context is added to a word it gradually becomes less ambiguous.

\index{ambiguity!lexical|)}


\index{uncertainty, representing|)}

\section{Outline of Possible Implementations}
\label{practical-issues}

We chose to describe the models of the preceding sections in a manner which was extremely general and also mathematically simple. This allowed us to present the concepts clearly without concern for how we could represent and compute with such models. Clearly it is impractical to explicitly represent a sentence as a sum over a (potentially infinite) number of dimensions. Instead, we imagine that in practice, systems that make use of the mathematics we have presented here will make use of standard representations for the logical aspect of the representation; the statistical or algebraic aspects can then be computed separately, while making use of the existing algorithms for computing with logic.

To make this clearer, we will outline how such a system may be constructed. 
We assume we have at our disposal a method for computing entailments between sentences of the logical language $\Lambda$. In most cases, $\Lambda$ will include the propositional calculus\index{propositional calculus} as a subset, and thus the equivalence classes of $\vdash$ will form a Boolean algebra. The main problem facing us is the function $p$ which is defined on sentences of $\Lambda$. In fact, we can do without $p$ itself, and assume we have at our disposal the function $p_\vdash$ which will be a probability measure on the Boolean algebra\index{Boolean algebra} of equivalence classes of $\vdash$. This means we will not have to compute sums of $p$ over sentences of $\Lambda$, a potentially impossible task. The function $p_\vdash$ can be assigned in many ways, for example:
\begin{itemize}
\item A simple heuristic could be used. For example, this could be an information-theory inspired measure based on the length of the logical expression: $p_\vdash(u) = k^{-|u|}$ where $k$ is a constant and $|u|$ is the length of the shortest member of the equivalence class of $u \in \Lambda$. This would have the advantage of being simple to compute yet fairly consistent with the requirements of $p_\vdash$: in general it is likely that if $u\vdash v$ then $p_\vdash(u) \le p_\vdash(v)$ will hold since $v$ can be expressed at least as simply as $u$.
\item A value for $p_\vdash$ could be assigned based on a probabilistic logic: this may be a fuzzy logic\index{fuzzy logic} such as \L ukasiewicz logic\index{Lukasiewicz logic@\L ukasiewicz logic} \citep{Kundu:94} or Basic Fuzzy Logic \citep{Hajek:98}, or a first order logic such as that of \cite{Nilsson:86} and later variations.
\end{itemize}
Both these approaches could make use of techniques that assign probabilities to concepts in an ontology; these are described in Chapter \ref{ontologies}. 

We assume for now that we are only interested in the representation of sentences; all uncertainty is described by a weighted sum over representations of sentences. Given a natural language sentence the system may for example parse the sentence and perform word sense disambiguation and anaphora resolution. Each of these can result in probabilistic information about which parses, senses and referents are intended, thus we will be left with a probability distribution over possible interpretations of the sentence.  Each interpretation is completely unambiguous and thus ready to be translated into logic; for efficiency purposes, these could be sorted by probability, and only the most probable interpretations retained. We will thus have a logical expression for each interpretation and we can compute the probability $p_\vdash$ for each of these.

At the end of this process, a sentence $s$ is represented as a list of pairs $\langle u_i, \alpha_i\rangle$, each specifying a logical translation $u_i$ and the probability $\alpha_i$ of the combined statistical information. At this point we can compute probabilities for sentences using the linear functional $\phi$:
$$\phi(\tilde{s}) = \sum_i \alpha_i p_\vdash(u_i),$$
where $p_\vdash(u_i)$ is the probability of the logical sentence $u_i$. However these probabilities are unlikely to coincide with our normal conception of the probability of a string, since they are the combination of probabilities assigned by the parser and probabilities of logical expressions, which need not necessarily coincide with the probabilities of strings. We can compensate for this however, by making use of a function $q'(u_i | s_i)$ which specifies the probability of the logical expression $u_i$ given that it is translated from the specific interpretation $s_i$ of the sentence under consideration, similar to the function $q$ we discussed earlier --- however this is clearly a difficult value to estimate directly. On the other hand, the problem of estimating the probability of a string is well understood; we can make use of one of the many language modelling techniques to do this, for example we could use an $n$-gram. This value of the probability of a string then provides a renormalising condition which allows us make the vector representing the string fit the expected probability of the string; define a constant $c_s = l(s)/\phi(\tilde{s})$, where $l(s)$ is the probability assigned to the string $s$ by the language model. The renormalised string is then represented by the list of pairs $\langle u_i, \beta_i\rangle$ where $\beta_i = c_s\alpha_i$. The renormalising constant $c_s$ thus plays the role of the function $q'$.

Computing the degree of entailment between the representations of strings causes some difficulties. This is because we have represented strings as sums over the vector representations of logical sentences which are not disjoint in the vector lattice. Given two such representations $\tilde{s}_1 = \sum_i \beta^{(1)}_i\tilde{u}^{(1)}_i$ and $\tilde{s}_2 = \sum_i \beta^{(2)}_i\tilde{u}^{(2)}_i$, where $\tilde{u}^{(k)}_i$ is the vector representation of the logical expression $u^{(k)}_i$, we need to compute $\phi(\tilde{s}_1 \land \tilde{s}_2)$ in order to obtain the degree of entailment. However addition does not distribute with respect to the lattice meet operation except when the addition is between disjoint elements of the vector lattice; since the vector representations of the logical sentences are not in general disjoint, there is no way to find $\tilde{s}_1\land\tilde{s}_2$ in terms of the meets of their summands. The solution to this is to find a set $S$ of disjoint logical sentences such all the $u_i$ can be written as a disjunction of elements of $S$. This is possible using the canonical form\index{Boolean algebra!canonical form} of a Boolean algebra in which each element is written as a join of \emph{minterms} \citep{Birkhoff:48}. (This could potentially be computationally expensive --- given $n$ sentences there could be $2^n$ minterms.) For disjoint positive elements $a$ and $b$ of a vector lattice, $a \lor b = a + b$, so given the set $S$ each sentence $u_i$ can be written as a sum of disjoint elements, the meet operation becomes trivial and the degree of entailment can easily be computed.

An alternative approach is to compute a lower bound on the degree of entailment\index{entailment!degree of}. Since $\beta^{(k)}_iu^{(k)}_i \le \tilde{s}_k$ for each $i$ and $s_k$, we have
$\beta^{(1)}_i\tilde{u}^{(1)}_i \land \beta^{(2)}_j\tilde{u}^{(2)}_j  \le \tilde{s}_1\land\tilde{s}_2$
for all $i$ and $j$, and hence
$$\phi_\mathrm{min}(\tilde{s}_1\land\tilde{s}_2) = \max_{i,j}\left[ \phi(\beta^{(1)}_i\tilde{u}^{(1)}_i \land \beta^{(2)}_j\tilde{u}^{(2)}_j)\right]  \le \phi(\tilde{s}_1\land\tilde{s}_2).$$
The left hand side thus provides us with a lower bound $\phi_\mathrm{min}(\tilde{s}_1\land\tilde{s}_2)$ on the context-theoretic probability of the meet of the representations of the two sentences, which can be used to calculate the degree of entailment. This lower bound can be thought of as the greatest probability obtained by taking meets between individual interpretations of the two sentences. It is straightforward to calculate:
$$\phi_\mathrm{min}(\tilde{s}_1\land\tilde{s}_2) = \max_{i,j}\left[ (\min\{\beta^{(1)}_i,\beta^{(2)}_j\})p_\vdash(u^{(1)}_i\land u^{(2)}_j) \right].$$
Note that  $\land$ here is logical conjunction from the language $\Lambda$. This is possible since in a logic with the propositional calculus as a subset, the vector representation of the logical conjunction of two sentences will be the same as the vector lattice meet of the representations of the individual sentences, for the reason discussed in Section \ref{propositional}. The lower bound on the degree of entailment $\Ent(s_1,s_2)$ is then given by $\phi_\mathrm{min}(\tilde{s}_1\land\tilde{s}_2)/\phi(\tilde{s}_1)$.

\subsection{Entailment between words and phrases}

Computing entailment between words and phrases using the ideas of Section \ref{context-theoretic-analysis} and subsequent sections is clearly more challenging than computing entailment between sentences, since we need to calculate a sum over all contexts $(a,b) \in A^*$. One approach to this problem would be to use a Monte-Carlo technique to estimate the entailment by taking a sample of contexts. In fact, only those contexts which give a sentence in $\lambda$ will contribute to the sum, and heuristics could be used to skew the sample towards those contexts which are likely to be important for the string under consideration.

\section{Conclusion}

We have presented a context-theoretic analysis of logical semantics for natural language, and shown how the flexibility of the vector representation that comes with the context-theoretic framework allows the incorporation of statistical information about uncertainty into the representation. This provides us with a principled way of reasoning with uncertainty and ambiguity in meaning.

We discussed some requirements that we may expect of a system that represents ambiguity and uncertainty in natural language, namely:
\begin{itemize}
\item that the system be able to reason with uncertainty in a probabilistic fashion, following a Bayesian\index{Bayesianism} philosophy. The mathematics we have described allows for the incorporation of information about the probability of meaning, in the Bayesian sense.
\item that the system deals with ambiguity in a way that agrees with our intuition and that incorporates statistical information about this ambiguity. This is true of the ideas presented here: an ambiguous word or phrase is represented as a weighted sum of its unambiguous meanings. It is the weights given to these meanings that allow statistical information to be incorporated.
\end{itemize}

We have also shown how a system may be implemented using the ideas presented here, and outlined how the computational problems involved may be solved.

It should be noted that the approaches presented in this chapter are just a few ways of dealing with the problems of ambiguity and uncertainty in logical semantics within the context-theoretic framework; it is likely that future work within the framework will bring to light new approaches and computational techniques.

Among the problems that need addressing are questions surrounding multi-word expressions and non-compositionality --- is there a way to identify context-theoretic properties of words and phrases that may indicate non-compositionality, and how may existing approaches to representing non-compositionality be incorporated into the framework? These are questions that we hope to address in future work.

Other areas of interest for future work include looking at how different probabilistic logics relate to the framework and which are best suited to it and to representing natural language, looking at computational procedures for calculating or estimating the degree of entailment when using logical semantics, especially between words and phrases, and other ways to make logical semantics more robust, for example by combining them with other context theories.

\index{logical semantics|)}

 \chapter{Taxonomies and Vector Lattices}
 \label{ontologies}
 \index{taxonomy|(}
 
A crucial feature that we require of the context-theoretic framework\index{context-theoretic!framework} is that we are able to make use of logical representations of meaning within the framework. Ontologies\index{ontologies} form an important part of many systems that deal with logical representations of natural language, thus it is important to examine the relationship between ontological representations of meaning and vector based ones. In this chapter we show how an important part of an ontology, a taxonomy, can be represented in terms of vectors in a vector lattice, by means of \emph{vector lattice completions}\index{vector lattice completion}, a concept that we define. The ideas presented in this chapter marry the vector-based representations of meaning with the ontological ones by considering both from the unifying perspective of vector lattice theory.

The constructions presented here may have several practical benefits:
\begin{itemize}
\item  They provide a link with statistical representations of meaning such as latent semantic analysis\index{latent semantic analysis} and distributional similarity\index{distributional similarity} measures by showing that taxonomic properties of meaning can be represented within the vector space structures of these techniques. Through this, the ideas presented here in combination with such techniques may lead to new methods of automatic ontology construction. For example, by relating semantic analysis vectors to the taxonomy vectors it may be possible to place a new concept in the vector space of the taxonomy based on its latent semantic analysis vector.
\item The vector-based representation of a taxonomy can be used to build context theories that make use of the taxonomy whilst remaining entirely vector-based, allowing the use of techniques to combine vectors such as tensor or free products, discussed in the next chapter. These could lead to new approaches to natural language semantics that would potentially be more robust than logical approaches since they would be more amenable to incorporating statistical features of language, being entirely vector based.
\item Vector spaces give us a lot of flexibility: vectors can be scaled, rotated, translated, and the dimensionality of the vector space can be reduced. These properties may lead to new techniques for the efficient representation of meaning. For example, it may be possible to use a dimensionality reduction to efficiently represent a taxonomy in terms of vectors.
\end{itemize}
Perhaps more importantly though, the subject of this chapter is the nature of meaning itself, and the techniques we present here show that the vector lattice representations of meaning can be viewed as a generalisation of ontological ones. In addition to the lattice structure of ontologies, however, vector lattices allow a more subtle description of meaning that allows the quantification of nearness of meaning that cannot be described fully in the lattice structure of ontologies; part of the success of techniques such as latent semantic analysis is due to their ability to quantify nearness of meaning in this way. 

The contributions of this chapter are as follows:
\begin{itemize}
\item We give a definition for a vector lattice completion\index{vector lattice completion} as a way of representing a taxonomy in terms of a vector lattice.
\item We describe several vector lattice completions with different properties:
\begin{itemize}
\item A \emph{probabilistic completion}\index{vector lattice completion!probabilistic} (see Section \ref{probabilistic-completion-section}) allows the incorporation of the ``probability of a concept'' into the vector-based description.
\item We describe a \emph{distance preserving completion}\index{vector lattice completion!distance preserving} (see Section \ref{distance-section}) in which the distance between vectors in the vector lattice representation is the same as the distance in the ontology using a measure of \cite{Jiang:97}.
\item The vector space representation typically uses a large number of dimensions. In Section \ref{efficient-completions-section} we describe a vector lattice completion that in many cases uses a smaller number of dimensions than the probabilistic completion, and discuss its application to two real world ontologies.
\end{itemize}
\item The constructions we present allow the description of taxonomic concepts in terms of vectors; in Section \ref{representing-words-section} we discuss the representation of terms which may be ambiguous\index{ambiguity}, requiring the representations of the individual senses of a term to be combined.
\item In Section \ref{dist-sim-projections-section} we analyse certain measures of distributional similarity and show how they can be thought of in terms of projections on a vector space. This leads us to a representation of ambiguous terms in terms of sums of projections, described in Section \ref{ideal-projection}. In this representation, in addition to the vector lattice properties of previous vector lattice completions, multiplication is defined, meaning that we can define a context theory. This construction may be useful in situations where ontological representations are needed as part of a larger context theory.
\end{itemize}

\section{Taxonomies}
\label{taxonomy}

Ontologies describe relationships between concepts. They are considered to be of importance in a wide range of areas within artificial intelligence and computational linguistics. For example, WordNet\index{WordNet} \citep{Fellbaum:98} is an ontology that describes relations between word senses, or more accurately, senses of \emph{terms}, since WordNet also describes the meanings of collocations.

\begin{figure}
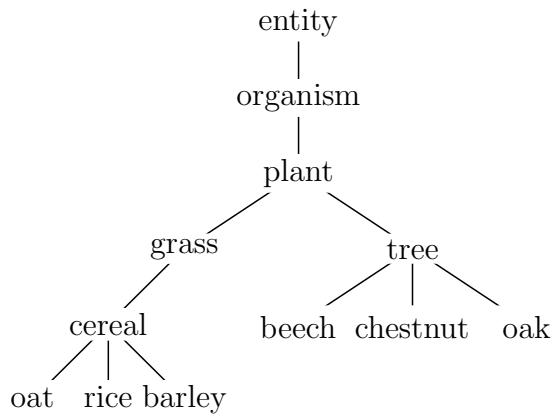

\begin{center}
\begin{graph}(7,6)(0,-5.6)
\textnode{entity}(3.5,0){entity}[\graphlinecolour{1}]
\textnode{organism}(3.5,-1){organism}[\graphlinecolour{1}]
\textnode{plant}(3.5,-2){plant}[\graphlinecolour{1}]
	\textnode{grass}(2,-3){grass}[\graphlinecolour{1}]
	\textnode{cereal}(1,-4){cereal}[\graphlinecolour{1}]
		\textnode{oat}(0,-5){\rule[-0.5ex]{0pt}{2.1ex}oat}[\graphlinecolour{1}]
		\textnode{rice}(1,-5){\rule[-0.5ex]{0pt}{2.1ex}rice}[\graphlinecolour{1}]
		\textnode{barley}(2,-5){\rule[-0.5ex]{0pt}{2.1ex}barley}[\graphlinecolour{1}]
	\textnode{tree}(5,-3){tree}[\graphlinecolour{1}]
		\textnode{beech}(3.5,-4){\rule{0pt}{2ex}beech}[\graphlinecolour{1}]
		\textnode{chestnut}(5,-4){\rule{0pt}{2ex}chestnut}[\graphlinecolour{1}]
		\textnode{oak}(6.5,-4){\rule{0pt}{2ex}oak}[\graphlinecolour{1}]

\edge{entity}{organism}
\edge{organism}{plant}
\edge{plant}{grass}
\edge{grass}{cereal}
	\edge{cereal}{oat}
	\edge{cereal}{barley}
	\edge{cereal}{rice}
\edge{plant}{tree}
	\edge{tree}{beech}
	\edge{tree}{chestnut}
	\edge{tree}{oak}


\end{graph}
\end{center}
\caption{A small example taxonomy extracted from WordNet \citep{Fellbaum:98}.}
\label{plant-taxonomy}
\end{figure}

Arguably the most important relation described in an ontology is the \textbf{is-a} relation (also called subsumption), which describes inclusion between classes of objects.  When applied to meanings of terms, the relation is called \emph{hypernymy}\index{hypernymy}. For example, a \emph{tree} is a type of \emph{plant} (the concept \emph{plant} subsumes \emph{tree}), thus the word ``plant'' is a hypernym of ``tree''. The converse relationship between terms is called hyponymy, so ``tree'' is a hyponym of ``plant''. A system of classification that only deals with the \textbf{is-a} relation is referred to as a \emph{taxonomy}. An example taxonomy is shown in figure \ref{plant-taxonomy}, with the most general concept at the top, and the most specific concepts at the bottom.

The \textbf{is-a} relation is in general a partial ordering, since
\begin{itemize}
\item it is always the case that an $a$ is an $a$ (reflexivity);
\item if an $a$ is a $b$ and a $b$ is an $a$ then $a$ and $b$ are the same (anti-symmetry).
\item if an $a$ is a $b$ and a $b$ is a $c$ then an $a$ is necessarily a $c$ (transitivity).
\end{itemize}

The taxonomy described by figure \ref{plant-taxonomy} has a special property: it is a tree\index{tree}, i.e.~no concept directly subsumed by one concept is directly subsumed by any other concept. This type of taxonomy will be studied in section \ref{distance-section}.

Later we will discuss ``distance measures'' on ontologies. These can be as simple as measuring the shortest number of links between two concepts \citep{Rada:89} or be information theoretic measures based on the ``probability of a concept'' \citep{Resnik:95,Jiang:97}.

\index{ontologies}

\subsection{Vector Lattice Embeddings of Taxonomies}


Vector representations of meaning do not seem to sit nicely with ontological representations of meaning --- the former make use of vector spaces and the latter make use of lattices. In fact, what we will show in this chapter is that the two types of representation can be combined within the structure of a vector lattice\index{vector lattice}, a space that is simultaneously a vector space and a lattice. Taxonomies can be embedded in a vector lattice in such a way that the lattice structure is preserved, and existing vector representations of meaning can be considered as implicitly carrying a lattice structure.


The relationship between concepts in a taxonomy is expressed by means of a partial order\index{partial ordering}, and we wish to embed the partial ordering representation in a vector lattice; we call such an embedding a \emph{vector lattice completion}. The partial ordering of the vector lattice representation must still therefore contain the partial ordering of the taxonomy, but in addition, we provide each meaning with a concrete position in some $n$-dimensional space. We define this formally as follows:
\begin{defn}[Vector Lattice Completion]\index{vector lattice completion|textbf}
Let $S$ be a partially ordered set. A \emph{vector lattice completion} of $S$ is a vector lattice $V$ and a function $\psi$ from $S$ to $V$ that is a partial ordering homomorphism, i.e.~$\psi(s_1) \le \psi(s_2)$ if and only if $s_1 \le s_2$, for all $s_1, s_2 \in S$.
\end{defn}

Because the embedding will necessarily be a \emph{lattice completion}, it will introduce new operations of meet and join on elements (see section \ref{lattices}). Many taxonomies may already have some of the properties of a lattice, for example, most taxonomies are join semilattices. However the existing join operation is not usually directly useful since it does not correspond with our usual idea of logical disjunction. For example, in figure \ref{plant-taxonomy}, the join of the concepts \emph{beech} and \emph{oak} is \emph{tree}. If something is a \emph{beech} or an \emph{oak}, it is definitely a \emph{tree}, however the converse provides problems: if something is a \emph{tree} it does not follow that the thing is necessarily a \emph{beech} or an \emph{oak}---since it could also be a \emph{chestnut}. Thus the logical disjunction of the concepts \emph{beech} and \emph{oak} should sit somewhere between these two concepts and \emph{tree}.

\subsection{Probabilistic Completion}
\label{probabilistic-completion-section}
\index{vector lattice completion!probabilistic|(}

We are also concerned with the probability of concepts. This is an idea that has come about through the introduction of ``distance measures''\index{distance measures!on taxonomies} on taxonomies \citep{Resnik:95}. Since terms can be ascribed probabilities based on their frequencies of occurrence in corpora, the concepts they refer to can similarly be assigned probabilities. The probability of a concept is the probability of encountering an instance of that concept in the corpus, that is, the probability that a term selected at random from the corpus has a meaning that is subsumed by that particular concept. This ensures that more general concepts are given higher probabilities, for example if there is a most general concept (a top-most node in the taxonomy, which may correspond for example to ``entity'') its probability will be one, since every term can be considered an instance of that concept.

We give a general definition based on this idea which does not require probabilities to be assigned based on corpus counts:
\begin{defn}[Real Valued Taxonomy]
A real valued taxonomy\index{taxonomy!real valued|textbf} is a finite set $S$ of \emph{concepts} with a partial ordering $\le$ and a positive real function $p$ over $S$. The \emph{measure} of a concept is then defined in terms of $p$ as
$$\hat{p}(x) = \sum_{y \in \down{x}} p(y).$$

The taxonomy is called \emph{probabilistic}\index{taxonomy!probabilistic|textbf} if $\sum_{x \in S} p(s) = 1$. In this case $\hat{p}$ refers to the \emph{probability of a concept}.
\end{defn}
Thus in a probabilistic taxonomy, the function $p$ corresponds to the probability that a term is observed whose meaning corresponds (in that context) to that concept. The function $\hat{p}$ denotes the probability that a term is observed whose meaning in that context is subsumed by the concept.

Note that if $S$ has a top element $I$ then in the probabilistic case, clearly $\hat{p}(I) = 1$. In studies of distance measures on ontologies, the concepts in $S$ often correspond to senses of terms, in this case the function $p$ represents the (normalised) probability that a given term will occur with the sense indicated by the concept. The top-most concept often exists, and may be something with the meaning ``entity''---intended to include the meaning of all concepts below it.

The most simple completion we consider is into the vector lattice $L^\infty(S)$, the real vector space of dimensionality $|S|$, with basis elements $\{e_x : x\in S\}$.
\begin{prop}[Ideal Vector Completion]\index{ideal vector completion}
Let $S$ be a probabilistic taxonomy with probability distribution function $p$ that is non-zero everywhere on $S$. The function $\psi$ from $S$ to $L^\infty(S)$ defined by
$$\psi(x) = \sum_{y \in \down{x}} p(y)e_y$$
is a completion of the partial ordering of $S$ under the vector lattice order of $L^\infty(S)$, satisfying $\|\psi(x)\|_1 = \hat{p}(x)$.
\end{prop}
\begin{proof}
The function $\psi$ is clearly order-preserving: if $x \le y$ in $S$ then since $\down{x} \subseteq \down{y}$, necessarily $\psi(x) \le \psi(y)$. Conversely, the only way that $\psi(x) \le \psi(y)$ can be true is if $\down{x} \subseteq \down{y}$ since $p$ is non-zero everywhere. If this is the case, then $x \le y$ by the nature of the ideal completion. Thus $\psi$ is an order-embedding, and since $L^\infty(S)$ is a complete lattice, it is also a completion. Finally, note that $\|\psi(x)\|_1 = \sum_{y\in\down{x}} p(y) = \hat{p}(x)$.
\end{proof}
This close connection with the ideal completion is what leads us to call it the \emph{ideal vector completion}. The completion allows us to represent concepts as elements within a vector lattice so that not only the partial ordering of the taxonomy is preserved, but the probability of concepts is also preserved as the size of the vector under the $L^1$ norm.

\index{vector lattice completion!probabilistic|)}

\subsection{Distance Preserving Completion}
\label{distance-section}
\index{vector lattice completion!distance preserving|(}

Some attempts have been made to link ontological representations with statistical techniques. These centre around measures of semantic distance which attempt to put a value on semantic relatedness between concepts.

\cite{Jiang:97}\index{Jiang and Conrath|(} defined a distance measure 
based on the information content of concepts \citep{Resnik:95}, which can be derived from their probabilities. We will show that this measure has the following property: concepts can be embedded in a vector lattice in such a way that the distance between concepts in the vector lattice is equal to the Jiang-Conrath distance measure.\footnote{Further investigation is required to determine whether other distance measures possess this property.}


We are able to show that the distances are preserved in certain types of taxonomy: the concepts must form a \emph{tree}:
\begin{defn}[Trees]\index{tree|textbf}
A partially ordered set $S$ is called a \emph{tree} if every element $x$ in $S$ has at most one element $y\in S$ such that $x \prec y$ and there is an element $I$ such that $z \le I$ for all $z \in S$. The unique element preceding $x$ is called the \emph{parent} of $x$, it is denoted $\Par(x)$ if it exists.
\end{defn}\noindent
Note that in a tree only the topmost element $I$ has no parent.

The Jiang-Conrath measure makes use of a particular property of trees. It is easy to see that a tree forms a \emph{semilattice}: for each pair of elements $x$ and $y$ there is an element $x \lor y$ that is the \emph{least common subsumer} of $x$ and $y$. For example, in figure \ref{plant-taxonomy}, the least common subsumer of \emph{oat} and \emph{barley} is \emph{cereal}; the least common subsumer of \emph{oat} and \emph{beech} is \emph{plant}.

The measure also makes use of the information content\index{information content} of a concept; this is simply the negative logarithm of its probability. In our formulation, the information content $\mathit{IC}(x)$ of a concept $x$ is defined by
$$\mathit{IC}(x) = -\log \hat{p}(x).$$
The information content thus decreases as we move up the taxonomy; if there is a most general element $I$, it will have an information content of zero.

The Jiang-Conrath distance measure $d(x,y)$ between two concepts $x$ and $y$ is then defined as
$$d(x,y) = \mathit{IC}(x) + \mathit{IC}(y) - 2\mathit{IC}(x\lor y).$$
There is a notable similarity between this expression and a relation that holds in vector lattices:
\begin{equation*}\tag{$*$}\label{vlid}|u - v| = u + v - 2(u\land v),\end{equation*}
for all $u$ and $v$ in the vector lattice. This formula provides the starting point for preserving distances in the vector lattice completion.

In its current form, in building a vector lattice we cannot simply replace the function $\hat{p}$ with the information content, since $\hat{p}$ must increase as we move up the taxonomy; instead we must invert the direction of the lattice. This allows us to embed concepts in the lattice while retaining the information content as the norm, and changes joins into meets, so that distances correspond to the Jiang-Conrath\index{Jiang and Conrath|)} measure.
\begin{prop}[Distance Preserving Completion]
Let $S$ be a probabilistic taxonomy which forms a tree with partial ordering $\le$. The function $\mathit{IC}$ defines a positive real-valued function $f_\mathit{IC}$ by
$$f_\IC(x) = \IC(x) - \IC(\Par(x)).$$
for $x \in S - \{I\}$, and $f_\IC(I) = 0$. We define a new partial ordering $\le'$ on $S$ by $x \le' y$ iff $y \le x$ (thus $\le'$ is the \emph{dual} of $\le$). Then $f_\IC$ together with the new partial ordering defines a real-valued taxonomy on $S$. Call the function that maps an element of $S$ to its completion in the new taxonomy $\psi'$. The vector lattice completion of the new taxonomy satisfies $\|\psi'(x)\|_1 = \IC(x)$ and $\|\psi'(x) - \psi'(y)\|_1 = d(x,y)$. 
\end{prop}

\begin{proof}
For the results about vector lattices used here see section \ref{vector-lattices}. Because the taxonomy is a tree, $f_\IC$ is clearly a positive function satisfying $\|\psi'(x)\|_1 = \IC(x)$. To see the second part, we need to know that the vector lattice $L^\infty(S)$ with the $L_1$ norm is an \emph{AL-space}; this means that $\|s + t\| = \|s\| + \|t\|$ whenever $s \land t = 0$. We have here
$$(u - u\land v)\land (v - u\land v)  = \tfrac{1}{2}(u + v - 2(u\land v) - |u - v|) = 0,$$
where we have used the above identity twice. Thus, using the same identity, we have
\begin{eqnarray*}
 \|u - v\|_1 = \| |u - v| \|_1&=& \|u + v - 2(u \land v)\|_1\\
		&=& \|(u - u\land v) + (v - u \land v)\|_1\\
		&=& \|u - u\land v \|_1 + \|v - u \land v \|_1\\
		&=& \|u\|_1 + \|v\|_1 -2\|u\land v\|_1
\end{eqnarray*}
For the last step, we used the fact that we are dealing with positive elements, with $u - u\land v \ge 0$ and thus, using the additive property of the $L_1$ norm, $\|u\| = \|u - u\land v + u\land v\| = \|u - u\land v\| + \|u\land v\|$.

Finally, note that the lattice completion is built from the dual of a tree, which is a join semilattice. Joins are preserved as meets in the completion since $\down{x} \cap \down{y} = \down{x \land y}$, and thus we have $\psi'(x) \land \psi'(y) = \psi'(x\lor y)$. This completes the proof.
\end{proof}

Thus we have shown that it is possible to simultaneously preserve the partial ordering of an ontology and the distance between concepts (as measured by Jiang and Conrath) within a vector lattice representation. We believe this particular result opens up the potential for a wide range of techniques combining statistical methods of determining meaning with ontological representations. For example, we might expect that distributional similarity measures can be used as a predictor of semantic similarity --- i.e.~that the distributional similarity of two terms is correlated to the semantic distance between the concepts\footnote{This assumes the terms are unambiguous; ambiguity would make the detection of correlation more difficult.} the terms represent; this idea would be compatible with Harris' distributional hypothesis \citep{Harris:68}.
Indeed measures of distributional similarity have been used to place terms within a semantic hierarchy such as WordNet \citep{Alfonseca:02, Pekar:03}. In addition to providing new avenues for research in this task, our results may allow terms to be automatically placed within the fine-grained structure allowed by the vector lattice representations. Measures of distributional similarity could also be used to refine vector lattice representations of existing taxonomies by moving concepts so that their position in the vector lattice matches what we would expect based on measuring the distributional similarity of the corresponding terms.

\index{vector lattice completion!distance preserving|)}

\subsection{Efficient Completions}
\label{efficient-completions-section}
\index{vector lattice completion!efficient|(}

\begin{figure}
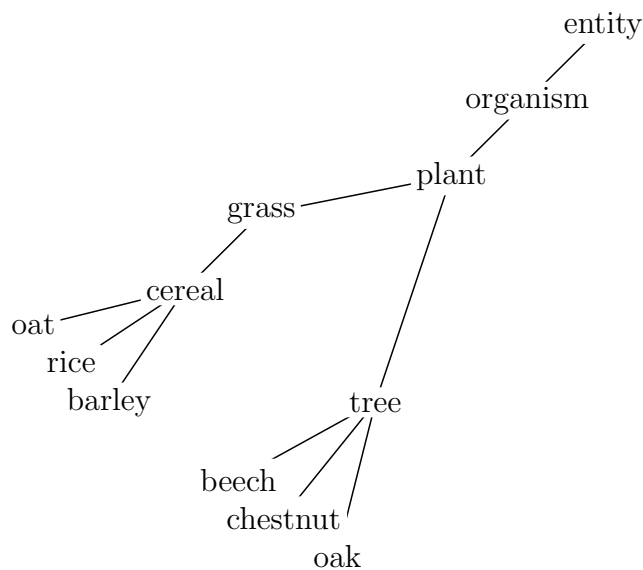

\begin{center}
\begin{graph}(7,7.5)(0,-7)
\textnode{entity}(6.5,0){entity}[\graphlinecolour{1}]
\textnode{organism}(5.5,-1){organism}[\graphlinecolour{1}]
\textnode{plant}(4.5,-2){plant}[\graphlinecolour{1}]
	\textnode{grass}(2,-2.5){grass}[\graphlinecolour{1}]
	\textnode{cereal}(1,-3.5){cereal}[\graphlinecolour{1}]
		\textnode{oat}(-1,-4){\rule[-0.5ex]{0pt}{2.1ex}oat}[\graphlinecolour{1}]
		\textnode{rice}(-0.5,-4.5){\rule[-0.5ex]{0pt}{2.1ex}rice}[\graphlinecolour{1}]
		\textnode{barley}(0,-5){\rule[-0.5ex]{0pt}{2.1ex}barley}[\graphlinecolour{1}]
	\textnode{tree}(3.5,-5){tree}[\graphlinecolour{1}]
		\textnode{beech}(1.7,-6){\rule{0pt}{2ex}beech}[\graphlinecolour{1}]
		\textnode{chestnut}(2.3,-6.5){\rule{0pt}{2ex}chestnut}[\graphlinecolour{1}]
		\textnode{oak}(3,-7){\rule{0pt}{2ex}oak}[\graphlinecolour{1}]

\edge{entity}{organism}
\edge{organism}{plant}
\edge{plant}{grass}
\edge{grass}{cereal}
	\edge{cereal}{oat}
	\edge{cereal}{barley}
	\edge{cereal}{rice}
\edge{plant}{tree}
	\edge{tree}{beech}
	\edge{tree}{chestnut}
	\edge{tree}{oak}


\end{graph}
\end{center}
\caption{It is possible to embed a tree into a two dimensional vector lattice in such a way that the partial ordering is preserved. Two concepts $s_1$ and $s_2$ satisfy $s_1 \le s_2$ if $s_1$ is to the left or level with and below or level with $s_2$.}
\label{plant-taxonomy-rot}
\end{figure}

In this section we discuss the question of how many dimensions are necessary to maintain the lattice structure in the vector lattice completion. The representations discussed previously use a very large number of dimensions: one for each node in the ontology. To see that this is more than is generally needed, consider an ontology whose Hasse diagram\index{Hasse diagram!planar} is planar: that is it can be rearranged so that no lines cross (see figure \ref{plant-taxonomy-rot}). If we then position the nodes in the diagram such the lines between nodes are at an angle of less than $45^\circ$ to the vertical (this can always be done by stretching the diagram out vertically), and we rotate the diagram by $45^\circ$ to the right, and set an origin, the position of each node in the two dimensional diagram can be considered as a representation of the concept in the vector lattice $\R^2$. It is easy to see that the partial ordering is preserved --- if $x\le y$ in the partial ordering, then this will also hold in the two-dimensional vector lattice, although care has to be taken in the positioning of concepts to ensure that other unwanted relations can't be derived in the new space.

One problem with this simplistic vector lattice representation is that there is no obvious way to interpret the two dimensions. Another, more serious problem is that it is not unique: in general there are many ways we can draw the Hasse diagram, and each will correspond to a different representation. Concepts will necessarily be positioned arbitrarily according to which way the diagram is drawn, leaving us in doubt as to whether the vector aspect of the representation is meaningful. This arbitrary positioning means that distances between nodes are dependent on how we draw the Hasse diagram: in one representation a pair of nodes may be close together, while in another they may be far apart. For example, in the Hasse diagram of a tree, we can swap leaf nodes any way we wish to make a pair of nodes arbitrarily close or far apart.

We call representations that don't have this property \emph{symmetric}: a representation is symmetric if the distances $\|x - y\|$ between the representation of a pair of nodes is only dependent on the lattice properties of the nodes represented by $x$ and $y$. Clearly symmetry comes with uniqueness: if there is only one representation of a given lattice, the vector properties must be determined by the lattice.

Instead of this two dimensional representation then, we propose an efficient symmetric representation suitable for any partial ordering, in which dimensions correspond to \emph{chains} or \emph{totally ordered subsets} of the partial order. For taxonomies which are trees this representation is unique up to isomorphism; this more efficient representation can then be used in place of the vector ideal completion.

\begin{defn}[Chains]\index{chain|textbf}
Let $S$ be a partially ordered set. A \emph{chain} $C$ of $S$ is a totally ordered subset of $S$, that is, a subset of $S$ which is a partially ordered set under the partial ordering of $S$ such that $x \le y$ or $y \le x$ for all $x,y \in C$.

A collection of chains $\mathcal{C}$ is called \emph{covering} if $\bigcup_{C\in \mathcal{C}} C = S$. Clearly every partially ordered set has at least one covering collection of chains: that collection consisting of all chains containing just one node of $S$.
\end{defn}

\begin{defn}[Chain completion]\index{vector lattice completion!chain completion}
\newcommand{\Ch}{\mathrm{Ch}_\mathcal{C}}
Let $S$ be a real valued taxonomy and $\mathcal{C} = \{C_1, C_2, \ldots C_n\}$ be a covering collection of chains for $S$. Let $\Ch(x) = \{i : x \in C_i\}.$ Then define the function $\xi_0$ from $S$ to $\R^n$ by
$$\xi_0(x) =  \sum_{i \in \Ch(x)} \frac{p(x)}{|\Ch(x)|}e_i,$$
where $e_i$ are the basis elements of $\R^n$. Then the chain completion $\xi$ is defined by:
$$\xi(x) = \sum_{y \le x} \xi_0(y).$$
\end{defn}
\begin{prop}
The function $\xi$ defines a vector lattice completion of $S$ satisfying $\|\xi(x)\|_1 = \hat{p}(x)$.
\end{prop}

\begin{proof}
By the definition of $\xi$ it is clear that $u \le v$ in $S$ implies $\xi(u) \le \xi (v)$ in $\R^n$. Conversely, if it is not true that $u \le v$ then there will be some chain in $\mathcal{C}$ containing $v$ but not $u$, so it will never be true that $\xi(u) \le \xi(v)$, showing that $\xi$ defines an embedding of the partial ordering of $S$, and since $\R^n$ is a vector lattice, it also defines a vector lattice completion. Finally, note that
$$\|\xi(x)\|_1 = \sum_{y\le x} \|\xi_0(x)\|_1 = \sum_{y \le x} p(x) = \hat{p}(x)$$
since all the vectors are positive, which completes the proof.
\end{proof}

Providing we can find a covering with a low number of chains $n$, the previous proposition gives us an efficient vector lattice representation using $n$ dimensions. The representation as it stands is not unique, since there are in general many ways we can cover a partially ordered set with chains. The task then, is to find an efficient, unique way of determining a covering collection of chains. We achieve this by considering \emph{maximal chains}, chains containing as many elements of $S$ as possible whilst remaining a totally ordered set:
\begin{defn}[Maximal chains]
A maximal chain $C$ for $S$ is a chain such that there is no element $x$ of $S - C$ such that if $x$ were added to $C$ then $C$ would remain a chain. Let $\mathcal{C}$ be the covering collection of chains consisting of all maximal chains of $S$. $S$ is said to be \emph{uniquely minimally covered} by $\mathcal{C}$ if for each $C \in \mathcal{C}$ there is at least one element $x \in C$ such that $x$ is not in any other chain of $\mathcal{C}$; in this case, $S$ is said to possess a \emph{unique minimal covering} $\mathcal{C}$.
\end{defn}

\begin{prop}
If $S$ has a unique minimal covering $\mathcal{C}$, then $\mathcal{C}$ is a covering for $S$ with the least possible number of chains.
\end{prop}
\begin{proof}
We will assume there is a covering $\mathcal{C}'$ with less chains than $\mathcal{C}$ and show a contradiction. We can convert every chain $C$ in $\mathcal{C}'$ into a maximal chain by adding elements to $C$ until we can no longer add any more. The resulting collection cannot contain all maximal chains (since $\mathcal{C}'$ was assumed to have less chains than $\mathcal{C}$). Each missing maximal chain must contain some element not in any other maximal chain, which must also have been missing from $\mathcal{C}'$. Thus $\mathcal{C}'$ cannot have been a covering collection, which shows a contradiction and completes the proof.
\end{proof}

Thus if $S$ has a unique minimal covering, we can represent it uniquely and efficiently using the number of dimensions corresponding to the number of chains in this covering. Any taxonomy that is a tree\index{tree} has a unique minimal covering: each maximal chain will have a leaf node that is not in any chain; in fact there will be a chain corresponding to each leaf node. Thus the chain completion gives a unique efficient representation for any taxonomy that is a tree, and we would expect taxonomies that are very tree-like to also have efficient representations.


\subsection{Analysis of Application to Ontologies}

While we know that the chain completion is a relatively efficient for trees, we don't know how useful it is likely to be in real-world applications. To find out, we analysed two real world ontologies. The first is the Semantic Network\index{Semantic Network} used in the Unified Medical Language System\index{Unified Medical Language System} \citep{National:98}, whose taxonomy consists of just 135 nodes representing broad categories of meanings related to medical concepts. In this case, the taxonomy has a simple tree structure, so each dimension corresponds to a leaf node. There are 90 leaf nodes, thus we can represent the 135 nodes using only 90 dimensions, a saving of a third.

It is also instructive here to consider a simple theoretical situation: a regular tree\index{tree!regular} of depth $n$ with each node having $r$ branches. In this case, the total number of nodes is
$$\sum_{i=1\ldots n} r^i = \frac{r^{n+1} - 1}{r - 1} - r \simeq \frac{r^{n+1}}{r - 1}$$
where the approximation is for large $n$ and $r > 1$. The number of leaf nodes is $r^n$, thus in this approximation the ratio of leaf nodes to the total number of nodes will be $r^n (r-1)/r^{n+1} = (r-1)/r$. Thus the saving in the chain completion is greatest for low $r$: in a binary tree, half the nodes will be concentrated in the leaf nodes. The semantic network we considered above has a saving corresponding to $r = 3$.

The second taxonomy we considered was that of WordNet\index{WordNet} \citep{Fellbaum:98}. This is a very different situation to that just considered, having a much greater number of nodes, and no tree structure --- quite a large number of nodes have more than one parent. We looked at a subset of around 43,000 nodes using the hypernymy relation of nouns only; each node corresponds to a ``synset'' or concept corresponding to senses of terms in WordNet. We found a covering collection of chains using around 35,000 chains: a saving in terms of dimensionality of around 20\%. This does not give a unique representation however, and thus potentially suffers from some of the same problems as the two dimensional representations. The total number of maximal chains was around 60,000, meaning the unique chain-based representation would be less efficient than the straightforward vector lattice completion in which each dimension corresponds to a node.

It seems that chain-based representations are able to provide modest improvements in the efficiency of vector lattice representations, especially in the case of taxonomies with a tree structure. It is our hope, however, that techniques such as dimensionality reduction will eventually provide a means to find much more efficient representations, as long as it is possible to find good quality approximations which retain as much structure as possible of the original vector lattice.

\index{vector lattice completion!efficient|)}




\section{Representing Ambiguous Terms}
\label{representing-words-section}

So far we have only really considered representing concepts, or \emph{senses} of terms; we have not been concerned yet with how to represent terms themselves, which may be ambiguous with meanings covering many senses. For example, we view the structure of WordNet\index{WordNet}, which describes senses of terms, as a partial ordering, or as elements of a vector lattice. If we want to combine the vector lattice representations of the senses of a term to form something representing the ambiguous meaning, what is the correct way to do this?

Context-theoretic techniques provide an answer: if we look at the most straightforward model of context, the representation of a term is given by the vector sum of the representations of its contexts. This can easily be seen by considering the model of context discussed in the first chapter: if we add sense tags to the terms occurring in a corpus, then look at the vector representations of the individual senses of a term, since the vector representation is formed linearly, summing these representations will give us the same vector as that arrived at by looking at occurrences of the term without sense tags. This also makes sense from a probabilistic perspective; the probability of the occurrence of a term in a corpus is the sum of the probability of the occurrences of its senses, and this property is carried over in the $L^1$ norm of the corresponding vector representations. Looking at the lattice structure, this construction behaves as we would expect: each sense of a term entails the term itself. Thus if a term $w$ has $n$ senses $s_1, s_2, \ldots s_n \in S$, then the context vector of $w$ would be
$$\hat{w} = \sum_{i=1}^n \hat{s}_i$$
where $\hat{s}_i$ is the context vector of sense $s_i$.

When it comes to making use of vector representations of taxonomies, however, we run into a problem. We have constructed our vectors so that the $L^1$ norm corresponds to the probability of the \emph{concept}, which depends on the taxonomic structure. According to the context-theoretic philosophy, the representation of a term should be constructed linearly from the representations of its senses, however the probability of the occurrence of a \emph{sense} does not coincide with the probability of a concept. For example, the meaning of the word ``entity'' corresponds to the most general concept in some taxonomies, and thus the probability of the concept \emph{entity} is 1. However the word itself occurs fairly rarely in corpora, and we would expect it to have a fairly low probability even with respect to terms representing much more general concepts.

Looking at the situation from a context-theoretic perspective helps us to find an answer. We can view each node in the taxonomy as a context that terms can occur in. In the ideal vector completion a concept $s$ is represented as a sum over basis vectors corresponding to the nodes representing concepts at least as general as $s$. When $s$ is the sense of a term, we view the term as occurring in sense $s$ in contexts corresponding to the concepts at least as general as $s$. We may know the probability of the sense, but we have no way to distribute this probability over the hypothetical contexts.

One way of getting around this problem is to renormalise the vectors representing the individual senses $s_i$ and scale them according to the probability $\pi_i$ that the term $w$ occurs in sense $s_i$ (so that $\sum_i \pi_i$ is equal to the probability of term $w$ occurring):
$$\bar{w} = \sum_{i=1}^n \frac{\pi_i}{\|\bar{s}_i\|_1}\bar{s}_i$$


Thus we have a plausible way of representing terms as vectors. If we are to make use of these representations as part of a context theory, however, we have to be able to consider them as elements of an algebra. We have already seen the use of projections to represent lattice structures in the previous chapter, and again it is an algebra formed from projections that we will use to represent meanings of words within the setting of a context theory. In fact, as we will show, work in measures of distributional similarity supports the idea of representing meanings as projections.

\subsection{Distributional Similarity and Projections}
\label{dist-sim-projections-section}
\index{distributional similarity|(}

The work of \cite{Lee:99} analyses distributional similarity measures with respect to the \emph{support} of the underlying distribution. Let $f_t(c)$ denote the observed frequency of term $t$ occurring in context $c$. The support $S(t)$ of $f_t$ is the set of contexts $c$ for which $f_t(c)$ is non-zero;
$$S(t) = \{c \in C : f_t(c) > 0\}$$
where $C$ denotes the set of possible contexts that terms may occur in, or the feature space. According to our previous analysis, we consider the function $f_t$ as a vector in the space $L^\infty(C)$.

Lee considers measures of the degree of similarity between two terms $u$ and $v$. She shows that the three best performing measures (which include the $L^1$ norm, $\|f_u - f_v\|_1$) all depend only on the behaviour of the functions $f_u$ and $f_v$ on the intersection of the supports of the two terms, $S(u,v) = S(u) \cap S(v)$. Those measures which placed emphasis on the behaviour of the functions outside of this set, such as the $L^2$ norm, generally performed poorly in comparison.

\cite{Weeds:03}\index{Weeds, Julie} takes this analysis further, considering different functions $D(t,c)$ measuring the degree of association between a term $t$ and context $c$. The support with respect to $D$ is defined as $S_D(t) = \{c \in C : D(t,c) > 0\}$. She then considers the \emph{precision} according to an ``additive model'' defined in terms of $D$:
$$\mathcal{P}^\textrm{add}(u,v) = \frac{\sum_{c \in S_D(u,v)} D(u,c)}{\sum_{c\in S_D(u)} D(u,c)};$$
\emph{recall} can then be defined as the dual of precision, $\mathcal{R}^\textrm{add}(u,v) = \mathcal{P}^\textrm{add}(v,u)$. Weeds goes on to show how a general framework to describe distributional similarity measures can be described in terms of measures of precision and recall, and evaluates a range of measures within her framework. The best performing measure made use of the additive model of precision and recall together with a mutual information based function for $D$.

The details of Weeds' analysis are not so relevant for us; what is important to note is that in Weeds' additive model there is a move away from considering terms merely as vectors, and that this move is experimentally successful. What we will show is that we can view the additive model as representing terms as \emph{projections}, special kinds of operators on a vector space.

The vector space we are considering is given by the set $C$ of contexts that terms may occur in; we denote it $L^\infty(C)$; each element $c$ of $C$ has a corresponding basis element $e_c \in L^\infty(C)$. Given a subset of contexts $X$, $X\subseteq C$, we can view the vector space $L^\infty(X)$ as a subspace of $L^\infty(C)$. This subspace defines a projection $P_X$ on $L^\infty(C)$.

To specify this in more detail, consider a vector $f$ defined on $L^\infty(C)$ in terms of its components $\alpha_c$, where $f = \sum_{c\in C} \alpha_c e_c.$
The effect of the projection $P_X$ is then defined as follows:
$$P_Xf = \sum_{c \in X} \alpha_c e_c.$$
Given two subsets $X$ and $Y$ of $C$, it is easy to see that $P_XP_Y = P_{X \cap Y}$, thus the projection encodes set-theoretic behaviour. Since the definitions of precision and recall depend on the intersection of supports, we can translate these definitions into ones based on projections:
$$\mathcal{P^\textrm{add}}(u,v) = \|P_uP_v\Omega_D(u)\|_1,$$
where $P_t = P_{S_D(t)}$ and $\Omega_D(u)$ is a vector in $L^\infty(C)$ given in terms of its components by
$$\Omega_D(u) = \frac{1}{\sum_{c\in C} D(u,c)}\sum_{c \in C}D(u,c)e_c.$$

This representation comes close to providing us with a context theory; words can be represented as operators on a vector lattice and thus are elements of an algebra; the difference is that there is not a unique linear functional under consideration, the linear functional (which depends on $\Omega_D(u)$) is different depending on what element we are considering precision with respect to. The preceding analysis does however, point to the representation of meanings as projections on a vector lattice; we will show how such representations allow us to combine representations of concepts to form representations of the meanings of words.

\index{distributional similarity|)}

\subsection{Combining Concept Projections}
\label{ideal-projection}

So far we have not discussed the relationship between vector lattice completions and the context-theoretic framework itself. The previous completions we have discussed cannot be considered as context theories since they deal only with the vector lattice structure: there is no definition of multiplication on this space. In general, when discussing taxonomies the concept of multiplication on the vector space is not relevant, however there may be situations where it is useful to be able to define multiplication. For example, we may wish to make use of ontological representations as part of a larger context theory, in which case it helps to have a description of ontologies within the context-theoretic framework.

In this section we will show how terms can be represented within the context-theoretic framework as projections on a vector space. First we show how concepts in a taxonomy can be represented in terms of projections together with a linear functional.

\begin{defn}[Ideal Projection Completion]\index{vector lattice completion!ideal projection completion} If $S$ is a probabilistic taxonomy with probability distribution function $p \in L^\infty(S)$, then the \emph{ideal projection} $P_x$ associated with $x\in S$ is the projection $P_{\down{x}}$ on the space $L^\infty(S)$. We define a linear functional $\phi$ on the space of operators on $L^\infty(S)$ by
$$\phi(A) = \|(Ap)^+\|_1 - \|(Ap)^-\|_1,$$
\end{defn}
\begin{prop}
The ideal projection completion defines a vector lattice completion for $S$, such that $\phi(P_x) = \hat{p}(x)$.
\end{prop}

\begin{proof} There is clearly a lattice isomorphism between the ideal completion representation $\down{x}$ of $x\in S$ and the projection $P_x$; for example
$$P_xP_y = P_{\down{x}\cap\down{y}}.$$
Then note that
$\phi(P_x) = \|P_xp\|_1 = \sum_{y \in \down{x}}p(y) = \hat{p}(x).$
\end{proof}


The ideal projection completion can in fact be used to define a context theory for an alphabet $A$ if we have a way of associating elements of $A$ with concepts in $S$; for example $A$ may be a set of terms and $S$ a taxonomy of their meanings. If the words are unambiguous they will be associated with just one concept in $S$. Thus we can associate with each term a projection on $L^\infty(S)$.

Following the reasoning of previous sections, we can sum these projections to obtain representations of ambiguous terms. If a term $w$ has $n$ senses $s_1, s_2, \ldots s_n \in S$, and the term $w$ occurs in the sense $s_i$ with probability $\pi_i$, then we can represent $w$ as a probabilistic sum of the projection representation of its senses:
$$\bar{w} = \sum_{i = 1}^n \frac{\pi_i}{\phi(P_{s_i})} P_{s_i},$$
where $\bar{w}$ is the representation of $w$ as an operator on $L^\infty(S)$. The factor $\pi_i/\phi(P_{s_i})$ ensures that $\phi(\bar{w})$ is equal to the probability of term $w$; it can be interpreted as the conditional probability that $w$ occurs in sense $s_i$ given that some term has occurred in some sense $t$ at least as general as $s_i$, that is $s_i \le t$.

Because we represent terms as operators, in addition to the usual lattice operations, which work in a similar way to the ideal vector completion, multiplication is also defined on the representations. We can think of the probabilistic sum of senses as representing our uncertainty about the meaning of a term. The product of two terms then, would represent our uncertainty about the conjunction of their meanings. For example, if we approximate the meaning\footnote{Meanings are based on Wordnet definitions \citep{Fellbaum:98}; probabilities are invented.} of the word \emph{line} by
$$\bar{w}_l = \tfrac{3}{10}P_{l_1} + \tfrac{1}{10}P_{l_2}$$
where $l_1$ represents the sense ``a formation of people or things one beside another'' and $l_2$ represents the sense ``a mark that is long relative to its width'', and the word \emph{mark} by
$$\bar{w}_m = \tfrac{1}{5}P_{m_1} + \tfrac{1}{10}P_{m_2}$$
where $m_1$ represents the sense ``grade or score'' and $m_2$ represents the sense ``a visible indication made on a surface'', then the product is given by
$$\hat{w}_l\hat{w}_m = \tfrac{3}{50}P_{l_1}P_{m_1} + \tfrac{3}{100}P_{l_1}P_{m_2} + \tfrac{1}{50}P_{l_2}P_{m_1} + \tfrac{1}{100}P_{l_2}P_{m_2} .$$
If we further assume that the meanings of senses are disjoint, except for those referring to the sense ``a mark that is long relative to its width'' and the sense ``a visible indication made on a surface''; that is we assume $P_xP_y = 0$ unless $x = l_2$ and $y = m_2$ or vice versa, in which case $P_{l_2}P_{m_2} = P_{l_2}$ since a line is a type of mark. Then $\hat{w}_l\hat{w}_m = \tfrac{1}{100}P_{l_2}$; the product has disambiguated the meaning of both words.

\section{Conclusions and Future Work}

In this chapter we have discussed ways to represent taxonomic structure in terms of vector lattices. We have given several constructions with various properties, enabling probabilistic information to be incorporated into the vector lattice, allowing distances between concepts to be preserved, and reducing the number of dimensions needed for a representation. We also discussed ways in which ambiguous terms may be represented in terms of the vectors representing concepts, and gave a construction for which multiplication is defined, giving us a context theory.

The ideas of this chapter give plenty of potential for future work. The constructions suggest new measures of semantic distance: for example, the $l^p$ norm could be used together with such representations as a distance measure. The representation also gives us a way to measure semantic distance between ambiguous terms, something that may prove useful in applications.

There may also be ways that the techniques of this chapter can be used to help build taxonomies automatically, by looking for correlation between semantic distance and measures of distributional similarity and using this to place concepts in the vector lattice, and hence in the taxonomies.

 \index{taxonomy|)}




 
%
%
%
%
%
%
%



%

\chapter{Context Theories and Syntax} 
\label{syntax-chapter}

In this chapter we look at ways of describing syntactic properties of language in terms of vector space operators and algebra. This will allow us to incorporate such properties into context theories for natural language. The ability to view syntax from a context-theoretic perspective has many potential benefits, for example, we describe a method to represent syntax in terms of matrices that may lead to fast computational methods for statistical parsing, and at the end of the chapter we describe some ideas for how separate context theories for syntax and semantics may be combined using a generalisation of the notion of independence to create a new form of natural language semantics in which both the semantic and syntactic aspects of a word may be represented as a single element of an algebra.


The context-theoretic framework places specific requirements on the nature of an implementation; these mean that certain grammar formalisms are more suited to the context-theoretic approach. We can identify several properties of the framework that are relevant:
\begin{itemize}
\item the framework requires that all information about the properties of a word are incorporated into its vector representation. This leads us to lexicalised formalisms for syntax in which syntactic properties of a word can be encapsulated independently of any external grammar. This makes a generative grammar less attractive for example, since the properties of a word are spread throughout the rules of the grammar, whereas categorial grammar can encapsulate the syntactic properties of a word purely by specifying its category.
\item the algebra of a context theory must be associative: $(ab)c = a(bc)$ for all $a,b$ and $c$ in the algebra, thus the grammatical formalism should be compatible with this idea.
\end{itemize}
Consideration of these properties has led us to two syntactic formalisms that are particularly suited to the context-theoretic approach, namely categorial grammars and link grammar. It is likely that other syntactic formalisms can also be described in terms of context theories, however we have concentrated on these two since they have the above properties and thus promise to be closest to the context-theoretic approach.

The contributions of this chapter are as folows:
\begin{itemize}
\item In Section \ref{syntax-background-section} we summarise various forms of categorial grammar including those that are algebraic in nature. We describe Bar-Hillel's and Lambek's formulations, bilinear logic and pregroups; in Section \ref{categorial-context} we discuss the relationship between categorial grammar formalisms and the context-theoretic framework, showing how the Lambek calculus can be incorporated into a context theory, and explaining why the other formalisms are not so suited to the framework.
\item The context-theoretic description of the Lambek calculus is difficult to handle: it is not obvious how to compute with the resulting representation. We have found that link grammar (introduced in Section \ref{link-grammar-section}) can be described as context theories in ways that do not have this limitation:
\begin{itemize}
\item We give a description of link grammar in terms of operators on an infinite dimensional vector space called Fock space in Section \ref{operator-formulation-section}. This gives us a new description of a simple form of stochastic link grammar (Section \ref{stochastic-link-grammar-section}) and enables us to describe link grammars in terms of matrices (Section \ref{lg-matrix-section}).
\item We give a context theory for link grammar in terms of semigroups in Sections \ref{alg-lg-section} to \ref{semigroup-context}. This brings to light a useful relationship between link grammar and inverse semigroups, allowing us to describe a link grammar parse as the Munn tree of a free inverse semigroup; this may ultimately have the potential to incorporate semantic information into the representation. This section also demonstrates the usefulness of using semigroups to construct context theories: a context theory can be tailored by building the required properties into a semigroup.
\end{itemize}
\item We discuss potential directions for future research in Section \ref{discussion-syntax-section}.
\end{itemize}
It is our hope that the contributions of this chapter will lead to new ways of combining vector representations of words to form representations of larger constituents in a way that incorporates syntactic structure, allowing complex vector-based representations of meaning to be built up from smaller ones. Such approaches could form a useful alternative to logic-based representations of meaning.

\section{Categorial Grammars}
\label{syntax-background-section}
\index{categorial grammar|(}


\subsection{Bar-Hillel Categorial Grammar}
\index{Bar-Hillel categorial grammar}

The simplest form of categorial grammar is due to \citeauthor{Bar-Hillel:50} (\citeyear{Bar-Hillel:50};\citeyear{Bar-Hillel:64} 1964) (based on earlier work of Ajdukiewicz) and is described as a deductive system with the following rewrite rules:
\begin{eqnarray*}
(A/B)\ B &\rightarrow& A\\
B\ (B\backslash A) &\rightarrow& A
\end{eqnarray*}
In a categorial grammar, words in a language are assigned one or more \emph{categories}, built up out of a number of \emph{basic types} and the operations $/$ and $\backslash$. For example, a transitive verb might be assigned the category $(\mathit{NP}\backslash S)/\mathit{NP}$, where $\mathit{NP}$ and $S$ are basic types representing the categories of noun phrases and sentences respectively. The category $(\mathit{NP}\backslash S)/\mathit{NP}$ can be thought of as describing those strings which form a sentence when they are both preceded and followed by a noun phrase.

\subsection{Lambek Calculus}
\index{Lambek calculus|(}

Based on Bar-Hillel's categorial grammar, \cite{Lambek:58}\index{Lambek,  Joachim} developed a calculus specifically for describing natural language. In its original form, it is defined as a deductive system, whose axioms\footnote{See also \cite{Wood:93}.} are:
\begin{eqnarray*}
A & \rightarrow & A\\
(AB)C & \leftrightarrow  & A(BC),
\end{eqnarray*}
where $A \leftrightarrow B$ is shorthand for $A\rightarrow B$ and $B\rightarrow A$, with the following rules of inference:
\begin{eqnarray*}
AB \rightarrow C & \mathrm{iff} & A\rightarrow C/B\\
AB \rightarrow C & \mathrm{iff} & B\rightarrow A\backslash C,
\end{eqnarray*}
and
$$\mathrm{if}\ A\rightarrow B\ \mathrm{and}\  B\rightarrow C\ \mathrm{then}\ A\rightarrow C$$
Using these rules, it possible to deduce many theorems of the calculus, for example
$$\begin{aligned}
(A/B)\ B \rightarrow A& &\quad &\text{(Ajdukiewicz's law)}\\
A \rightarrow (B/A)\backslash B & & &\text{(Type raising)}\\
(A/B)(B/C) \rightarrow A/C & & &\text{(Composition)}
\end{aligned}$$
and their equivalents with $/$ exchanged with $\backslash$; many of these are useful in describing features of natural language.

One way of modeling the Lambek calculus is with free semigroups\index{free semigroup!and Lambek calculus} (also called \emph{L-models}) --- the completeness of the Lambek calculus with respect to such models is described in \cite{Pentus:95}. The calculus can be viewed as operations on subsets  of a monoid $M$, with
\begin{eqnarray*}
XY &=& \{xy : a \in X, b \in Y\} \\
X \backslash Y &=& \{m \in M : Xm \subseteq Y\} \\
Y / X &=& \{m \in M : mX \subseteq Y\}
\end{eqnarray*}
where $X,Y\subseteq M$ and we also use $m$ as a shorthand for $\{m\}$.

More generally, the operations $/$ and $\backslash$ can be defined for certain semigroups called \emph{residuated lattices} \citep{Birkhoff:48}. The connection between the Lambek calculus and residuated lattices was noted in Lambek's original paper \citep{Lambek:58}.

\begin{defn}[Partially Ordered Semigroup]\index{semigroup!partially ordered|textbf}
A semigroup $S$ together with a partial ordering $\le$ is called \emph{partially ordered} if $x \le y$ implies $xz \le yz$ for all $x,y,z \in S$.
\end{defn}

\begin{defn}[Lattice Ordered Semigroup]\index{semigroup!lattice ordered|textbf}
A lattice ordered semigroup is a partially ordered semigroup $S$ in which the partial ordering defines a lattice with operations $\lor$ and $\land$ such that
\begin{eqnarray*}
x\cdot (y\lor z) &=& x\cdot y\  \lor\ x\cdot z\\
(y\lor z)\cdot x &=& y\cdot x\ \lor\ z\cdot x
\end{eqnarray*}
\end{defn}

\begin{defn}[Residuated Lattice]\index{residuated lattice|textbf}
A lattice ordered semigroup $S$ is called a residuated lattice, if for each $x,y \in S$ there exists a greatest element $x/y$ such that
$$x/y \cdot y \le x$$
and a greatest element $x\backslash y$ such that
$$y \cdot y\backslash x \le x.$$
The elements $x/y$ and $y\backslash x$ are called the right and left \emph{residuals} or \emph{quotients}.
\end{defn}

As \cite{Birkhoff:48} notes, if $S$ has a zero which is also the least element of the lattice then the residuation operations $/$ and $\backslash$ can be defined by
\begin{eqnarray*}
x/y &=& \bigvee\{z : zy \le x\}\\
y\backslash x &=& \bigvee\{z : yz \le x\}
\end{eqnarray*}

The notion of residuated lattice is useful for our purposes because it allows us to think of categorial grammar in purely algebraic terms, allowing us to see how it relates to the context theoretic framework, and how it compares to other algebraic approaches.

\index{Lambek calculus|)}

\subsection{Bilinear Logic}
\index{bilinear logic}

\cite{Lambek:93}\index{Lambek,  Joachim} and \cite{Abrusci:91}\index{Abrusci, V.~M.}, based on earlier work of \cite{Girard:87}\index{Girard, J.~Y.}, developed a new version of Lambek's calculus called \emph{(classical) bilinear logic}. This adds two constants, 1 (introduced at an earlier stage by Lambek) and 0, to Lambek's original definition, which satisfy
\begin{gather*}
1A \leftrightarrow A \leftrightarrow A1\\
(0/A)\backslash 0 \leftrightarrow A \leftrightarrow 0/(A\backslash 0) 
\end{gather*}
As a shorthand notation, $A\backslash 0$ is written $A^r$ and $0/A$ is written $A^l$. It can be shown that
$$(B^rA^r)^l \leftrightarrow  (B^lA^l)^r$$
which is written as $(A\oplus B)$. Some theorems of bilinear logic \citep{Casadio:02} are
\begin{gather*}
1^r \leftrightarrow 0 \leftrightarrow 1^l\\
A\oplus 0 \leftrightarrow A \leftrightarrow 0\oplus A\\
(A\oplus B)\oplus C \leftrightarrow A\oplus(B\oplus C)\\
\begin{aligned}
A^lA \rightarrow 0& &\quad &AA^r \rightarrow 0\\
1 \rightarrow A\oplus A^l& & &1 \rightarrow A^r\oplus A\\
A/B  \leftrightarrow A\oplus B^l& & &B \backslash A \leftrightarrow B^r \oplus A\\
(A\oplus B) C  \rightarrow A\oplus BC& & &C(A\oplus B)  \rightarrow CA\oplus B
\end{aligned}
\end{gather*}

\subsection{Pregroups}

Pregroups \citep{Lambek:01} arose as a simplification of bilinear logic called \emph{compact bilinear logic}, in which it is additionally assumed that $0 \leftrightarrow 1$ and $AB \leftrightarrow A\oplus B$. In this case there is a simpler description in terms of partially ordered monoids:

\begin{defn}[Pregroup]\index{pregroup|textbf}
Let $S$ be a partially ordered monoid. Then $S$ is called a pregroup if for each $x\in S$ there are elements $x^l$ and $x^r$ in $S$ such that
\begin{eqnarray*}
x^lx \le &1& \le xx^l\\
xx^r \le &1& \le x^rx
\end{eqnarray*}
\end{defn}





\subsection{Categorial Grammar and Context Theories}
\label{categorial-context}
\index{categorial grammar!and context theories|(}

We would like to be able to describe the syntactic formalisms we have discussed within the context-theoretic framework; firstly to demonstrate the generality of the framework, and secondly, because we hope new techniques in parsing and semantic representation to arise by doing so. When it comes to categorial grammars, we seem to be well-placed since there are algebraic interpretations of many versions of the formalism. However, on closer inspection, making direct use of these formalisms within the context theoretic framework appears difficult.

For example, if we want to make use of a residuated lattice\index{residuated lattice} $S$, we could try and represent the structure within a lattice ordered algebra\index{lattice ordered algebra}. Like any semigroup, the vector space $L^1(S)$ can be considered as a lattice ordered algebra (see Section \ref{algebras}). However, the lattice ordering of $L^1(S)$ is not connected to the lattice ordering of $S$. If we wished to connect them, we may try to use one of the constructions described in the previous chapter to embed partial orderings within vector lattices. However, then it is not clear how we are to define multiplication on the vector lattice in a way that is consistent with multiplication in $S$.

We face similar problems with pregroups:\index{pregroup} it is not clear how we can incorporate the pregroup partial order into a vector lattice partial order whilst maintaining the multiplication defined in the pregroup.

Bilinear logic\index{bilinear logic} appears closer to being a vector space with an ``addition'' operation, $\oplus$, however, this operation is not defined to be commutative, something which is essential for a vector space. Requiring $\oplus$ to be commutative results in multiplication also being commutative, something not generally desirable for describing natural language syntax.

There is one way to represent categorial grammars within the framework however: we can make use of free semigroup\index{free semigroup} models to describe the Lambek calculus. Instead of using subsets of a free monoid $A^*$, however, we use elements of the algebra $L^\infty(A^*)$. A set $X \subset A^*$ is represented as the element $\tilde{X} \in L^\infty(A^*)$:
$$\tilde{X}(z) = \begin{cases}
1 & \text{if } z\in X\\
0 & \text{otherwise,}
\end{cases}$$
for $z \in A^*$. Multiplication in this algebra is defined by multiplication of the underlying free monoid, while vector space and lattice operations are defined since $L^\infty(A^*)$ is a vector lattice. We are thus able to represent the syntactic properties of a word by taking weighted sums of the representation of its syntactic categories, with weights corresponding to the probability that a word will take the respective category.

We can use this idea to make a context theory if we define a linear functional $\phi$ on $L^\infty(A^*)$ by
$$\phi(u) = \sum_{x \in A^*} p(x)u(x)$$
where $p$ is a probability distribution over elements of $A^*$. In this way, the context-theoretic probability of a category is the sum of the probabilities of all the strings in that category.

This representation raises computational issues similar to the ones that arose in dealing with logical semantics in Section \ref{practical-issues}; and a similar solution can be used. The problem again is that a word may be represented as a sum of categories whose vector representations are not disjoint in the vector lattice. The same method for computing a lower bound for the degree of entailment\index{entailment!degree of} between sentences can be used to estimate a degree of entailment between a desired parse and the syntactic representation of a sentence, or  to estimate a syntactic ``entailment'' between sentences.

Note that the algebra $L^1(A^*)$ is \emph{not} a residuated lattice\index{residuated lattice} under the vector lattice ordering, since it is not a lattice ordered semigroup under this ordering. The subsemigroup of elements of this algebra generated by the representation of categories does, however, form a lattice ordered semigroup under this ordering, and is also a residuated lattice, since it is isomorphic as a lattice ordered semigroup to the semigroup of subsets of the free monoid $A^*$. This means, that while we can represent categories within the algebra and take weighted sums of them, we cannot form new categories from these weighted sums --- something that is not a limitation for representing natural language syntax.


\index{categorial grammar!and context theories|)}
\index{categorial grammar|)}

\section{Link Grammar}
\label{link-grammar-section}
\index{link grammar|(}

Link grammar \citep{Sleator:91}\index{Sleator and Temperly} is a lexicalised syntactic formalism which describes properties of words in terms of \emph{links} formed between them, and which is context-free in terms of its generative power. Apart from determining which sequences are grammatical, the links also encapsulate the nature of the relationships between words.

As an example, a transitive verb in English may link (simultaneously) to a subject on the left and an object on the right. This is represented in link grammar as the \emph{disjunct} $\ket{s}\bra{o}$ where $s$ and $o$ stand for `subject' and `object' respectively.\footnote{We are introducing our own, quantum mechanical, notation for link grammars from the beginning so as to be consistent, however we will describe the intended interpretation of this notation later.}

\begin{defn}[Link Grammar]
Let $L$ be a set of \emph{link types}. Then we define a set of \emph{left connectors} $D_l(L) = \{\ket{x} : x \in L\}$ and a set of \emph{right connectors} $D_r(L) = \{\bra{x} : x \in L\}$.

A disjunct is an element of $D_l(L)^*D_r(L)^*$. That is, a disjunct consists of a string of left connectors $\ket{x_1}\ket{x_2}\ldots\ket{x_n}$ followed by a string of right connectors $\bra{y_1}\bra{y_2}\ldots\bra{y_m}$.

The syntactic representation of a word is a set of disjuncts, each one corresponding to a different syntactic r\^ole played by the word. A sequence of words is in the language generated by the grammar if there is a corresponding sequence of disjuncts and a set of arcs, or \emph{links} drawn above the disjuncts such that:
\begin{itemize}
\item each disjunct in the sequence is a disjunct of the corresponding word in the sequence of words;
\item each left connector is connected to a right connector of the same type at any position to the right of it by drawing a link from one to the other;
\item each connector in each disjunct in the sequence is connected to exactly one other connector;
\item no links cross.
\end{itemize}
\end{defn}

Table \ref{link-table} shows a fragment of a link grammar. The grammar is clearly highly simplified, and is presented merely to explain the concept; for example in our fragment, \emph{way} and \emph{mud} can only occur as objects. Link grammars generally include a special symbol called the `wall' to indicate the beginning of the sequence \citep{Sleator:91}, which is then included in the grammar, but again we have omitted this for simplicity.

\begin{table}
\begin{center}
\caption{A small link grammar.}
\begin{tabular}{lp{7cm}}
\hline\noalign{\smallskip}
\emph{word} & \emph{disjuncts}\\
\noalign{\smallskip}
\hline\hline
\noalign{\smallskip}
they	& $\bra{s}$\\
mashed & $\ket{s}\bra{o}\qquad \ket{s}\bra{m}\bra{o}$ \\
way, mud & $\ket{d}\ket{o}\qquad \ket{d}\ket{j}\qquad \ket{a}\ket{d}\ket{o}\qquad \ket{a}\ket{d}\ket{j}$ \\
their, the & $\bra{d}$\\
through & $\ket{m}\bra{j}$\\
thick & $\bra{a}$\\
\noalign{\smallskip}
\hline
\noalign{\smallskip}
Link types: &  $s$: subject, $o$: object, $m$: modifying phrases, \mbox{$a$: adjective}, $j$: preposition, $d$:~determiner.\\
\noalign{\smallskip}
\hline
\end{tabular}
\label{link-table}
\end{center}
\end{table}

\begin{figure}[b]
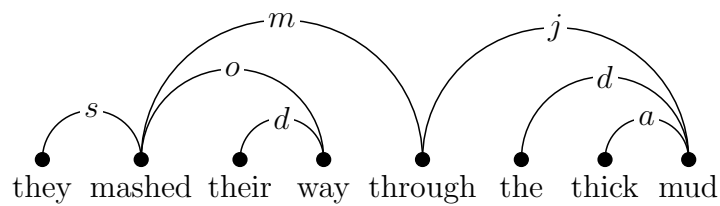

\begin{center}
\begin{graph}(10,4)(-1,-1)
	\graphlinecolour{1}
\newcommand{\ntext}[4]{\roundnode{#1}(#2,#3)[\graphlinecolour{0}]  \nodetext{#1}(0,-.5){\raisebox{0pt}[1pt][1pt]{#4}}}
\newcommand{\btext}[3]{\bow{#1}{#2}{0.5}[\graphlinecolour{0}] \bowtext{#1}{#2}{0.5}{#3}}
\ntext{They}{0}{0}{they}
\ntext{Mashed}{1.3}{0}{mashed}
\ntext{Their}{2.6}{0}{their}
\ntext{Way}{3.7}{0}{way}
\ntext{Through}{5}{0}{through}
\ntext{The}{6.3}{0}{the}
\ntext{Thick}{7.4}{0}{thick}
\ntext{Mud}{8.5}{0}{mud}
\btext{Thick}{Mud}{$a$}
\btext{The}{Mud}{$d$}
\btext{Through}{Mud}{$j$}
\btext{Their}{Way}{$d$}
\btext{Mashed}{Way}{$o$}
\btext{Mashed}{Through}{$m$}
\btext{They}{Mashed}{$s$}
\end{graph}
\caption{A link grammar parse.}
\label{parse}
\end{center}
\end{figure}

A parse for a sentence is drawn as a set of links above the sentence, as in Figure \ref{parse} for the sentence `they mashed their way through the thick mud'. The disjuncts that are used in the parse are not generally drawn, but can be inferred from the links drawn above the sentence.

An efficient parsing\index{parsers} algorithm for link grammar based on dynamic programming is described by \cite{Sleator:91}. Their link grammar for English can handle transitive, ditransitive and modal verbs; prepositions, adverbs, complex noun phrases and relative clauses; questions and question inversion; number agreement is also taken into account.

\subsection{Operator Formulation of Link Grammar}
\label{operator-formulation-section}

In this section we begin our description of link grammar in terms of operators on a vector space. The mathematics we will make use of is in fact derived from that of quantum mechanics: links are described as combinations of ``creation''\index{creation and annihilation operators} and ``annihilation'' operators referring to the creation and annihilation of a particle in a quantum mechanical\index{quantum mechanics} system.


The mathematics of quantum mechanics has proved useful in retrieval\index{retrieval} applications for removing unwanted components of meaning in a search query \citep{Widdows:03} on latent semantic analysis\index{latent semantic analysis} vectors.
Quantum mechanics deals with a kind of vector space that is particularly well behaved and frequently occurring, so called \emph{Hilbert space}\index{Hilbert space|(} (see Section \ref{completeness-section} for details).
We make use of a special kind of infinite dimensional Hilbert space called \emph{Fock space}\index{Fock space}. As we will show, we can describe syntactic properties of words in terms of link grammars as operators on such a space.

One immediate benefit of this discovery is an entirely new perspective on link grammars, which may open up research on this type of grammar. For example, we will show how this view of link grammars can be used to describe the grammar in terms of matrix\index{matrices} operations, opening up the possibility of (potentially very efficient) computational procedures for statistical parsing using matrices.






Our exposition is inspired by the study of \emph{free probability}\index{free probability} \citep{Voiculescu:97}, wherein the study of non-crossing diagrams is very closely connected to link grammars; our main result in this section is more or less a direct translation of a standard result in free probability theory.

Our syntactic vectors will reside in \emph{Fock Space}, a Hilbert space which is like the sum of an infinite series of Hilbert spaces.


Let $H$ be a finite dimensional complex Hilbert space and $\Omega$ a distinguished vector in $H$ with norm 1. The Fock space\index{Fock space|textbf} $\mathcal{F}$ of $H$ is then defined as
$$\mathcal{F} = \mathbb{C}\Omega \oplus H \oplus (H \otimes H) \oplus (H \otimes H \otimes H) \oplus \cdots$$
i.e.~it is the direct sum of all finite tensor product powers of $H$, where $\oplus$ denotes the direct sum and $\otimes$ the tensor product (see section \ref{new-vector-spaces}), and $\mathbb{C}\Omega$ is a one dimensional Hilbert space which is viewed as the zeroth power of $H$.


\index{creation and annihilation operators|(}
We are now able to form the connection between quantum mechanics and syntax.
In the physical interpretation of Fock space, different powers of the Hilbert space $H$ correspond to states of different numbers of particles. Special operators called \emph{creation operators} map states in $n$ powers of $H$ to states in $n+1$ powers of $H$, effectively `creating' an additional particle. Similarly, \emph{annihilation operators} reduce the number of powers of $H$ in a state by one, `annihilating' a particle. It is these operators that we will use to represent syntax.

Let $u$ be a vector in $H$. The creation operator $\ket{u}$ on $\mathcal{F}$ is defined such that
$$\ket{u} v_1\otimes v_2 \otimes \cdots \otimes v_n = u \otimes v_1\otimes v_2 \otimes \cdots \otimes v_n.$$
The \emph{dual} of $\ket{u}$ is the annihilation operator $\bra{u}$ and maps vectors according to:
$$\bra{u}v_1\otimes v_2 \otimes \cdots \otimes v_n = \inprod{u}{v_1}v_2 \otimes \cdots \otimes v_n$$
and $\bra{u}\Omega = 0$. The action of the operators on sums of tensor products can be deduced from their linearity.

The effect of `creating' and then `annihilating' is just a scalar product times the identity operator, 1:
$$\bracket{u}{v} = \inprod{u}{v}1;$$
the notation $\bracket{u}{v}$ is used whenever a creation operator follows an annihilation operator.

\subsection{Syntactic Interpretation}
\label{link-context-1}

In the syntactic interpretation of Fock space, the set of links $L$ are represented as a set of vectors $L_H$ which are assumed to form an \emph{orthonormal basis} for $H$. Disjuncts for words are then formed by concatenating creation and annihilation operators, in exactly the same way that left and right connectors are concatenated in link grammar. The representation of the syntactic characteristics of a word can then be represented by taking the sum of its disjuncts. For example the word \emph{mashed} in our simple link grammar in Table \ref{link} can be represented as the operator
$$\hat{\mathit{mashed}} = \ket{s}\bra{o} + \ket{s}\bra{m}\bra{o},$$
where we assume the vectors $s,o,m,a,j \in H$ form an orthonormal basis for $H$.

Our formulation will require that the link grammar parses are ``strict'' in the following sense: there must not be any connectors left unlinked; thus the parse must start with a right connector and end with a left connector.

In order to determine whether a sequence of words is in the language determined by the link grammar, we define a linear functional $\phi$ on $B(\mathcal{F})$ (the set of bounded linear operators on $\mathcal{F}$) by
$$\phi(\hat{a}) = \inprod{\Omega}{\hat{a}\Omega},$$
where $\hat{a} \in B(\mathcal{F})$.  We then have the following:
\begin{prop}
Let $W$ be a set of words, and $\Gamma$ a function that assigns a set of link grammar disjuncts to every word in $W$, with link types from a set $L$.

For every $w \in W$ we denote its corresponding Fock space operator $\hat{w}$ on the Fock space generated by the Hilbert space with basis vectors $L_H$  corresponding to the link types in $L$. Then $w_1w_2\ldots w_n$ is in the link grammar language defined by $\Gamma$ if and only if $\phi(\hat{s}) \ge 1$, where $s = \hat{w_1}\hat{w_2}\ldots\hat{w_n}$. $\phi(\hat{s})$ indicates the number of valid link grammar parses.
\end{prop}
\begin{proof}
Let us first assume each word has only one disjunct.

The product of an annihilation operator with a creation operator satisfies
$$\bracket{x}{y} = \left\{\begin{array}{ll}
0 & \textrm{if $x \neq y$}\\
1 & \textrm{if $x = y$}\end{array}\right.,$$
where $x, y \in L_H$. Thus any operator $\hat{s}$ which is given by a product of creation and annihilation operators reduces either to $0$, $1$, or a product of a (possibly empty) sequence of creation operators followed by a (possibly empty) sequence of annihilation operators. In the latter case, as in the case of $0$, $\phi(\hat{s})$ will be zero since if there are annihilation operators in the sequence their operation on $\Omega$ will give zero (they operate on $\Omega$ first as they are on the right), and if there are no annihilation operators the creation operators will operate on $\Omega$ to give a vector disjoint with $\Omega$.

If the sequence satisfies any of the following the product will be zero and the sentence will not parse:
\begin{itemize}
\item A left connector is not matched by a right connector; in this case the product of the corresponding operator will map $\Omega$ to a different dimension in the Fock space and $\phi(\hat{s})$ will be zero.
\item The left connector is matched by a right connector of a different type; in this case the product of the corresponding operators will be zero.
\item The connectors match but the corresponding links cross; in the case there will again be a product of the form $\bracket{x}{y}$ where $x\neq y$ and the product will be zero.
\end{itemize}
Conversely, $\phi(\hat{s})$ will be zero just in case one of the above conditions holds and thus the sentence will not parse.

On the other hand, if none of the above conditions are met the sentence must parse and if the parse is strict the corresponding operator must map $\Omega$ to itself, so $\phi(\hat{s}) = 1$.

If words are now allowed more than one disjunct, then since these are added as operators and distribute with respect to multiplication each possible parse will be a term in the resulting sum of disjuncts, and thus $\phi(\hat{s})$ will indicate the number of valid link grammar parses.
\end{proof}

Note that this representation defines a strong context theory: the original Hilbert\index{Hilbert space|)} space $H$ is a vector lattice under the ordering induced by the basis associated with the set of link types, and thus $\mathcal{F}$ is also a vector lattice since we can define a basis for it using the basis of $H$. Thus the space of operators on this space also form a vector lattice, as well as an algebra; specifically we are interested in the algebra $\mathcal{A}$ generated by creation and annihilation operators. Together with the linear functional $\phi$ and the translation from strings to operators, where we assume that the empty string translates to the identity operator, we have a context theory. Moreover, the subspace $I = \{u \in \mathcal{A} : \phi(u) = 0\}$ is a sub-vector lattice of $\mathcal{A}$ since it is the space formed from all linear combinations of sequences of creation and annihilation operators which do not map $\Omega$ onto itself, thus we have a strong context theory.

\index{creation and annihilation operators|)}

\subsection{Stochastic Link Grammar}
\label{stochastic-link-grammar-section}
\index{link grammar!stochastic}

In applications requiring robust parsing of natural language stochastic grammars are vital in order to help in dealing with the large number of parses, which in general for wide coverage parsers increases exponentially with sentence length \citep{Manning:99}.

In the case of our implementation of link grammar we are not restricted to using sums of the basis vectors $L_H$, but can take any linear combination of these vectors when constructing the grammar, enabling us to form a type of stochastic link grammar similar to the supertagging models of \cite{Bangalore:99}. The representation of a word would be a weighted sum of the representation of its disjuncts; the weight attached to each disjunct can be interpreted as the probability that the word occurs in that syntactic r\^ole. For products of words, the weights attached to disjuncts will in general sum to less than 1 since some disjuncts will have a product of zero; it is thus necessary to renormalise the weights after taking the product to account for disjuncts whose product is zero in order to interpret them as probabilities.

Probabilistic link grammars were described by \cite{Lafferty:92}\index{Lafferty et al.}, where the probability of each link occurring with a word is conditioned on several factors, including the words occurring on either side. Such a model provides a probability distribution over the language generated by the grammar. They showed their formalism to be a generalisation of \emph{trigrams}\index{trigrams} which have proved very successful in language modelling. Our formalism does not allow conditioning of the probability directly, as Lafferty et al's does, however this information can be incorporated by including extra links describing the features one wishes to condition the probability on, and weighting these links accordingly.



An advantage of this simpler formulation of stochastic link grammar in comparison to that of \cite{Lafferty:92} is that it allows an entirely lexicalised description of syntax: the grammar can be described by assigning each word its disjuncts and corresponding probabilities. The ultimate advantage however, we believe, will be in opening up new computational procedures for statistical parsing using matrices.

\subsection{Link Grammar and Matrices}
\label{lg-matrix-section}
\index{matrices!and link grammar|(}

The operators described in the previous section operate on an infinite-dimensional vector space --- something that is clearly difficult to implement. In practice, it may be possible to consider a finite-dimensional subspace of this vector space. This can be done by placing a limit on the number of left or right links that can be concatenated together. For example, we could use the subspace
$$\mathcal{F}_3 = \mathbb{C}\Omega \oplus H \oplus (H \otimes H) \oplus (H \otimes H \otimes H)$$
of the Fock space which is made up of $1 + n + n^2 + n^3$ dimensions, where $n$ is the number of dimensions of $H$. This would allow up to three left links and up to three right links to be concatenated. In general, allowing the concatenation of $k$ links would need $\sum_{i=0}^k n^i = \frac{n^{k+1} - 1}{n - 1}$ dimensions. 

The matrix representation of a link grammar can be built up using the standard definitions of tensor product and direct sum for matrices. For example, for a two dimensional vector space with basis vectors $a$ and $b$, for $k = 2$ we can assign the seven dimensions the following interpretations:
$$[\Omega, a, b, a\otimes a, a\otimes b, b\otimes a, b\otimes b]$$
The creation operator (left link) $\ket{a}$ would then have the matrix representation
$$\left(\begin{array}{ccccccc}
0&0&0&0&0&0&0\\
1&0&0&0&0&0&0\\
0&0&0&0&0&0&0\\
0&1&0&0&0&0&0\\
0&0&1&0&0&0&0\\
0&0&0&0&0&0&0\\
0&0&0&0&0&0&0
\end{array}\right)$$
since it maps $\Omega$ to $a$, $a$ to $a\otimes a$ and $b$ to $a\otimes b$. The corresponding annihilation operator $\bra{a}$ is represented by the matrix transpose of the representation of $\ket{a}$.

An important question to be addressed in future work is what the maximum number of concatenations is likely to be for a particular grammar and application; if this number is high the technique may become impractical because of the exponential increase in the number of required dimensions. One way to get around this problem may be to make use of a dimensionality reduction\index{dimensionality reduction}, such as that of random projections\index{random projections} \citep{Papadimitriou:98, Sahlgren:02}. In this technique, each basis vector in the original vector space is represented as a random vector in a new vector space of much lower dimensionality; this defines a transformation (a random projection) from the old vector space to the new. If the dimensionality of the new vector space is sufficiently high, it is highly likely that distances and scalar products between vectors will be preserved to within some threshold, however some further work is required to investigate the suitability of this technique for representing syntax.

\index{matrices!and link grammar|)}






\subsection{Parsing with Operators}
\index{parsers!and operators}

So far we have only really treated the problem of acceptance of a language defined by a link grammar: we can tell if a sentence is in the language, but we are left with no record of the parse itself. This is not very useful in applications, since we are normally interested in finding out the structure of the sentence. In order to determine this structure as we multiply the operator representations, we need to be able to keep a record of which disjunct was used with each word. This can be done by defining a new vector space $H_d$ of dimensionality $d$, where $d$ is the greatest number of disjuncts that any word has in the grammar. We then form the Fock space\index{Fock space} $\mathcal{F}_d$ of this vector space and take the tensor product with the original Fock space in which the link grammar is represented. We now alter our original operators so that they operate on the new space $\mathcal{F} \otimes \mathcal{F}_d$. If a word has the original representation $x_1 + x_2 + \ldots x_d$ where the $x_i$ are the representations of the individual disjuncts, then in the new representation it becomes
$$x_1\otimes \ket{e_1} + x_2 \otimes \ket{e_2} + \ldots x_d\otimes \ket{e_d},$$
where the $e_i$ are basis vectors for $H_d$.

As these representations are multiplied, the product will be a sum of disjuncts; the right hand side of each disjunct will be a product of creation operators, each specifying the number of the disjunct used in the corresponding word. Those disjuncts of a word which cannot be used to form sentences will have a product of zero, and thus will not feature in the sum; nor will their tensor product with $\mathcal{F}_d$, thus only those disjuncts that can be used to form valid sentences will be represented in the product.

\subsection{Algebraic Formulation of Link Grammars}
\label{alg-lg-section}
\label{link}

The vector space formulation of link grammar we have just described provides us with a way to describe syntax within a context theory; it has also provided us with a way of computing with link grammars using matrices. However we are interested in combining representations of syntax with representations of meaning, and the formulation just described does not seem to be ideally suited to this. Describing words as operators on Fock space\index{Fock space} would allow meanings of larger constituents to be built up using tensor products only in a limited fashion: Fock space vectors work like a stack, and vectors can only be ``pushed'' or ``popped'' on this stack.

If we can describe syntax in algebraic terms, specifically in terms of semigroups, then we will be on much stronger ground because of the tools available for combining such representations. In particular, free inverse semigroups allow the representation of trees in algebraic terms. As we will see, we will not lose the flexibility of vector space representations; the vector space nature will be regained by considering the algebra $L^1(S)$ that can be associated with each semigroup $S$.

First we will describe a semigroup to represent link grammar in terms of strings of left and right connectors.
\begin{defn}[Bracket Semigroup]\index{bracket semigroup|textbf}
We define
$$D(L) = D_l(L) \cup D_r(L) \cup \{0\},$$
and let $\equiv$ be the minimal congruence on $D(L)^*$ satisfying
$$\bracket{x}{y} \equiv \left\{ \begin{array}{ll}
	0 & \text{if }x \neq y \\
	1 & \text{if }x = y
\end{array}\right.,$$
for all $x,y \in L$ and $0x \equiv x0 \equiv 0$ for all $x \in D(L)^*$, where $1$ is the empty string. Then the \emph{bracket semigroup on $L$} is defined as $D(L)^*/ \equiv$. We identify the equivalence classes of the bracket semigroup by their shortest elements.
\end{defn}

Note that the identities that form the congruence are similar to those satisfied by the creation and annihilation operators; in fact, the bracket semigroup is not more than an algebraic description of these operators. By combining this representation with the one we are about to describe we will have a description of syntax that combines the best of both the representations.





\subsection{Inverse Semigroups}
\label{inverse}

The bracket semigroup defined previously falls within a more general category of semigroups: that of inverse semigroups.

\begin{defn}[Inverse Semigroup]\index{inverse semigroups|textbf}
An inverse semigroup $S$ is a semigroup such that each element $x \in S$ has a unique element $x^{-1} \in S$ such that $xx^{-1}x = x$ and $x^{-1}xx^{-1} = x^{-1}$.
\end{defn}

\begin{prop}
A bracket semigroup is an inverse semigroup.
\end{prop}

\begin{proof}
Define $\bra{x}^{-1} = \ket{x}$ and $\ket{x}^{-1} = \bra{x}$. Let $x_1x_2\ldots x_n$ be a representative element of an equivalence class of a bracket algebra, then define
$$(x_1x_2\ldots x_n)^{-1} = x_n^{-1}x_{n-1}^{-1}\ldots x_1^{-1}.$$
Then the operation as given defines a unique inverse satisfying the requirements of an inverse semigroup.
\end{proof}

The identification of link grammars as a type of inverse semigroup  has led us to consider other kinds of inverse semigroup as a possible means of incorporating semantics into the formalism. We recount some basic properties of inverse semigroups \citep{Howie:76}.

Let $S$ be an inverse semigroup with set of idempotents\index{idempotents} $E(S)$. Then:
\begin{itemize}
\item $(a^{-1})^{-1} = a$ for all $a\in S$.
\item $aa^{-1} \in E(S)$ for all $a \in S$.
\item $aea^{-1} \in E(S)$ for all $a \in S$, $e\in E(S)$.
\item $e^{-1} = e$ for all $e \in E(S)$.
\item $ef=fe$ for all $e,f \in E(S)$, i.e.~idempotents commute, and thus form a subsemigroup of $S$.
\item A partial order $\le$ can be defined on $S$ by $a \le b$ if there exists $e \in E(S)$ such  that $a=eb$. If $a \le b$ then:
\begin{itemize}
\item[$\diamond$] $aa^{-1} = ba^{-1}$
\item[$\diamond$] $a = ab^{-1}a$
\item[$\diamond$] There exists $e \in E(S)$ such that $a=be$
\end{itemize}
\item The partial order is easily seen to be a generalisation of the semilattice order on a commutative semigroup of idempotents, defined by $e \le f$ if $ef = e$, and $e \land f = ef$.
\end{itemize}

\subsection{Free Inverse Semigroups}
\index{inverse semigroups!free}

The bracket semigroup does not store the `parse' of a sentence, it merely informs us whether a sentence parses or not. An alternative construction that is of great importance for our studies is the notion of a \emph{free} inverse semigroup. We can use this structure to represent syntax; as we will see, a link grammar parse of a sentence corresponds to an idempotent in a corresponding free inverse semigroup. In this representation, the parse can be deduced from the idempotent itself; the semigroup effectively stores information about the parse of the sentence. This allows us to build context theories in which the sentence structure is built up as words are concatenated; the sentence structure is represented by the context theory, which is an important step towards incorporating semantic information into this structure.

The crucial work on free inverse semigroups was done by \cite{Munn:74}\index{Munn, W.~D.} in which he proves that free inverse semigroups are isomorphic to \emph{birooted word-trees}\index{birooted word-trees}, also called Munn trees\index{tree!Munn}.

Informally, the free inverse semigroup on a set $A$ is formed from elements of $A$ and their inverses, $A^{-1} = \{a^{-1} : a \in A\}$, satisfying no other condition than those of an inverse semigroup. Formally, the free inverse semigroup is defined in terms of a congruence relation on $(A \cup A^{-1})^*$ specifying the inverse property and commutativity of idempotents --- see \cite{Munn:74} for details. We denote the free inverse semigroup on $A$ by $\FIS(A)$.

\subsection{Equivalence to Birooted Word-Trees}

A birooted word-tree on a set $A$ is a directed acyclic graph whose edges are labelled by elements of $A$ which does not contain any subgraphs of the form $\bullet \stackrel{a}{\longrightarrow} \bullet \stackrel{a}{\longleftarrow} \bullet$ or $\bullet \stackrel{a}{\longleftarrow} \bullet \stackrel{a}{\longrightarrow} \bullet$, together with two distinguished nodes, called the start node, $\Box$ and finish node, $\circ$.

A element in the free semigroup $\FIS(A)$ is denoted as a sequence $x_1^{d_1}x_2^{d_2}\ldots x_n^{d_n}$ where $x_i \in A$ and $d_i \in \{1,-1\}$.

 We construct the birooted word tree by starting with a single node as the start node, and for each $i$ from 1 to $n$:
\begin{itemize}
\item Determine if there is an edge labelled $x_i$ leaving the current node if $d_i = 1$, or arriving at the current node if $d_i = -1$.
\item If so, follow this edge and make the resulting node the current node.
\item If not, create a new node and join it with an edge labelled $x_i$ in the appropriate direction, and make this node the current node.
\end{itemize}
The finish node is the current node after the $n$ iterations.

As an example consider the set $A = \{a,b,c,d\}$, and the element in $\FIS(A)$ given by the sequence
$$aaa^{-1}bcdbb^{-1}aa^{-1}d^{-1}c^{-1}ac.$$
This has the following graph:

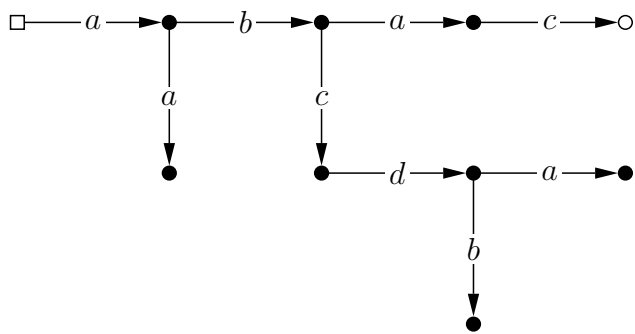
\begin{center}
\begin{graph}(10,6)(-1,-5)
\newcommand{\diredgetext}[3]{\diredge{#1}{#2}\edgetext{#1}{#2}{#3}}

\squarenode{R1}(0,0)[\graphnodecolour{1}]
\roundnode{R2}(2,0)
\roundnode{R3}(2,-2)
\roundnode{R4}(4,0)
\roundnode{R5}(4,-2)
\roundnode{R6}(6,-2)
\roundnode{R7}(6,-4)
\roundnode{R8}(8,-2)
\roundnode{R9}(6,0)
\roundnode{R10}(8,0)[\graphnodecolour{1}]
\diredgetext{R1}{R2}{$a$}
\diredgetext{R2}{R3}{$a$}
\diredgetext{R2}{R4}{$b$}
\diredgetext{R4}{R5}{$c$}
\diredgetext{R5}{R6}{$d$}
\diredgetext{R6}{R7}{$b$}
\diredgetext{R6}{R8}{$a$}
\diredgetext{R4}{R9}{$a$}
\diredgetext{R9}{R10}{$c$}
\end{graph}
\end{center}

The product of two elements $x$ and $y$  in the free inverse semigroup can be computed by finding the birooted word-tree of $x$ and that of $y$, joining the graphs by equating the start node of $y$ with the finish node of $x$ (and making it a normal node), and merging any other nodes and edges necessary to remove any subgraphs of the form  $\bullet \stackrel{a}{\longrightarrow} \bullet \stackrel{a}{\longleftarrow} \bullet$ or $\bullet \stackrel{a}{\longleftarrow} \bullet \stackrel{a}{\longrightarrow} \bullet$.

The inverse of an element has the same graph with start and finish nodes exchanged.

\subsection{Syntactic Equivalence}

We can represent parses of sentences in link grammar by translating words to syntactic categories in the \emph{free inverse semigroup} instead of the bracket algebra. In this case sentences are represented as idempotents. For example, the parse shown earlier for ``they mashed their way through the thick mud'' can be represented in the inverse semigroup on $A = \{s,m,o,d,j,a\}$ as
$$ss^{-1}modd^{-1}o^{-1}m^{-1}jdaa^{-1}d^{-1}j^{-1}$$
which has the following birooted word-tree:

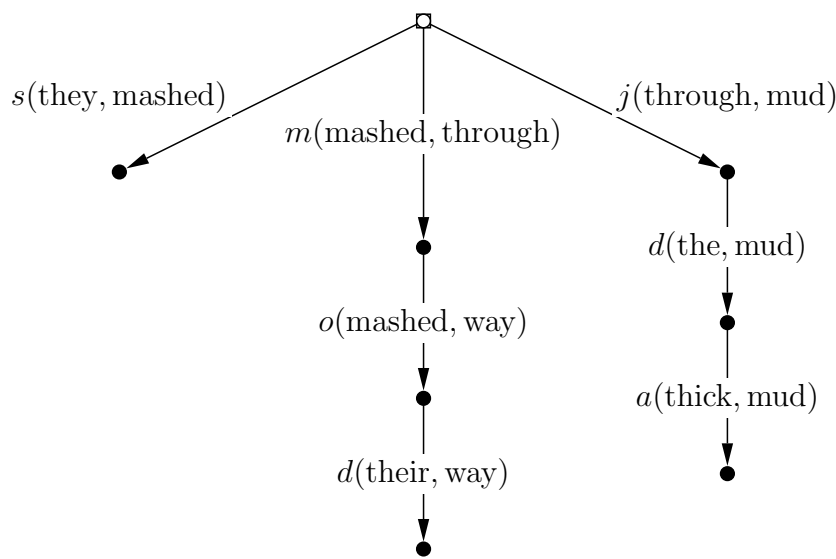
\begin{center}
\begin{graph}(8,8)(0,-7.5)
\newcommand{\diredgetext}[3]{\diredge{#1}{#2}\edgetext{#1}{#2}{#3}}
\squarenode{R1}(4,0)[\graphnodecolour{1}]
\roundnode{R2}(0,-2)
\roundnode{R3}(4,-3)
\roundnode{R4}(4,-5)
\roundnode{R5}(4,-7)
\roundnode{R6}(8,-2)
\roundnode{R7}(8,-4)
\roundnode{R8}(8,-6)
\roundnode{R1Copy}(4,0)[\graphnodecolour{1}]
\freetext(0,-1){$s(\text{they},\text{mashed})$}
\diredge{R1}{R2}
\diredgetext{R1}{R3}{$m(\text{mashed},\text{through})$}
\diredgetext{R3}{R4}{$o(\text{mashed},\text{way})$}
\diredgetext{R4}{R5}{$d(\text{their},\text{way})$}
\diredge{R1}{R6}
\freetext(8,-1){$j(\text{through},\text{mud})$}
\diredgetext{R6}{R7}{$d(\text{the},\text{mud})$}
\diredgetext{R7}{R8}{$a(\text{thick},\text{mud})$}
\end{graph}
\end{center}

In this graph, the fact that start and finish nodes overlap indicates that the element is idempotent. The nodes linked by the grammar are indicated in brackets; later we will be able to attach the meanings of these words to the links in the grammar.

We formalise the equivalence with the following proposition:
\begin{prop}
Let $S$ be the free inverse semigroup on the set of link types. The inverse semigroup representation of a disjunct is the element of $S$ formed by replacing each left and right link of type $a$ with elements $a \in S$ and $a^{-1} \in S$ respectively. Then if a sequence of disjuncts is a link grammar parse the product of the inverse semigroup representation of the disjuncts is idempotent.
\end{prop}
\begin{proof}
Let $x$ be the concatenation of disjuncts (we can also interpret $x$ as an element of $S$), and let $a$ be the first (leftmost) element of the sequence $x$. If the sequence of disjuncts is a link grammar parse then for each left connector there is a corresponding right connector on its right, and each connector is connected to exactly one other connector, so the first connector must be a left connector and there must be a corresponding $a^{-1}$ to represent its right connector on the right. Let $y$ be the subsequence of $x$ such that $x = aya^{-1}z$ for some sequence $z$. If $y$ and $z$ are both the empty string then $x$ is idempotent since $aa^{-1}$ is idempotent. Since no links cross, both $y$ and $z$ must satisfy the same conditions as $x$, and hence by induction, $x$ is idempotent, since $aea^{-1}$ and $aa^{-1}e$ are both idempotent for any idempotent $e$ in the inverse semigroup.
\end{proof}

Note that the converse implication does not hold in general since $a^{-1}a$ is also idempotent; thus this formulation allows right connectors to precede left connectors just as well as succeed them. In practice this should not be a problem since it is likely that the grammar can be redesigned in such a way that unwanted idempotents do not occur.

\subsection{A Semigroup for Syntax}

Both the bracket semigroup\index{bracket semigroup} and the free inverse semigroup\index{inverse semigroup!free} accurately represent syntax according to link grammar, however both have advantages and disadvantages for practical application in representing syntax. The free inverse semigroup stores information about the parse in a Munn tree\index{tree!Munn}, however combinations which don't parse will be `left over'. In the bracket semigroup, combinations which don't parse have a product of zero, so are ignored, but there is no memory of the parse.

For example, suppose nouns may optionally be preceded by an adjective ($a$) before taking a determiner ($d$) which we represent as $n_f = a^{-1}d^{-1} + d^{-1}$ in the $L^1$ algebra of the free inverse semigroup, and as $n_b = \ket{a}\ket{d} + \ket{d}$ in the $L^1$ algebra of the bracket semigroup. If the noun is now preceded by a determiner, $d$ or $\bra{d}$ respectively, then in the free inverse semigroup we have
$$dn_f = da^{-1}d^{-1} + dd^{-1}$$
while in the bracket semigroup we have $\bra{d}n_b = 1$ since $\bracket{d}{a} = 0$. Thus the free inverse semigroup correctly stores the idempotent $dd^{-1}$ but leaves the non-syntactic construction $da^{-1}d^{-1}$, while the bracket semigroup correctly cancels out this construction, but has no memory of the parse.

To get around this problem we combine the two structures; to do this we will need the \emph{direct product}. Given two semigroups $S_1$ and $S_2$ the direct product is the cartesian product $S_1 \times S_2$ with the semigroup product defined by
$$ (x_1,y_1)\cdot (x_2,y_2) = (x_1x_2,y_1y_2). $$
If $S_1$ and $S_2$ are inverse semigroups, then $S_1 \times S_2$ is an inverse semigroup, with inverse $(x,y)^{-1} = (x^{-1},y^{-1})$.

Given a set $A$ of links, we take the direct product of the free inverse semigroup on $A$ and the bracket semigroup on $A$, modulo an equivalence which makes elements zero in the bracket semigroup zero in the product. That is, the semigroup for syntax  is defined as
$$\FIS(A) \times B(A)\ / \equiv,$$
where $\equiv$ is defined by $(x,0) \equiv (y,0)$ for all $x,y \in \FIS(A)$. We are actually interested in the subsemigroup $S_s(A)$ generated by elements of the form $(a, \bra{a})$ and $(a^{-1},\ket{a})$ for all elements $a \in A$. We denote these elements $\sbra{a}$ and $\sket{a}$ respectively, and the idempotent $(aa^{-1},1)$ as $\sbracket{a}{a}$.

Our example then becomes
$$\sbra{d}n_s = \sbra{d} \Big(\sket{a}\sket{d} + \sket{d}\Big) = \sbracket{d}{d}$$
where $n_s = \sket{a}\sket{d} + \sket{d}$ is the representation of the noun in $S_s(A)$.

\subsection{From Semigroups to Context Theories}
\label{semigroup-context}

In this section we show how a context theory can be constructed from a semigroup. First we construct an algebra $L^1(S)$ of functions on a semigroup $S$ with multiplication defined by convolution (see section \ref{algebras}). This makes an algebra from a semigroup as we would expect intuitively; if $a,b,c,d\in S$ and $\alpha,\beta,\gamma,\delta \in \R$ then in $L^1(S)$ we have
$$(\alpha e_a + \beta e_b)(\gamma e_c + \delta e_d) = \alpha\gamma e_ae_c + \alpha\delta e_ae_d + \beta\gamma e_be_c + \beta\delta e_be_d$$
where $e_a$ is the basis element of $L^1(S)$ corresponding to $a$, that is the function that is 1 on $a$ and 0 elsewhere.

If, however, $S$ possesses a zero $\theta$, then this will not automatically be the zero of the algebra, instead it will be a function of $\theta$. What we will do is effectively equate the part of the algebra relating to $\theta$ to zero. Let $\bm{\theta}$ denote the ideal generated by $\theta$,
$$\bm{\theta} = \{\alpha\theta : \alpha \in \R\},$$
(assuming a real vector space). Then we are interested in the vector space $L^1(S)/\bm{\theta}$, that is the vector space of equivalence classes $x + \bm{\theta} = \{x + y: y \in \theta\}$. Addition and scalar multiplication in this space is defined by
\begin{gather*}
(x + \bm{\theta}) + (y + \bm{\theta}) = x + y + \bm{\theta}\\
\alpha(x + \bm{\theta}) = \alpha x + \bm{\theta}
\end{gather*}
Since $L^1(S)$ is also an algebra, we can define multiplication on $L^1(S)/\bm{\theta}$ by
$$(x + \bm{\theta})(y + \bm{\theta}) = xy + \bm{\theta}.$$
The equivalence class $0 + \bm{\theta}$ is now a zero of the vector space and the algebra; when there is no ambiguity, we shall simply denote it by $0$. If $ab = \theta$ in $S$, then in the algebra $L^1(S)/\bm{\theta}$ we have $e_ae_b = 0$.

Since $\bm{\theta}$ is a vector lattice supspace of $L^1(S)$, the space $L^1(S)/\bm{\theta}$ is a vector lattice; clearly it is also a lattice ordered algebra under the multiplication of $S$.

Since the $L^1$ norm is finite in the space $L^1(S)$ we can use it to define a linear functional:
$$\phi(u) = \|u^+\|_1 - \|u^-\|_1$$
Thus together with an assignment from words to elements of $L^1(S)$, we have a context theory.

\subsection{Relating Link and Categorial Grammars}
\index{categorial grammar!and link grammar}

The inventors of link grammar describe a translation from Bar-Hillel type categorial grammars to link grammar \citep{Sleator:93}. They describe it recursively in terms of a function $E$ that takes a categorial expression and returns a link grammar expression. In our notation, it can be expressed as follows:
\begin{itemize}
\item The set of link types $L$ is the set of categorial expressions.
\item If a word has a set $\{x_1,x_2,\dots,x_n\}$ of categorial expressions, then it is represented by the sum
$$E(x_1)+E(x_2)+\ldots E(x_n).$$
\item The representation of a basic type $A$ is
$$E(A) = \ket{A} + \bra{A}.$$
\item The representation of other categories is given by
\begin{eqnarray*}
E(x/y) &=& \ket{x/y} + \bra{x/y} + E(x)\bra{y}\\
E(y\backslash x) &=& \ket{y\backslash x} + \bra{y\backslash x} + \ket{y}E(x)
\end{eqnarray*}
\end{itemize}
As \citeauthor{Sleator:93} note, the size of the link grammar representation is linear in the size of the categorial grammar representation, thus they expect that translating to link grammar would be an effective method of parsing categorial grammars. From our perspective, there is an additional potential use for the translation: the connection enables a new way of implementing statistical categorial grammars, using the statistical link grammar formalisms.

\index{link grammar|)}

\section{Discussion and Further work}
\label{discussion-syntax-section}

Using the constructions of the previous section, we have described a formalism that parses\index{parsers} sentences in a purely algebraic fashion. The advantage of this algebraic description over the operator-based description is that the parse itself is stored in algebraic form and does not need to be reconstructed from information about which disjunct was used with each word. This is due to the extra structure provided by the free inverse semigroup which allows tree-like structures to be represented. It is this structure that we believe will also be useful for constructing representations of meaning directly within the context theoretic framework. For example, it may be possible to find link grammars for natural language such that the Munn tree\index{tree!Munn} of a sentence describes relationships between the words in the sentence. This can already be seen to be true to some degree: for example, in the tree we showed for the sentence ``They mashed their way through the thick mud'', the branch relating to ``thick'' comes off the branches relating to ``mud'', in terms of idempotents, the idempotent representing ``the thick mud'' is more specific than that representing ``the mud''. The trees still bear little resemblance to the dependency trees that we are familiar with, however.

We have now described methods for representing meaning and syntax in algebra. The question arises how one may combine such methods to produce lexicalised algebraic representations of language incorporating both meaning and syntax. One may wish to choose a particular vector-based semantic formalism and a particular syntactic formalism and combine them. One way of doing this may be the mathematics of \emph{free probability}\index{free probability|textbf} \citep{Voiculescu:97}. The concept of freeness generalises the concept of independence to the case of non-commutative variables. Two sub-algebras $A_1$ and $A_2$ of an algebra are said to combine freely with respect to a linear functional $\phi$ if $\phi(x_1x_2\ldots x_n) = 0$ whenever all the $x_i$ satisfy $\phi(x_i) = 0$ and no two adjacent $x_i$ are in the same sub-algebra. Given two non-commutative probability spaces, one can construct their free product as an algebra which has sub-algebras isomorphic to the original algebras and satisfying the condition of freeness. Thus one could choose a context-theory to represent meaning and a context-theory to represent syntax and build a combined context-theory using the free product, in which each word would map to a product of its original syntactic and semantic representations. The idea that meaning and syntax combine freely is appealing since we are used to thinking of these two aspects of language separately; the concept of freeness may encapsulate this idea well, however exactly how it would work in practice remains to be seen.

We have left the question of how to compute with these new representations largely unanswered, however we are representing existing formalisms for which computational procedures already exist. Thus it may be possible to make use of existing algorithms with small adjustments to compensate for the differences that the context-theoretic perspective requires. It is our hope however that new and more efficient computational procedures will be brought to light by considering the algebraic approach, particularly in the area of statistical parsing. One area that seems particularly worthy of further investigation is the use of matrices\index{matrices} to approximate elements of algebras, along the lines of the description we gave for Fock space operators in terms of matrices.


%

%


\chapter{Conclusions and Future Work}

We have presented a context-theoretic framework for natural language semantics. The framework is founded on the idea that meaning in natural language can be determined by context, and is inspired by techniques that make use of statistical properties of language by analysing large text corpora. Such techniques can generally be viewed as representing language in terms of vectors. These techniques are currently used in applications such as textual entailment recognition, however the lack of a theory of meaning that incorporates these techniques means that they are often used in a somewhat ad-hoc manner. The purpose behind the framework is to provide a unified theoretical foundation for such techniques so that they may used in a principled manner.

\section{Summary of Part I}

In Part I of the thesis we define the framework, giving background to our definition and developing a model of meaning as context; we then abstract this theory by choosing certain properties of the model to form the basis of the framework.

\subsection*{Chapter \ref{background}}

In this chapter we gave an overview of the philosophical background of the notion of meaning as context. Our purpose in doing this was both to show how the techniques which make use of context can be viewed as arising out of these ideas, and to provide a grounding for the ideas we present in our framework.

Wittgenstein proposed that ``meaning is use'': knowing the meaning of a word is the same as knowing how to use it. Firth was interested in the ``context'' of a word in the sense of the situation in which a word was used and the objects that are physically present; he said ``you shall know a word by the company it keeps''. He also said however that part of the meaning of a word may be by collocation. The example he gives is of the word ``ass'', part of whose meaning is by collocation with expressions such as a preceding ``you silly'' --- this idea is much closer in spirit to more recent ideas of meaning as context, however Firth does not go as far as claiming all words may be considered in this way. The first to do this was Harris, whose famous distributional hypothesis discussed the relationship between meanings of words and the distribution of their contexts. He said that words will occur in similar contexts if and only if they have similar meanings.

More recently, it was proposed by \cite{Weeds:04} that the property of distributional generality may be correlated to semantic generality, i.e.~that if a word occurs in a wider set of contexts than another word, its meaning is also likely to be more general, and vice-versa. This idea was taken up by \cite{Geffet:05} specifically for lexical semantics with their distributional inclusion hypotheses, which effectively equate the meaning of a word in terms of entailment to specific features of the contexts that the word occurs in.

In this chapter we also introduced the techniques which inspired the framework, looking at latent semantic analysis and its variations and measures of distributional similarity. We first discussed the different methods for building context vectors that are common to all the techniques --- for example whether the context is considered to be a certain window of text surrounding the word or whether dependency features are used.

We discussed three related techniques: latent semantic analysis, probabilistic latent semantic analysis and latent Dirichlet allocation. These can all be viewed as attempting to build a model that extracts ``latent'' information from the context vectors that is assumed to relate to their meaning. Latent semantic analysis does this by means of a singular value decomposition which allows the important information in the vectors to be extracted by reducing the dimensionality of the representation. Doing this forces vectors to become closer to one another allowing ``latent'' relationships between words to be detected that would not be seen looking at the context vectors alone. This technique is not ideal however as it is not probabilistic in nature and the resulting representation can be hard to interpret, for example it may contain negative values. Both probabilistic latent semantic analysis and latent Dirichlet allocation provide a probabilistic analysis of the situation. In the former it is postulated that there is a hidden variable that is responsible for the occurrence of words and contexts, which are conditionally independent given this variable. The latter technique defines a generative model of a corpus which describes an infinite array of documents --- this overcomes a perceived limitation in probabilistic latent semantic analysis which considers only a fixed number of possible contexts.

We then looked at measures of distributional similarity. Instead of building models based on context vectors, these attempt to measure the similarity or difference between the vectors directly. Certain of the measures used can be given a geometric interpretation, while some are information theoretic in nature, for example being based on the Kullback-Leibler divergence. Some measures are based on the ``features'' of a term --- those contexts that are considered to provide useful information about a word, this may for example be those with positive mutual information.

\subsection*{Chapter \ref{meaning-context}}

This chapter forms the heart of the thesis. In it we build an abstract mathematical model that gives a specific interpretation of what it means for meaning to be determined by context. We propose that a string is represented by a vector that represents the contexts it occurs in in a corpus model --- an abstraction of a text corpus that allows infinite possible documents. We examine mathematical properties of the model which then form the basis of the context-theoretic framework later in the chapter.

We discuss the concept of distributional generality and show how it relates to the model. Specifically we show that space of context vectors can be thought of as a vector lattice, and we propose that the partial ordering of the lattice structure can be thought of as describing entailment between strings: generally if one context vector is less than or equal to another in this partial ordering it is because the latter occurs in a wider range of contexts, thus we make the assumption that it is also more general in meaning.

We then define the context-theoretic probability in terms of the model of the meaning as context. This provides us with a way to measure the size of context vectors that is similar to familiar notions of the probability of a string. This definition enables us to define a degree of entailment between strings. This is important because we assume that it is unlikely that a string will completely entail another according to the partial order of their context vectors, instead we allow degrees of entailment between strings based on a Bayesian interpretation of the context-theoretic probability.

We show how the vector space generated by context vectors can be considered as an algebra over a field where concatenation of strings is interpreted as multiplication in the vector space. This is important because it informs us about the allowed nature of multiplication in the vector space: addition must distribute with respect to the product. We incorporate this property as a central feature of the framework.

In the second part of the chapter we define the framework itself. We call implementations of the framework context theories since we view them as describing theories about the contexts that strings occur in. We require that context theories have some of the properties of the model of meaning as context; specifically, words must be represented as vectors in an algebra that also incorporates a compatible lattice structure and the context-theoretic probability must be described.

\section{Summary of Part II}

In Part II we look at applications of the framework, in order to demonstrate its effectiveness in applications. We look at four areas: that of textual entailment, representing ambiguity and uncertainty in logical semantics, relating taxonomies to the framework and representing syntax.

\subsection*{Chapter \ref{entailment-chapter}}

In this chapter we review approaches to recognising textual entailment, the task of determining, given two sentences, whether the first entails the second. We summarise the PASCAL Recognising Textual Entailment Challenge and approaches taken by entrants to the challenge. We examine in detail the probabilistic framework of \citeauthor{Glickman:05} for textual entailment, which has similar aims as our own framework. Theirs requires the second sentence (the hypothesis) to be given a logical interpretation however; this is not ideal for many approaches to textual entailment since they often make use of context-based representations of meaning that cannot easily be given logical interpretations.

We then examine approaches to the task that made use of logical representations of meaning and methods of making such systems robust. These included systems which used scoring or cost-based systems to allow simplifications to logical representations in order to reason effectively, systems that made use of classifiers  in which the results of logical inference are one ``feature'' in the classifier, and systems which used model building. What was lacking in many of these approaches was a firm theoretical foundation that would provide guidelines as to how to make the systems more robust in a principled manner.

In the second part of this chapter we showed how existing approaches to textual entailment can be considered within the context-theoretic framework. We first describe context theories for simple approaches to the task such as lexical and subsequence matching. We then analyse \citeauthor{Glickman:05}'s approach to the task, showing how it relates to the framework. This leads us to an adaptation of their approach which deals with the problem of data sparseness using latent Dirichlet allocation to form a context theory.

\subsection*{Chapter \ref{model-theoretic-chapter}}

A major aim of the framework is to provide insight into the problems of uncertainty and ambiguity in natural language, especially when making use of logical representations of language. Because logical systems\index{logical semantics} on their own are generally brittle, in natural language applications such as that of recognising textual entailment ways have to be found to make the systems more robust, and reasoning about uncertainty is one way in which this can be done. In this chapter we showed how logical semantics can be described in terms of context theories; this provided us with guidance as to how to represent uncertainty and ambiguity within the framework. We described how uncertainty arising from parsing and other sources can be incorporated into the  representation, and discussed the representation of word sense ambiguity within the framework. We outlined a possible way of implementing these ideas, showing how computational problems in dealing with the representation can be overcome. We also discuss how entailment can be represented between words and phrases within the framework, using logical semantics.


\subsection*{Chapter \ref{ontologies}}

In this chapter we discussed the relationship between ontological and vector-based representations of lexical semantics, presenting several ways of constructing vector based representations of a taxonomy using what we call vector lattice completions. We described several such completions: a probabilistic completion that allows the incorporation of the probability of a concept into the taxonomy, a distance preserving completion that preserves the Jiang-Conrath measure of distance in the ontology in the vector lattice structure, and a completion that uses a smaller number of dimensions for certain taxonomies. We also discuss how the representations of concepts in the vector lattice may be combined to form representations of ambiguous words.

\subsection*{Chapter \ref{syntax-chapter}}

In the last chapter we discussed the representation of syntactic structure within the framework, identifying categorial grammar and link grammar as being most suited to the context-theoretic approach. We give an overview of some of the different types of categorial grammar and show that the Lambek calculus can be described in terms of a context theory; we also discuss why the other types of categorial grammar are not so well suited to the context-theoretic approach.

We give a detailed analysis of link grammar and describe it in terms of context theories in two different ways: in terms of operators on an infinite dimensional vector space, and in terms of inverse semigroups. The former leads us to a new description of a simple form of stochastic link grammar as well as a method of computing with link grammars using matrices. The latter shows how the structure of a sentence may be incorporated into its vector representation, bringing us a step nearer to incorporating semantic information into such representations.

\section{Future Work}

We divide future work into two sections, possible practical investigations arising directly from the work we have discussed, and theoretical issues that remain unresolved.

\subsection{Practical Investigations}

Chapters \ref{entailment-chapter} and \ref{model-theoretic-chapter} raise many possibilities for the design of systems to recognise textual entailment within the framework:
\begin{itemize}
\item Variations on substring matching: experiments with different weighting schemes for substrings, allowing partial commutativity of words or phrases, and replacing words with vectors representing their context, using tensor products of these vectors instead of concatenation.
\item Extensions of Glickman and Dagan's approach and our own context-theoretic approach using latent Dirichlet allocation, perhaps using other corpus models perhaps based on $n$-grams or other models in which words don't commute, or a combination of context theories based on commutative and non-commutative models.
\item Implementations of the outline described in Chapter \ref{model-theoretic-chapter} for representing uncertainty in logical semantics: here there are many possible variations based on what information on uncertainty is included (from parsing, part-of-speech tagging, word sense disambiguation and so on), what logical representation is used, what methods of assigning probabilities to logical statements and to strings are used, and what computational procedure is used to calculate or estimate the degree of entailment.
\item Weighted combinations of any of the above techniques. A question that will then need to be addressed is how to find an optimum weighting scheme for the different context theories.
\end{itemize}
All of these ideas could be evaluated using the data sets from the Recognising Textual Entailment Challenges.

Several areas of investigation are suggested by Chapter \ref{ontologies}. The first concerns the construction of ideal completions: it would be interesting to construct ideal completions for real taxonomies such as that of WordNet, and investigate the consequences of performing various types of dimensionality reduction (for example singular valued decomposition or random projections) on the resulting representations, to see for example how the quality of the representation degrades as the size of the vector space is reduced.

The vector lattice completions suggest new ontological distance measures, for example the $l^p$ norms can be used on any of the vector lattice completion representations. It would be interesting to see how these compare to existing measures of ontological distance in applications.

Another area of investigation suggested by this chapter is the possibility of using vector lattice completions together with measures of distributional similarity to construct ontologies automatically. If enough of a correlation can be found between distance in the vector lattice and measures of distributional distance or similarity then the latter could be used to attempt to place a concept in the vector lattice and hence in the ontology. A related idea is that of ontological smoothing: the positions of concepts in the vector lattice could be altered based on measures of distributional similarity and the resulting representations compared to the original representation in applications.

The most promising area for investigation arising from Chapter \ref{syntax-chapter} is that of matrix-based parsing of link grammars. The practicality of this idea needs to be investigated in detail; one possibility is that dimensionality reductions may be used to perform statistical parsing efficiently with matrices. Other areas include looking at computational procedures for dealing with the context-theoretic representation of the Lambek calculus.

\subsection{Theoretical Investigations}

Several areas for investigation are suggested by Chapter \ref{meaning-context}. Proving Conjecture \ref{conjecture} may provide further insight into the nature of meaning as context, as well as giving evidence for our requirement that a context theory should be a lattice ordered algebra (instead of just a partially ordered algebra). A further interesting question is which algebras are isomorphic to the context algebra of some corpus model. It may be that stronger conditions can be placed on context theories to restrict the formalism to such algebras --- such implementations would truly deserve to be called ``context theories''\index{context theory}.

In Chapter \ref{ontologies} we only considered the representation of the ``is-a'' relation of ontologies. An interesting question is how other ontological relationships such as meronymy and antonymy may be expressed within the vector lattice structure.

An area of interest suggested by the work in Chapter \ref{syntax-chapter} is the use of free probability\index{free probability} to combine context theories --- this seems to us a very promising area for future work that may lead to entirely new representations of language. Our hope is that we will eventually be able to take a vector based model of meaning and combine it with a statistical model of syntax to produce a complete vector-based semantics for natural language, by combining the corresponding context theories using free probability. Because of the statistical nature of both aspects of this construction, such a semantics would be robust, and there may be potential for computing efficiently with this semantics using matrices and dimensionality reductions.

A subject that we have not considered much in this thesis is the issue of multi-word expressions and non-compositionality. What predictions does the context-theoretic framework make about non-compositionality? Answering this may lead us to new techniques for recognising and handling multi-word expressions and non-compositionality.


Of course it is hard to predict the benefits that may result from what we have presented, since we have given a way of thinking about meaning in natural language that in many respects is new. This new way of thinking opens the door to the unification of logic-based and vector-based methods in computational linguistics, and the potential fruits of this union are many.


\appendix
 



 \chapter{Mathematical Methods for Computational Linguistics}

This appendix provides a reference for foundational mathematical concepts that are necessary for an understanding of the thesis.

\section{Semigroups, Groups and Fields}
\label{semigroups}

\begin{defn}[Semigroup]\index{semigroup|textbf}
A \emph{binary operation} on a set $S$ is a function from $S\times S$ to $S$. The value of the binary operation $\cdot$ on two elements $x$ and $y$ in $S$ is denoted $x \cdot y$. A semigroup $(S,\cdot)$ is a set $S$ with a binary operation $\cdot$ which is \emph{associative}:
$$(x\cdot y)\cdot z = x \cdot (y \cdot z).$$
This product is often denoted $x\cdot y \cdot z$ or simply $xyz$.

An element $e$ of $S$ is called \emph{unity} if $es = se = s$ for all $s \in S$. There can only ever be at most one unity in $S$: if $e_1$ and $e_2$ are unities then $e_1e_2 = e_1 = e_2$. A semigroup with unity is often called a \emph{monoid}.
\end{defn}
 
 
 \begin{defn}[Free Semigroup]\index{free semigroup|textbf}
Let $A$ be a set. The set $A^*$ is the set of all finite sequences of symbols of elements of $A$. Then $A^*$ is a semigroup (called the \emph{free semigroup} on $A$) under concatenation of sequences, $x\cdot y = xy$ for $x,y \in A^*$.
 \end{defn}
 
 
 \begin{defn}[Group]\index{group|textbf}
 A group $G$ is a monoid with unity $e$ such that for each element $x \in G$ there is an element $x^{-1}$, called the \emph{inverse} of $x$, such that $xx^{-1} = x^{-1}x = e$. A group is called \emph{abelian} or \emph{commutative} if $xy = yx$ for all $x,y\in G$.
 \end{defn}
 
 \begin{defn}[Field]\index{field|textbf}
 A field is a set $F$ together with two operations $+$ and $\cdot$ called addition and multiplication such that $F$ is an abelian group under addition with (additive) identity $0 \in F$, and a commutative monoid under multiplication, with (multiplicative) identity $1 \in F$, with $1 \neq 0$, such that every element $x \in F$ except $0$ has a multiplicative inverse $x^{-1}$ (that is, $F - \{0\}$ is an abelian group under multiplication) and multiplication distributes over addition:
$$x\cdot(y + z) = x\cdot y + x \cdot z$$
\end{defn}

\begin{defn}[Congruence] \index{congruence|textbf}
A congruence on a semigroup $S$ is an equivalence relation $R$ on $S$ that is preserved under multiplication, so that if $aRb$ then $xaRxb$ and $axRbx$. Let $aR$ denote the set $\{x : aRx\}$, called the equivalence class of $a$. We can define a product on equivalence classes by $aR \circ bR = abR$. This semigroup is denoted $S/R$, and is called the quotient of $S$ with respect to $R$.

If $R_1$ and $R_2$ are congruences on $S$ then $R_1 \cap R_2$ is also a congruence relation. Since the universal relation $U$ defined by $xUy$ for all $x,y \in S$ is a congruence, we can find for every relation $R$ on $S$ the smallest congruence containing $R$ as the intersection of all congruences $R'$ with $R' \supseteq R$.
\end{defn}

\section{Vector Spaces}
\label{vectors}

\begin{defn}[Vector Space]\index{vector space|textbf}
A vector space over a field $F$ is a set $V$ with two operations: addition, $V \times V \rightarrow V$, denoted $u + v$ where $u,v \in V$, and scalar multiplication: $F \times V \rightarrow V$, denoted $\alpha v$ where $\alpha \in F$ and $v \in V$,
satisfying the following conditions:
\begin{itemize}
\item $V$ is closed under addition and scalar multiplication;
\item the vector space under addition forms an \emph{abelian group}: addition is associative and commutative and there is an additive identity $0 \in V$ such that for every element $v \in V$ there is an element $-v$ such that\footnote{In general we write $x - y$ for $x + (-y)$.} $v + (-v) = 0$;
\item scalar multiplication is associative: $\alpha (\beta v) = (\alpha \beta) v$ for $\alpha, \beta \in F$ and $v \in V$;
\item $1v = v$ where $1$ is the multiplicative identity of $F$;
\item scalar multiplication is distributive with respect to vector and scalar addition:
\begin{eqnarray*}
\alpha(u + v) & = & \alpha u + \alpha v\\
(\alpha + \beta)v & = & \alpha v + \beta v
\end{eqnarray*}
\end{itemize}
When the field $F$ is that of the real or complex numbers $\R$ or $\mathbb{C}$, the vector space is called `real' or `complex' respectively. Unless otherwise stated, we shall be dealing exclusively with real vector spaces.
\end{defn}

\begin{defn}[Finite-dimensional Real Vector Spaces]\index{vector space!finite dimensional|textbf}
The most important examples for computational linguists are the $n$-dimensional real vector spaces, denoted $\R^n$. An element of $\R^n$ is denoted
$$x = (x_1,x_2,\ldots x_n),$$
where the $x_i$ are the real valued \emph{components} of $x$. The operations on $\R^n$ are defined as follows:
\begin{eqnarray*}
x + y & = & (x_1 + y_1, x_2 + y_2, \ldots, x_n + y_n)\\
\alpha x & = & (\alpha x_1, \alpha x_2, \ldots, \alpha x_n)\\
-x & = & (-x_1, -x_2, \ldots, -x_n);
\end{eqnarray*}
the zero element is the element all of whose components are zero.
Given a finite set $S$, we write $\R^S$ for the vector space $\R^{|S|}$; then each element of $S$ corresponds to a dimension in $\R^S$.
\end{defn}

\subsection{Notions of Distance}

The following sequence of definitions are to do with the notion of ``distance'' and ``size'' of objects. These concepts are of key importance in computational linguistics because we are often interested in ``distances'' between words---for example semantic distance. The types of space, in order of generality, are \emph{metric space}, \emph{normed space} and \emph{inner product space}.
\begin{defn}[Metric]\index{metric|textbf}
A metric $d$ is a function on a set $X$ satisfying:
\begin{center}
\begin{tabular}{ll}
$d(x,y) \ge 0$ & (non-negativity)\\
$d(x,y) = 0$ if and only if $x=y$ & (identity of indiscernibles)\\
$d(x,y) = d(y,x)$ & (symmetry)\\
$d(x,z) \le d(x,y) + d(y,z)$ & (triangle inequality)
\end{tabular}
\end{center}
for all $x,y,z \in X$. A metric space is a set $X$ together with a metric $d$.
\end{defn}

The definition of metric is very general: it does not require the set $X$ to be a vector space. In contrast, a more common way of defining distances on a vector space is via a \emph{norm}:
\begin{defn}[Norm]
If $V$ is a vector space over a field $F$ which is a subfield of the complex numbers, a norm $\|\cdot\|$ is a function from $V$ to the real numbers satisfying:
\begin{center}
\begin{tabular}{ll}
$\|x\| \ge 0$ & (positivity)\\
$\|\alpha x\| = |\alpha|\cdot\|x\|$ & (positive scalability)\\
$\|x + y\| \le \|x\| + \|y\|$ & (triangle inequality)\\
$\|x\| = 0$ if and only if $x = 0$ & (positive definiteness)
\end{tabular}
\end{center}
A \emph{normed vector space} is a vector space together with a norm.
\end{defn}
It is fairly straightforward to see that a norm $\|\cdot\|$ on a vector space $V$ defines a metric $d$ on $V$ by $d(x,y) = \|x - y\|$.

\begin{defn}[$l^p$ Norms]\index{lp norm@$l^p$ norm|textbf}
The most important examples are given by the $l^p$ norms, for $p$ a real number $\ge 1$. For the vector space $\R^n$, the $l^p$ norm of an element $x = (x_1,x_2,\ldots,x_n)$ is given by
$$\|x\|_p = \left(|x_1|^p + |x_2|^p + \ldots + |x_n|^p\right)^{1/p}$$
The $l^\infty$ norm of $x$ is defined as the supremum of $|x_i|$ over all components $x_i$ of $x$.
\end{defn}

Some of the most important instances of vector spaces, namely the Hilbert spaces, are those with an \emph{inner product}, which corresponds to the familiar dot product on finite dimensional vector spaces. We give the definition here in terms of complex numbers for generality; we shall only ever need real vector spaces.
\begin{defn}[Inner Product]\index{inner product|textbf}
An inner product on a complex vector space is a function $\langle \cdot ,\cdot \rangle : V \times V \rightarrow \mathbb{C}$ satisfying for all $u,v,w \in V$ and $\alpha \in F$:
\begin{center}
\begin{tabular}{lc}
Additivity: &
$\inprod{u}{v+w}  =  \inprod{u}{v} + \inprod{u}{w}$\\
\vspace{0.1cm}
&$\inprod{u+v}{w}  =  \inprod{u}{w} + \inprod{v}{w}$\\
\vspace{0.1cm}
Nonnegativity: &
$\inprod{v}{v} \ge 0$\\
\vspace{0.1cm}
Nondegeneracy: &
$\inprod{v}{v} = 0\quad \textrm{iff}\quad  v = 0$\\
\vspace{0.1cm}
Conjugate symmetry: &
$\inprod{u}{v} = \overline{\inprod{v}{u}}$\\
\vspace{0.1cm}
Sesquilinearity: &
$ \inprod{u}{\alpha v} = \alpha\inprod{u}{v}$
\end{tabular}
\end{center}
where $\overline{\alpha}$ denotes the complex conjugate of $\alpha$. The definition clearly also holds when $V$ is a real vector space. A vector space with an inner product defined is called an \emph{inner product space}.

Note that conjugate symmetry implies that $\inprod{x}{x}$ is real for all $x$, and that conjugate symmetry and sesquilinearity together imply that
$$\inprod{\alpha x}{y} = \overline{\alpha}\inprod{x}{y}.$$
An inner product naturally defines a norm $\|\cdot\|$ on a vector space, by $\|x\| = \sqrt{\inprod{x}{x}}$.
\end{defn}

\begin{example}[Dot Product]\index{dot product}
The inner product or dot product on $\R^n$ is defined by
$$\inprod{x}{y}  = \sum_{i = 1\ldots n} x_iy_i.$$
The norm of a vector in $\R^n$ corresponds to its length: $\|x\| = \sqrt{\sum_{i=1\ldots n} x_i^2}$.
\end{example}

\subsection{Bases}
\label{basis}

Almost every vector space considered in computational linguistics comes with some basis, which can usually be conceptually linked to the notion of context. The notion of a \emph{basis} in a vector space is also very important in relation to \emph{vector lattices} (see Section \ref{vector-lattices}).

\begin{defn}[Basis]\index{basis|textbf}
A basis is a set $B$ of elements of a vector space $V$ over a field $F$, such that the elements are \emph{independant}, i.e., if
$$\sum_{b_i \in B} \alpha_i b_i = 0$$
for some set of $\alpha_i \in F$, then necessarily $\alpha_i = 0$ for all $i$; and $B$ \emph{spans} $V$, i.e., for each element $x \in V$,
$$x = \sum_{b_i \in B} \beta_i b_i$$
for some set of values $\beta_i \in F$.

Two elements $x,y$ in an inner product space $V$ are called \emph{orthogonal} if $\inprod{x}{y} = 0$. An orthonormal basis for $V$ is a basis $B$ such that any two distinct elements of $B$ are orthogonal and the magnitude of each element in $B$ is 1, i.e. $\inprod{b}{b} = 1$ for all $b\in B$.
\end{defn}

\begin{example}[Orthonormal Basis for $\R^n$]\index{basis!orthonormal|textbf}
An orthonormal basis for the vector space $\R^n$ is given by the set $\{e_1,e_2,\ldots e_n\}$ where $e_1 = (1,0,0,\dots 0)$, $e_2 = (0,1,0,\dots 0)$, \ldots, $e_n = (0,0,0,\dots 1)$. In this way, for the vector space $\R^S$, we can associate a basis element $e_s$ with each element $s \in S$.
\end{example}

\subsection{Completeness}
\label{completeness-section}

Completeness is a property of vector spaces which is difficult to grasp conceptually, and is not that important to understand in relation to applications in computational linguistics. However, it is a property that is possessed by a lot of interesting vector spaces, and is often required of vector spaces since it leads to things being mathematically very ``well behaved''.

\begin{defn}[Limit]\index{limit}
Let $a_1,a_2\ldots$ be an infinite sequence of real numbers. A real number $a$ is said to be the limit of the sequence if and only if for every real number $\epsilon > 0$, there is a natural number $n_0$ such that for all $n > n_0$, $|a_n - a| < \epsilon$.
\end{defn}

\begin{defn}[Completeness]\index{completeness|textbf}
Given a metric space $X$ with metric function $d$, a sequence $x_1, x_2, \ldots$ is called \emph{Cauchy} if for every positive real number $a$, there is an integer $n_0$ such that for all integers $m,n > n_0$, $d(x_m,x_n) < a$. If every Cauchy sequence has a limit in $X$, the metric space is called \emph{complete}.

A \emph{Banach space}\index{Banach space|textbf} is a normed vector space which is complete with respect to the metric $d$ defined by $d(x,y) = \|x - y\|$. A \emph{Hilbert space}\index{Hilbert space|textbf} is a vector space with an inner product which is complete with respect to the metric defined by the inner product norm, $d(x,y) = \sqrt{\inprod{x-y}{x-y}}$. A Hilbert space is thus a special kind of Banach space.
\end{defn}

\subsection{$l^p$ and $L^p$ Spaces}
\index{lp space@$l^p$ space|textbf}
\label{lp-space-section}

We shall often need to deal with infinite dimensional vector spaces, for example, we shall often want to associate a dimension with each sequence in a set of sequences $A^*$. When we do this, not all vectors will have finite norm, and precisely which ones do depends on which norm we use. We can thus categorise subspaces according to which norms are guaranteed to be finite. For $p \ge 1$ we define the $l^p$ space to be the vector space of all infinite sequences $x$ of real numbers $x = (x_1,x_2,\ldots)$ such that $\sum_i |x_i|^p$ is finite, together with the $l^p$ norm. The $l^\infty$ space is the set of all vectors with finite components, together with the $l^\infty$ norm.
All the $l^p$ spaces are Banach spaces, and only the $l^2$ space is a Hilbert space.

If $S$ is a countable set, we shall often want to consider real valued functions $f$ on $S$ as vectors. We can consider such functions as sequences of real numbers: writing $S = \{s_1, s_2, \ldots\}$, we can think of $f$ as a sequence $(f(s_1), f(s_2), \ldots)$. We denote by $L^p(S)$ \index{Lp space@$L^p$ space|textbf} the set of functions on $S$ which are in the corresponding $l^p$ space when viewed as sequences. A function $f$ in $L^\infty(S)$ is called a bounded function\index{bounded function|textbf} since there exists some bound $b \in \R$ such that $|f(x)| < b$ for all $x$.

\subsection{New vector spaces from old}
\label{new-vector-spaces}
\begin{defn}[Direct Sum]\index{direct sum|textbf}
Given two vector spaces $U$ and $V$ we can construct a vector space $U \oplus V$ called the \emph{direct sum} of $U$ and $V$. The direct sum is simply the cartesian product $U \times V$ with vector operations defined component-wise:
\begin{eqnarray*}
(u_1,v_1) + (u_2,v_2) & = & (u_1+u_2,v_1+v_2)\\
\alpha(u,v) & = & (\alpha u, \alpha v)
\end{eqnarray*}
where $u_i \in U, v_i \in V, \alpha \in F$. If $U$ and $V$ are Hilbert spaces, then $U \oplus V$ denotes the Hilbert space with the inner product defined by
$$\inprod{(u_1,v_1)}{(u_2,v_2)} = \inprod{u_1}{u_2} + \inprod{v_1}{v_2}$$
The dimension of $U\oplus V$ is equal to the sum of the dimensions of $U$ and $V$.
\end{defn}
\begin{defn}[Tensor Product]\index{tensor product|textbf}
The \emph{tensor product} $U \otimes V$ of two vector spaces $U$ and $V$ is constructed by taking the vector space generated by the cartesian product $U \times V$ and factoring out the subspace generated by the equations:
\begin{eqnarray*}
(u_1 + u_2) \otimes v & = & u_1 \otimes v + u_2 \otimes v\\
u \otimes (v_1 + v_2) & = & u \otimes v_1 + u \otimes v_2\\
\alpha u\otimes v & =  &u \otimes \alpha v = \alpha(u \otimes v) 
\end{eqnarray*}
where $u_i,u \in U$, $v_i,v \in V$ and $\alpha \in F$.

If $U$ and $V$ are Hilbert spaces, the tensor product is again a Hilbert space, with inner product defined by
$$\inprod{u_1\otimes v_1}{u_2 \otimes v_2} = \inprod{u_1}{u_2}\inprod{v_1}{v_2}.$$
The dimension of $U \otimes V$ is equal to the product of the dimensions of $U$ and $V$.
\end{defn}

\section{Lattice Theory}
\label{lattices}

The concepts described in this section deal with relationships between objects. One of the most important types of relationship that we consider on sets of objects is that of a \emph{partial ordering}. An example of this is the hypernymy relation between words (or equivalently the \textbf{is-a} or subsumption relation between concepts), discussed in Section \ref{taxonomy}. Another example is the subset relation on a set of sets.

These relations often satisfy much stronger conditions, which we classify in sequence: semilattices, lattices, modular lattices, distributive lattices and Boolean algebras. All of these have important characteristics which may also be expressed in algebraic terms.


\begin{defn}[Partial Ordering]\index{partial ordering|textbf}
\index{partial ordering|textbf}
A partial ordering on a set $S$ is a relation $\le$ that satisfies, for all $x,y,z \in S$:
\begin{center}
\begin{tabular}{ll}
$x \le x$ & (reflexivity)\\
if $x \le y$ and $y \le x$ then $x = y$ & (antisymmetry) \\
if $x \le y$ and $y \le z$ then $x \le z$ & (transitivity)
\end{tabular}
\end{center}
\end{defn}
If $a \le b$ then we say $a$ is \emph{contained in} or \emph{is less than} $b$. An example of a partial ordering is the set inclusion relation, $\subseteq$ on a set of subsets of a set, or the `less than or equal' relation on the natural numbers.

The following definition is useful for describing properties of partial orderings, and drawing diagrams of them:
\begin{defn}[Preceding elements]
    Write $x < y$ if $x \le y$ and $x \neq y$ in $L$. We say that $x$ 
    \emph{precedes} $y$ and write $x \prec y$ if $x < y$ and there is no 
    element $z$ such that $x < z < y$.
\end{defn}

Partial orderings are often depicted using \emph{Hasse diagrams}\index{Hasse diagram}. Some examples are shown in figure \ref{hasse}. Elements of the lattice are shown as nodes, while the relation $\prec$ between elements is shown by connecting nodes with an edge, such that the lesser element is below the greater element in the diagram. For example, figure \ref{hasse:notlattice} shows a four element set with a partial ordering which may be described by the relation $\le$ on the set $\{a,b,c,d\}$ defined by $a \le c$, $b \le c$, $a \le d$, $b\le d$. Hasse diagrams such as these are used to show partial orderings \emph{up to isomorphism}, that is, when we are not interested in the labeling of the nodes, only the nature of the partial order itself. 
\begin{figure}
\begin{center}

\subfigure[A partial ordering that is not a lattice]{
	\label{hasse:notlattice}
	\begin{graph}(4,3.5)(-0.25,-3.5)
	\graphnodesize{0.15}
	\roundnode{x1}(1,-1)
	\roundnode{x2}(2.5,-1)
	\roundnode{x3}(1,-2.5)
	\roundnode{x4}(2.5,-2.5)
	\edge{x1}{x3}
	\edge{x1}{x4}
	\edge{x2}{x3}
	\edge{x2}{x4}
	\end{graph}
}
\hfill
\subfigure[An embedding of the partial ordering in a lattice]{
	\label{hasse:lattice}
	\begin{graph}(4,3.5)(-0.25,-3.5)
	\graphnodesize{0.15}
	\roundnode{x1}(1,-1)
	\roundnode{x2}(2.5,-1)
	\roundnode{x3}(1,-2.5)
	\roundnode{x4}(2.5,-2.5)
	\roundnode{x5}(1.75,-1.75)
	\edge{x1}{x5}
	\edge{x2}{x5}
	\edge{x5}{x3}
	\edge{x5}{x4}
	\end{graph}
}
\hfill
\subfigure[The five element non-modular lattice]{
	\label{hasse:notmodular}
	\begin{graph}(4,3.5)(0,-3.25)
	\graphnodesize{0.15}
	\roundnode{x1}(2,-0.25)
	\roundnode{x2}(1,-1)
	\roundnode{x3}(1,-2)
	\roundnode{x4}(3,-1.5)
	\roundnode{x5}(2,-2.75)
	\edge{x1}{x2} \edge{x2}{x3} \edge{x3}{x5}
	\edge{x1}{x4} \edge{x4}{x5}
	\end{graph}}
\hfill
\caption{Hasse diagrams}
\label{hasse}
\end{center}
\end{figure}

\begin{defn}[Semilattice and Lattice]\index{semilattice|textbf}
An \emph{upper bound} of a subset $T$ of a partially ordered set $S$ is an element $s$ such that $t \le s$ for all $t \in T$. The \emph{least upper bound} of $T$ if it exists (also called \emph{supremum} or \emph{join}) is the upper bound which contains every upper bound. The join of a set $T$ is denoted $\bigvee T$, or if $T$ consists of two elements $x$ and $y$ their join is denoted $x \lor y$.

Similarly a \emph{lower bound} of $T$ is an element $s'$ such that $s' \le t$ for all $t \in T$. The \emph{greatest lower bound} if it exists (also called the \emph{infimum} or \emph{meet} of $T$) is the lower bound which is contained in every other lower bound. The meet of $T$ is denoted $\bigwedge T$; the meet of two elements $x$ and $y$ is denoted $x\land y$.

\index{lattice|textbf}
A \emph{meet semilattice} (or simply \emph{semilattice}) is a partially ordered set in which every pair of elements has a greatest lower bound. Similarly, a \emph{join semilattice} is a partially ordered set in which every pair of elements has a greatest lower bound.

A \emph{lattice} is a partially ordered set in which any two elements have a least upper bound and a greatest lower bound; a lattice is thus both a join and a meet semilattice. A lattice is called \emph{complete} if every subset of $S$ has a least upper bound and greatest lower bound; all finite lattices are complete.
\end{defn}

Figure \ref{hasse:notlattice} shows a partial ordering that is \emph{not} a lattice: the join of the two lesser elements is not well defined, similarly, the meet of the two greater elements is not defined. Figure \ref{hasse:lattice} does show a lattice: the new element acts as the missing join and meet.

A semilattice can be characterised as a semigroup $S$ with the binary operation $\land$ satisfying \emph{idempotence} and \emph{commutativity}:
\begin{eqnarray*}
x \land x & = & x\\
x \land y & = & y \land x
\end{eqnarray*}
respectively. The partial ordering can be recovered by defining $x \le y$ iff $x\land y = x$. Similarly, a lattice can be characterised as a set $S$ together with two operations $\land$ and $\lor$ such that $(S,\land)$ and $(S,\lor)$ are semilattices (according to the above characterisation), satisfying the \emph{absorption} laws:
\begin{eqnarray*}
x \lor (x \land y) & = & x\\
x \land (x \lor y) & = & x
\end{eqnarray*}

\begin{defn}[Modularity]\index{lattice!modular|textbf}
\index{lattice!modular|textbf}
A modular lattice is a lattice $L$ satisfying the \emph{modular identity}: if $x \le z$ then
$$x \lor (y \land z) = (x \lor y) \land z,$$
for all $x,y,z \in L$.
\end{defn}

Figure \ref{hasse:lattice} shows a five element modular lattice, while \ref{hasse:notmodular} shows a lattice that is not modular; it is the only five element non-modular lattice (up to isomorphism).

The proof of the following proposition is in \cite{Birkhoff:48}:
\begin{prop}
The modular lattices are those which do not have the five element non-modular lattice of figure \ref{hasse:notmodular} as a sub-lattice.
\end{prop}

\begin{defn}[Distributivity, Complement and Boolean Algebra]
\index{lattice!distributive|textbf}
A lattice is called \emph{distributive} if it satisfies
\begin{eqnarray*}
x\lor(y\land z) & = & (x\lor y)\land(x\lor z)\\
x\land(y\lor z) & = & (x\land y)\lor(x\land z)
\end{eqnarray*}
A lattice is \emph{complemented} if for every element $a$ there is an element $a'$ such that $a \lor a' = 1$ and $a \land a' = 0$. A complemented distributive lattice is called a \emph{Boolean algebra}\index{Boolean algebra|textbf}.
\end{defn}

\subsection{Functions between partial orders}

It is very important for our work to characterise the nature of functions between partial orderings. Of special importance are those that preserve the partial ordering, and in the case of lattices, preserve meets and joins. We define some important types of functions, and give examples.

\begin{defn}[Order Embeddings]\index{order embedding|textbf}
A function $f$ from one partially ordered set $S$ to another $T$ is called \emph{monotone} or \emph{order-preserving} if $a \le b$ in $S$ implies $f(a) \le f(b)$ in $T$. Conversely, if $f(a) \le f(b)$ implies $a \le b$ then $f$ is called \emph{order-reflecting}. An \emph{order embedding} is a function that is both order-preserving and order-reflecting. A \emph{completion} of a partially ordered set $S$ is an order embedding of $S$ into a complete lattice.
\end{defn}

\begin{defn}[Lattice Homomorphisms]\index{lattice homomorphism|textbf}
If $S$ and $T$ are semilattices, a function $f$ from $S$ to $T$ is a \emph{semilattice homomorphism} if $f(a \land b) = f(a) \land f(b)$ (where $\land$ can represent the meet or the join operation). If $S$ and $T$ are lattices, a \emph{lattice homomorphism} is a function that is both a meet semilattice and join semilattice homomorphism, i.e.~$f(a \land b) = f(a) \land f(b)$, and $f(a \lor b) = f(a) \lor f(b)$. A \emph{lattice isomorphism} is a bijective lattice homomorphism, i.e.~for each element $b$ in $T$ there is exactly one element $a$ in $S$ such that $f(a) = b$. If a lattice isomorphism exists between two lattices they are said to be \emph{isomorphic}.
\end{defn}

Often we may be dealing with partial orders\index{partial ordering} but require something with more structure than that relation provides. For example, we may like to be able to define meets and joins to make the partial ordering into a lattice. The concepts of \emph{principal ideals} and their duals, \emph{principal filters}, allow us to do this:

\begin{defn}[Ideals and Filters]\index{ideal (lattice)}\index{filter}
A \emph{lower set} in a partially ordered set $S$ is a set $T$ such that for all $x,y \in S$, if $x \in T$ and $y \le x$ then $y \in T$. Similarly, an upper set in $S$ is a set $T'$ such that for all $x,y \in S$, if $x\in T$ and $y \ge x$ then $y \in T$.

The \emph{principal ideal generated by an element $x$}\index{principle ideal} in a partially ordered set $S$ is defined to be the lower set $\down{x} = \{y \in S : y \le x\}$. Similarly, the \emph{principal filter generated by $x$} is the upper set $\up{x} = \{y \in S : y \ge x\}$.
\end{defn}

\begin{prop}[Ideal Completion]\label{ideal-completion-prop}\index{ideal (lattice)!completion|textbf}
If $S$ is a partially ordered set, then $\down{\cdot}$ can be considered as a function from $S$ to the powerset $2^S$. Under the partial ordering defined by set inclusion, the set of lower sets form a complete lattice, and $\down{\cdot}$ is a completion of $S$, the \emph{ideal completion}. Similarly, the function $\up{\cdot}$ is the \emph{filter completion} of $S$: it is an embedding into the complete lattice of upper sets, again ordered by inclusion.
\end{prop}





\section{Riesz Spaces and Positive Operators}
 \label{vector-lattices}
 
The previous sections have described formalisms commonly used to describe meaning: broadly speaking, that of vector spaces and that of lattices. Until now, little attention within computational linguistics has been paid to how to combine these two areas. There is a large body of research within mathematical analysis into an area which merges the two formalisms: the study of \emph{partially ordered vector spaces}, \emph{vector lattices} (or \emph{Riesz spaces}), and \emph{Banach lattices}, and a special class of operators on these spaces called \emph{positive operators}.


The definitions and propositions of this section can be found in \cite{Abramovich:02} and \cite{Aliprantis:85}.

\begin{defn}[Partially ordered vector space]\index{vector space!partially ordered|textbf}
A partially ordered vector space $V$ is a real vector space together with a partial ordering $\le$ such that:
\vspace{0.1cm}\\
\indent if $x \le y$ then $x + z \le y + z$\\
\indent if $x \le y$ then $\alpha x \le \alpha y$
\vspace{0.1cm}\\
for all $x,y,z \in V$, and for all $\alpha \ge 0$. Such a partial ordering is called a \emph{vector space order} on $V$. If $\le$ defines a lattice on $V$ then the space is called a \emph{vector lattice}\index{vector lattice|textbf} or \emph{Riesz space}.

\begin{example}[Lattice Structure of $l^p$ Spaces]\index{lp space@$l^p$ space}
The $l^p$ spaces defined earlier are vector lattices under the component-wise partial ordering defined by $x \le y$ if and only if $x_i \le y_i$ for all $i$, where $x = (x_1, x_2, \ldots)$ and $y = (y_1, y_2, \ldots)$.
\end{example}

A vector $x$ in $V$ is called \emph{positive}\index{positive vector|textbf} if $x \ge 0$. The \emph{positive cone} of a partially ordered vector space $V$ is the set $V^+ = \{x \in V : x \ge 0\}$
\end{defn}

The positive cone has the following properties:
\begin{eqnarray*}
X^+ + X^+ \subseteq X^+\\
\alpha X^+ \subseteq X^+\\
X^+ \cap (-X^+) = \{0\}
\end{eqnarray*}
for $\alpha \ge 0$. Any subset $C$ of $V$ satisfying the above three properties is called a \emph{cone} of $V$.

\begin{prop}
If $C$ is a cone in a real vector space $V$, then the relation $\le$ defined by $x \le y$ iff $y - x \in C$ is a vector space order on $V$, with $X^+ = C$.
\end{prop}

Operators which map positive elements to positive elements are called \emph{positive}; there is a large body of work studying such operators.
This idea leads to some useful definitions of particular positive elements of a vector lattice corresponding to an arbitrary element $x$. The \emph{positive part} of $x$ is denoted $x^+$ and is defined by $x^+ = x \lor 0$. Similarly the \emph{negative part} is $x^- = (-x)\lor 0$, and the \emph{absolute value} is $|x| = x \lor -x$. There are a number of useful identities concerning these definitions:
\begin{prop}
The following identities hold for elements $x,y$ in a vector lattice:
\begin{enumerate}[\indent(a).]
\item $x = x^+ - x^-$
\item $|x| = x^+ + x^-$
\item $x\land y = \frac{1}{2}(x + y - |x - y|)$
\item $x\lor y =  \frac{1}{2}(x + y + |x - y|)$
\end{enumerate}
\end{prop}

\subsection{Abstract Lebesgue Spaces}
\label{ALspace}
\index{abstract Lebesgue space|textbf}

A Riesz space together with a norm is called a \emph{normed Riesz space}. If the space is complete with respect to the norm (that is, it is also a Banach space) it is called a \emph{Banach lattice}.

\begin{defn}[Abstract Lebesgue Space]
An Abstract Lebesgue (or AL) space is a Banach lattice $V$ such that 
$$\|x + y\| = \|x\| + \|y\|$$
for all $x,y$ in $V$ with $x \ge 0$, $y \ge 0$ and $x\land y = 0$.
\end{defn}

\section{Algebras}
\label{algebras}

The concept of an \emph{algebra over a field}\index{algebra!over a field|textbf} (or often simply ``an algebra'') is of importance in abstract analysis, and for example providing (in the case of a special type of algebra called a C*-algebra) an alternative formulation of the mathematics of quantum mechanics. They also provide the foundation for the theory of non-commutative probability\index{non-commutative probability}.


\begin{defn}
An algebra is a vector space $A$ over a field $K$ together with a binary operation $(a,b)\mapsto ab$ on $A$ that is bilinear, i.e.
\begin{align*}
a(\alpha b + \beta c) &= \alpha ab + \beta ac\\
(\alpha a+\beta b)c &= \alpha ac + \beta bc
\end{align*}
and associative, i.e. $(ab)c = a(bc)$ for all $a,b,c\in A$ and all $\alpha,\beta \in K$.\footnote{Some authors do not place the requirement that an algebra is associative, in which case our definition would refer to an \emph{associative algebra}.}
\end{defn}


\begin{defn}[Multiplication on $L^1(S)$]\index{algebra!formed from semigroup}
If $S$ is a semigroup then $L^1(S)$ is an algebra under multiplication defined by convolution:
$$(u\cdot v)(x) = \sum_{y,z:yz = x} u(y)v(z),$$
where $u,v \in L^1(S)$ and $x,y,z \in S$.
\end{defn}

\subsection{Linear Operators}
\label{operators}
\index{operator (linear)|textbf}

The concept of an algebra arose through abstraction of concrete examples of algebras, in particular the algebra of \emph{linear operators} on a vector space. These are special kinds of functions on the vector space that agree with the vector space structure.
They are defined as follows:
\begin{defn}[Linear Operator]
A linear operator from a vector space $U$ to a vector space $V$ both over a field $F$ is a function $A$ from $U$ to $V$ satisfying
$$A(\alpha x + \beta y) = \alpha Ax + \beta Ay$$
for all $x,y \in U$ and $\alpha, \beta \in F$.
\end{defn}
Note that the operation of $A$ on an element $x$ is denoted simply $Ax$ (without brackets). In addition, we shall often refer to a linear operator simply as an ``operator''---in this case linearity is assumed.

Linear operators themselves form a vector space, with vector space operations defined by
\begin{eqnarray*}
(A + B)x&=&Ax + Bx\\
(\alpha A)x&=&\alpha Ax\\
0x&=&0
\end{eqnarray*}
Since the operation of functions is necessarily associative, it is easy to see that the linear operators form an algebra under function composition.

We shall often refer to a \emph{linear functional} \index{linear functional} as  a linear function from a vector space to the real numbers; it is especially used in the context of an algebra over a field.

\subsection{Positive operators}
\label{positive-operators}

\index{operator (linear)!positive|textbf}
\begin{defn}[Positive Operators]
An operator $A$ on a partially ordered vector space $V$ is called \emph{positive} if $x \ge 0$ implies $Ax \ge 0$ for all $x\in V$. It is called \emph{regular} if it can be denoted as the difference between two positive operators.
\end{defn}

Surprisingly, the set of regular operators on a vector lattice themselves form a vector lattice:
\begin{prop}[Riesz-Kantarovi\v{c}]
The positive cone defines a vector space order on the vector space of operators on $V$. This order makes the space of regular operators a vector lattice. Specifically the meet and join of two regular operators $A$ and $B$ are given by
\begin{eqnarray*}
(A \land B)x & = & \inf\{Ay + Bz : y,z \in V^+ \text{and } y + z = x\}\\
(A \lor B)x & = & \sup\{Ay + Bz : y,z \in V^+ \text{and } y + z = x\}.
\end{eqnarray*}
\end{prop}

This means that we can use all the constructions of vector lattices with operators on vector lattices; for example we can define the positive and negative parts $A^+$ and $A^-$ of an operator $A$ as $A\lor 0$ and $(-A)\lor 0$ respectively.

\begin{defn}[Lattice Homomorphism]\index{lattice homomorphism (operator)|textbf}
A positive operator $A$ between two vector lattices is called \emph{lattice homomorphism} if $A(x \lor y) = Ax \lor Ay$. A lattice homomorphism that is a one-to-one function is called a \emph{lattice isomorphism}.
\end{defn}

The following proposition shows the importance of lattice homomorphisms:
\begin{prop}
For a positive operator $A$ between two Riesz spaces $U$ and $V$, the following statements are equivalent:
\begin{enumerate}[\indent(a).]
\item $A$ is a lattice homomorphism.
\item $A(x^+) = (Ax)^+$ for each $x \in U$.
\item $A(x \land y) = Ax \land Ay$ for all $x,y \in U$.
\item $|Ax| = A|x|$ for each $x \in U$.
\item $x\land y = 0$ in $U$ implies $Ax \land Ay = 0$ in $V$. 
\end{enumerate}
\end{prop}

 

\bibliographystyle{plainnat}

\addcontentsline{toc}{chapter}{Bibliography}
\bibliography{contexts}

\addcontentsline{toc}{chapter}{Index}
\printindex

\end{document}